\documentclass[10pt,twocolumn,letterpaper]{article}

\usepackage{cvpr}              

\usepackage[accsupp]{axessibility}

\usepackage{xcolor, colortbl}
\usepackage{outlines}
\usepackage{multirow}
\usepackage{algorithm,algorithmic}
\newcommand{\vect}[1]{\boldsymbol{\mathbf{#1}}}

\newlength\myindent
\usepackage{amsmath}
\usepackage{amssymb}
\usepackage{mathtools}
\usepackage{amsthm}
\usepackage{mdframed}
\usepackage{scalerel}

\usepackage{enumitem}
\usepackage{thmtools, thm-restate}
\theoremstyle{plain}
\newtheorem{theorem}{Theorem}[section]

\theoremstyle{definition}
\newtheorem{definition}[theorem]{Definition}
\newtheorem{assumption}[theorem]{Assumption}
\theoremstyle{remark}

\newlength{\textfloatsepsave} 
\def\NB#1{{\color{green}{\bf [NB:} {\it{#1}}{\bf ]}}}
\def\russ#1{{\color{blue}{\bf [russ:} {\it{#1}}{\bf ]}}}

\definecolor{mygray}{gray}{0.9}
\newcommand{\cc}{\cellcolor{gray!20}}

\newcommand{\adjust}[1]{{\leavevmode\color{black}#1}}
\newcommand{\simm}[1]{\sim_{#1}}

\usepackage{soul}
\usepackage{tikz}
\usetikzlibrary{bayesnet, positioning,calc}

\usepackage{etoc} 

\definecolor{cvprblue}{rgb}{0.21,0.49,0.74}
\usepackage[pagebackref,breaklinks,colorlinks,allcolors=cvprblue]{hyperref}

\def\paperID{14043} 
\def\confName{CVPR}
\def\confYear{2025}

\title{Open Set Label Shift with Test Time Out-of-Distribution Reference}

\author{Changkun Ye$^{1,2}$, 
Russell Tsuchida$^{3}$\thanks{Work done while at Data61 CSIRO.}, 
Lars Petersson$^{2}$, 
Nick Barnes$^{1}$\\
$^{1}$Australian National University, 
$^{2}$Data61 CSIRO, ACT, Australia, \\
$^{3}$Data Science and AI Group, Monash University \\ 
{\tt\small \{changkun.ye, nick.barnes\}@anu.edu.au, russell.tsuchida@monash.edu, lars.petersson@csiro.au}
}

\begin{document}

\maketitle

\begin{abstract}
Open set label shift (OSLS) occurs when label distributions change from a source to a target distribution, and the target distribution has an additional out-of-distribution (OOD) class.
In this work, we build estimators for both source and target open set label distributions using a source domain in-distribution (ID) classifier and an ID/OOD classifier. 
With reasonable assumptions on the ID/OOD classifier, the estimators are assembled into a sequence of three stages: 1) an estimate of the source label distribution of the OOD class, 2) an EM algorithm for Maximum Likelihood estimates (MLE) of the target label distribution, and 3) an estimate of the target label distribution of OOD class under relaxed assumptions on the OOD classifier.
The sampling errors of estimates in 1) and 3) are quantified with a concentration inequality.
The estimation result allows us to correct the ID classifier trained on the source distribution to the target distribution without retraining.
Experiments on a variety of open set label shift settings demonstrate the effectiveness of our model. 
\adjust{Our code is available at \url{https://github.com/ChangkunYe/OpenSetLabelShift}.}
\end{abstract}
\vspace{-1em}

\etocdepthtag.toc{mtchapter}
\etocsettagdepth{mtchapter}{subsection}
\etocsettagdepth{mtappendix}{none}

\section{Introduction}
Modern deep learning models demonstrate superior performance over classical models and even human beings in a variety of classification tasks. 
Despite this, real world application of these models can still suffer from unsatisfactory performance due to distribution shift between training data and real world testing data.

\textbf{Label shift}: Label shift is a common type of distribution shift, where the marginal distribution of label $p(y)$ is shifted between source (train) and target (test) domain whereas the conditional distribution $p(x|y)$ is invariant. 
Three main sub-problems of the broader label shift problem are: 1) \emph{detection:} detect if label shift happens; 2) \emph{estimation:} estimate the target domain label distribution given target unlabeled data; 3) \emph{correction:} adapt a classifier trained on the source domain to a target domain.
The label shift problem has been widely studied in the closed set setting, where the source domain and target domain have identical class label set \cite{alexandari2020maximum}. 
The state-of-the-art (SOTA) Closed Set Label Shift (CSLS) models estimate label shift via either solving a linear system \cite{BBSE, RLLS} or running an EM algorithm \cite{mapls, saerens2002adjusting}.  

\textbf{Open Set Label Shift (OSLS)}: Benefiting from the success in the Closed Set Label Shift problem, the more challenging but realistic Open Set Label Shift problem has recently begun to attract research interest~\cite{garg2022domain}. 
In the OSLS problem, the target domain includes data from both in-distribution (ID) classes that are identical to the source domain and an extra out-of-distribution (OOD) class. 
To tackle this problem, \citealt{garg2022domain} analyze the problem under a domain adaptation perspective and propose a model that aims to estimate the target domain label distribution for ID class and the OOD class and correct label shift. 

\begin{figure}
    \centering
    \includegraphics[width=1\linewidth]{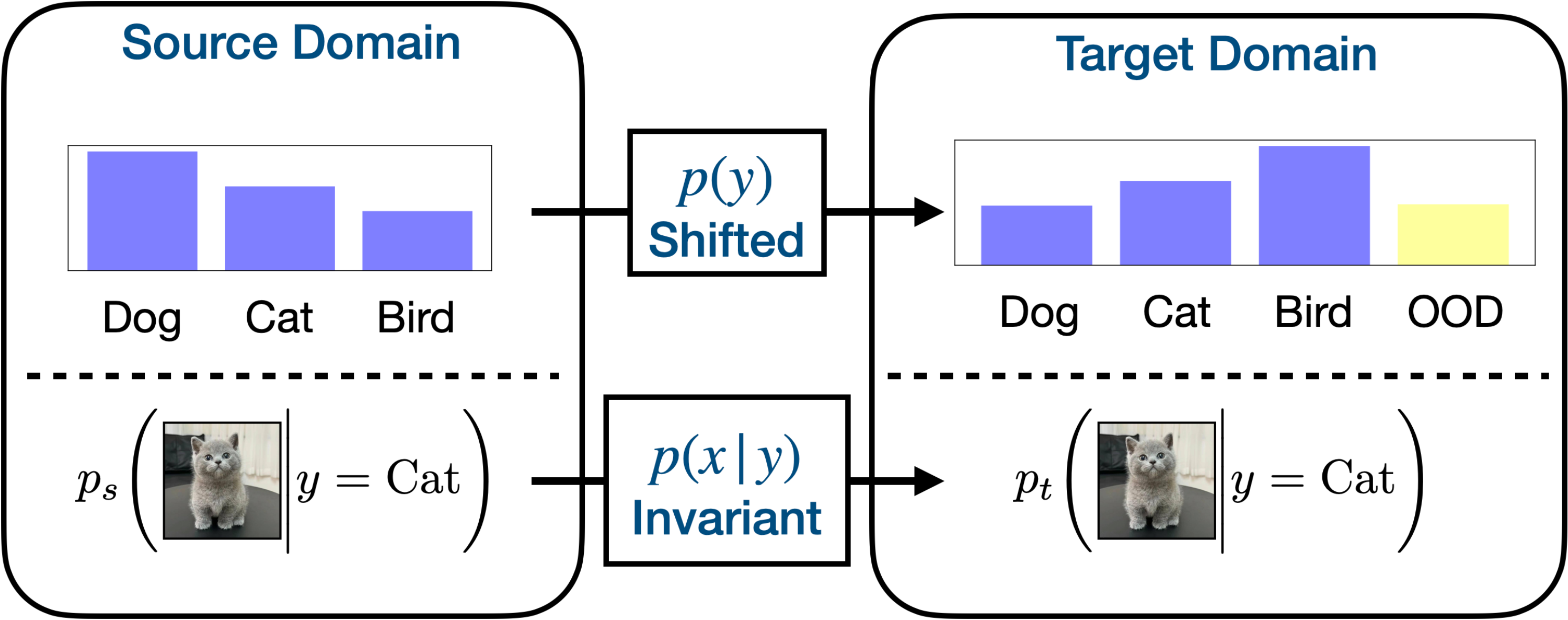}
    \vspace{-2em}
    \caption{\textbf{Open Set Label Shift (OSLS)} set up, where source and target domain have different label distributions $p(y)$ but identical conditional distribution of data given label $p(x|y)$. OSLS extends the Closed Set Label Shift (CSLS) with an extra Out-of-Distribution (OOD) class on the target domain.}
    \label{fig:fig1}
\end{figure}

\begin{figure*}
    \centering
    \includegraphics[width=0.85\linewidth]{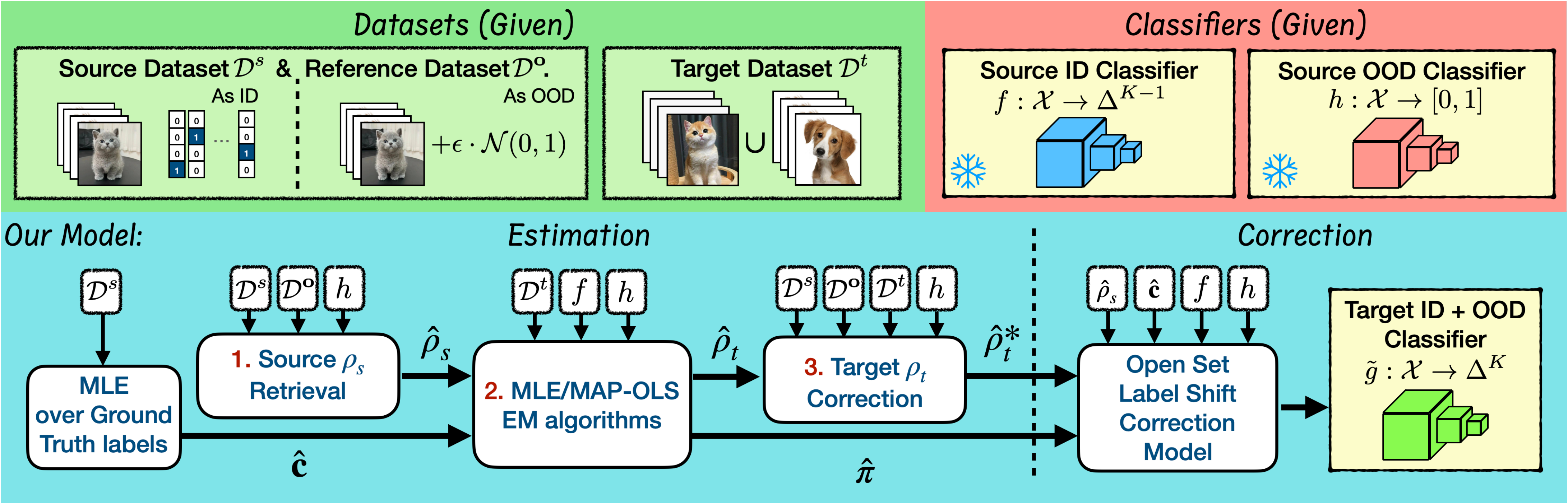}
    \vspace{-1em}
    \caption{\textbf{Structure of our proposed Open Set Label Shift estimation and correction method.} The target ID label distribution probabilities ${p_t(y=\cdot)=\vect{\pi}}$ and ID data ratio ${p_t(b=1)=\rho_t}$ are estimated through three steps: 1) retrieve source ID data ratio ${\rho_s}$ (Sec.~\ref{subsec:rho-s-retrieval}), 2) estimate target ID data ratio ${\rho_t}$ and target ID label distribution $\vect{\pi}$ via an EM algorithm under Assumption~\ref{assume:A1},\ref{assume:A2} (Sec.~\ref{subsec:em}) and 3) correct the target ID data ratio estimator $\hat{\rho}_t$ when OOD classifier $h(x)$ does not satisfy Assumption.~\ref{assume:A2}B (Sec.~\ref{subsec:rho-t-correction}). Based on the estimation result, our correction model constructs a new classifier to classify target domain images. Unlike \citealt{garg2022domain} that retrains the ID/OOD classifier, any OOD classifier proposed in previous OOD detection literature can be used in our model without retraining.}
    \vspace{-2em}
    \label{fig:model-structure}
\end{figure*}

\textbf{Contributions}:
This work focuses on the OSLS estimation and correction problems. 
We propose a novel method for estimating and correcting label shift with an ID classifier and an ID/OOD classifier, without retraining or fine-tuning, where the ID/OOD classifier can be imported from the vast suite of methods available in the OOD detection/Open-Set Recognition literature. 
We derive EM algorithms for the Maximum Likelihood Estimate (MLE) or Maximum {\it a Posteriori} (MAP) estimate of the target label distribution and target ID data ratio with the OOD reference dataset. 
We also propose models to estimate the source ID data ratio and target ID data ratio. 
We test the model on a number of datasets and show superior performance over baselines.

Our main contributions are as follows:
\begin{itemize}
    \item Based on a test time OOD dataset as reference, we propose a novel OSLS model that estimates and corrects open label shift without retraining the ID classifier and ID/OOD classifier. 
    Our method is able to utilize existing OOD detection works without re-training or fine-tuning.
    \item We derive an EM algorithm to obtain the MLE/MAP estimate of the target ID label distribution and target ID data ratio (Theorem~\ref{theorem:em-mle}).
    \item We propose estimators for the  source ID data ratio (which is required by our EM algorithm) and the target ID data ratio for an imperfect OOD classifier. 
    Upper bounds of the sampling error for the two estimators are also provided (Theorem~\ref{theorem:est-rho} and~\ref{theorem:estrho-linear}).
    \item Experimental results demonstrate the superior performance of our method on both label shift estimation error in ID classes and label shift correction accuracy over baselines on CIFAR10/100 and ImageNet-200 datasets with various OOD datasets (\S~\ref{sec:experiments}).
\end{itemize}

\section{Related Works}
\textbf{Closed Set Label Shift (CSLS)} works focus mainly on the \emph{estimation} task. 
Recent methods obtain point estimate of target label distribution based on either EM algorithms or solving linear systems.
EM algorithm based approaches estimate label shift by viewing target label as an unobserved latent variable in a latent variable model. 
\citealt{saerens2002adjusting} propose an EM algorithm model called MLLS to estimate target label distribution. \citealt{alexandari2020maximum} justifies the effectiveness of MLLS and prove that MLLS converges to a MLE estimate. 
\citealt{garg2020unified} provides consistency guarantees of MLLS. 
MAPLS~\cite{mapls} further extend MLLS and derive an EM algorithm to obtain the MAP estimate of the target label distribution and justify their model on large-scale datasets.
Linear system based approaches aim to model and estimate the joint distribution of the ground truth label and the classifier's predicted label and then solve a linear system. This approach can be traced back to last century~\cite{rayens1993discriminant}. 
Recently, BBSE~\cite{BBSE} provide an upper bound of the estimation error of the linear system approach. 
\citealt{garg2021leveraging} unify the EM algorithm and linear system approaches under an optimization perspective and provide consistency analysis on both approaches. 
RLLS~\cite{RLLS} introduces extra constraints over BBSE to reduce estimation error. LTF~\cite{guo2020ltf} utilizes the generative model to estimate label shift with Generative Adversarial Networks (GAN)~\cite{gan}. ELSA~\cite{tian2023elsa} tackles the CSLS estimation task with a fix point iteration algorithm.

\textbf{Open Set Label Shift (OSLS)} was first discussed by \citealt{garg2022domain} under the domain adaptation setting. 
The model leverages the Positive and Unlabeled (PU) learning~\cite{pulearning} approach and proposes a multi-stage unsupervised domain adaptation framework that retrains the ID/OOD classifier to estimate the target ID/OOD label distribution. 
Unlike their methods, our approach extends the CSLS problem setup and aims to solve the OSLS problem by utilizing OOD classifiers in the existing OOD detection literature without fine-tuning. Such a setup greatly increases the flexibility of the model, especially when the source domain classifier is frozen or re-training is expensive in practice. 

\textbf{Other Advanced Label Shift} settings are also studied in the recent years. For example, some works~\cite{zhao2021active,wu2021online,baby2024online,qian2023handling} explore the label shift problem in the active learning or online learning setting, \citealt{zhang2021coping} analyze the Adversarial Label Shift setting and \citealt{maity2022minimax} discuss the Supervised Label Shift problem. These methods have different problem setups and thus will not be compared. 

\textbf{OOD detection and Open Set Domain Adaptation} are also related to the OSLS problem, where~\cite{openmax, msp, odin, mds, opengan, ebo, gram, sun2021react, mls, vim, knn, ash, RotPred, godin, csi, aprl, vaze2021open, miller2021accuracy, hein2019relu, meinke2019towards, fang2022out, panareda2017open,saito2018open, liu2019separate, fang2020open, zhang2021learning,wang2023learning} are discussed in Appendix~\ref{Asec:related-works} due to the space limit.

\section{Problem Setup}


Let $\mathcal{X}\subseteq \mathbb{R}^d$ be the data space, $\mathcal{Y} = \{1,2,...,K\}$ be the label space and $\mathcal{Y}\cup\{K+1\}$ be the open label space with $K+1$ as the class assigned to OOD data. 
We use ${p_s(x,y=\cdot)}$ and ${p_t(x,y=\cdot)}$ respectively to denote the source and target domain joint data and label distributions, $\Delta^{K-1}$ to denote the $K$-dimensional probability simplex.
To model ID versus OOD data, we introduce binary random variables $B_s,B_t$ on the source and target domain respectively with $B_s,B_t=1$ and $B_s,B_t=0$ respectively mean ID and OOD.

\subsection{Graphical model setup}
\textbf{Likelihoods and priors:} 
Without loss of generality, we parameterise the source label distribution $Y_s|B_s=1\sim\text{Cat}(K,\vect{c})$ and target label distribution $Y_t|B_t=1\sim\text{Cat}(K,\vect{\pi})$ both as categorical distributions. 
Let the ID indicator $B_s,B_t$ follow Bernoulli distributions $B_s\sim\text{Bern}(\rho_{s})$ and $B_t\sim\text{Bern}(\rho_{t})$ , with $p_s(b=1)=\rho_s$ and $p_t(b=1)=\rho_t$ as the probability of the data being ID on source and target domain respectively. 
Formally, we are given:
\begin{equation}\label{eq:label-distribution}
\begin{aligned}
& p_s(y|b;\vect{c})=
\begin{cases}
  c_j, & \text{if } b=1, y\in\mathcal{Y}\\
  1, & \text{if } b=0, y=K+1 \\
  0, & \text{otherwise}\\
\end{cases},
\\
& p_t(y|b;\vect{\pi})=
\begin{cases}
  \pi_j, & \text{if } b=1, y\in\mathcal{Y}\\
  1, & \text{if } b=0, y=K+1 \\
  0, & \text{otherwise}.
\end{cases}
\end{aligned}
\end{equation}
We optionally place priors over the target label ID/OOD ratio $\rho_t$ and the target ID probabilities $\vect{\pi}$, which are further discussed in subsequent sections.
We treat the source label ID/OOD ratio $\rho_s$ and the source ID probabilities $\vect{c}$ as deterministic parameters.

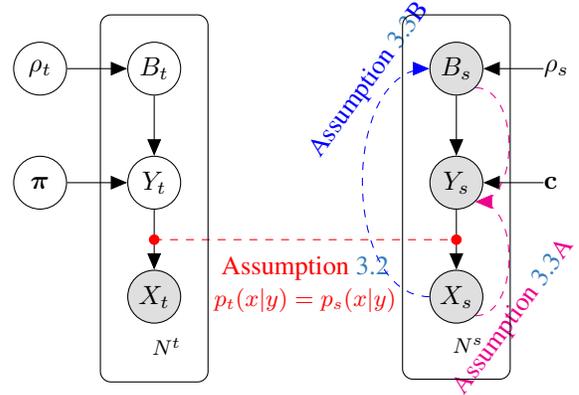
\begin{figure} 
    \centering
    \begin{tikzpicture}

\node[latent] (Bt) at (0, 2) {$B_t$};
\node[latent,   below=0.8cm of Bt] (Yt) {$Y_t$};
\node[obs,      below=0.8cm of Yt] (Xt) {$X_t$};

\node[latent,   left=0.8cm of Bt] (rho_t) {$\rho_t$};
\node[latent,   left=0.8cm of Yt] (pi) {$\vect{\pi}$};

\node[obs,      right=0.8cm of Bt, xshift=2.5cm] (Bs) {$B_s$};
\node[obs,      below=0.8cm of Bs] (Ys) {$Y_s$};
\node[obs,      below=0.8cm of Ys] (Xs) {$X_s$};

\node[const,    right=0.8cm of Bs] (rho_s) {$\rho_s$};
\node[const,    right=0.8cm of Ys] (c) {$\vect{c}$};

\edge {rho_t} {Bt};
\edge {pi} {Yt};
\edge {Bt} {Yt};
\edge {Yt} {Xt};

\edge {rho_s} {Bs};
\edge {c} {Ys};
\edge {Bs} {Ys};
\edge {Ys} {Xs};

\coordinate (midpoint1) at ($(Xt)!0.5!(Yt)$);
\coordinate (midpoint2) at ($(Xs)!0.5!(Ys)$);
\draw[dashed, color=red] (midpoint1) -- (midpoint2);
\coordinate (midpoint3) at ($(midpoint1)!0.5!(midpoint2)$);

\node[below=0.1cm of midpoint3, align=center] (labelshift) {\textcolor{red}{Assumption~\ref{assume:A1}} \\ \small \textcolor{red}{$p_t(x|y)=p_s(x|y)$}};

\fill[red] (midpoint1) circle (2pt);
\fill[red] (midpoint2) circle (2pt);

\draw[->, dashed, bend left=90, color=blue] (Xs) to node[pos=0.8, sloped, above, color=blue] {Assumption~\ref{assume:A2}B} (Bs);
\draw[->, dashed, bend left=270, color=magenta] (Xs.south east) to node[pos=0.2, sloped, below, color=magenta] {Assumption~\ref{assume:A2}A} (Ys.south east);
\draw[->, dashed, bend left=90, color=magenta] (Bs.south east) to node[pos=0.2, sloped, below, color=magenta] {} (Ys.south east);

\plate[inner sep=0.35cm] {plate1} {(Bt) (Yt) (Xt)} {$N^t$};
\plate[inner sep=0.35cm] {plate2} {(Bs) (Ys) (Xs)} {$N^s$};

\end{tikzpicture}
    \caption{
    \textbf{Graphical model of the Open Set Label Shift setting and our assumptions.} $X_s, X_t$ are data for the source and target domain, $Y_s, Y_t$ are the corresponding categorical-valued labels and $B_s, B_t$ are binary values representing ID/OOD data. 
    $\vect{c}, \vect{\pi}$ are source and target domain label distribution class probabilities. 
    Source domain data $X_s$ is observed with ground truth ID data in $\mathcal{D}^s$ and reference OOD data in $\mathcal{D}^{\textbf{o}}$. 
    $p(x|y)$ are invariant under the label shift assumption. 
    With the help of a reference OOD dataset $\mathcal{D}^{\text{o}}$ at test time, we first estimate $\rho_s, \vect{c}$ and then $\rho_t, \vect{\pi}$ without retraining. Optional prior distributions are employed on $\rho_t, \vect{\pi}$.
    }
    \label{fig:dag}
\end{figure}

\textbf{OSLS problem definition:} 
In the OSLS problem, we can also define the \emph{detection}, \emph{estimation} and \emph{correction} task by analogy with the CSLS problem~\cite{mapls}:
\begin{definition}\label{def:osls-problem}
(\textbf{Open Set Label Shift Problem})

    \noindent Under Assumption~\ref{assume:A1}, given:
    \begin{itemize}
        \item Source domain ID labeled data $\mathcal{D}^s = \{(x^s_i,y^s_i)\}^{N^s}_{i=1}$, where ${(x^s_i,y^s_i)\simm{i.i.d} p_s(x,y=\cdot|b=1)}$;
        \item Target domain unlabeled data $\mathcal{D}^t=\{x^t_i\}^{N^t}_{i=1}$, where ${x^t_i\simm{i.i.d} p_t(x)}$;
        \item Source ID $K$-class classifier $f:\mathcal{X}\rightarrow\Delta^{K-1}$;
        \item Source ID/OOD classifier $h:\mathcal{X}\rightarrow[0,1]$ ($0$ for OOD),
    \end{itemize}
    the open set label shift problem is to solve
    \begin{itemize}
        \item \emph{Detection}: Verify $p_s(y|b=1) \neq p_t(y|b=1)$;
        \item \emph{Estimation:} Estimate $p_t(y=\cdot|b=1)$, ${p_t(b=\cdot)}$;
        \item \emph{Correction:} Model $p_t(y=\cdot|x)$ with $f$ and $h$. 
    \end{itemize}
\end{definition}
A graphical model depiction of the OSLS estimation problem is given in Figure~\ref{fig:dag}.
We focus on the estimation and correction problems.
That is, our overarching goal is to first estimate the target ID label distribution $\vect{\pi}$ and target ID data ratio $\rho_t$, and then use these estimates to build a better classifier and OOD detector.

\subsection{Assumptions}
Similar to the closed set label shift problem~\cite{mapls}, we study the open label shift problem based on the assumption that source and target domain have identical conditional distributions of data $x$ given label $y$:
\begin{assumption}{(\textbf{Open Set Label Shift Assumption})}\label{assume:A1}
\[p_s(x|y=i) = p_t(x|y=i) \qquad \text{for all\,}  i\in\mathcal{Y}\cup \{K+1\}.\]
\end{assumption}
Here we naturally extend the closed set label shift assumption for $K$ classes to include the OOD $K+1$ class.

In Definition~\ref{def:osls-problem}, the ID classifier $f$ can be obtained by training a NN model on source domain data via supervised learning. 
Any of the models proposed in the OOD detection literature can be used for the OOD classifier $h$.
Due to the absence of OOD training data, $h$ is usually established on intuitive principles~\cite{godin} and hence their test performance lacks theoretical guarantees.
Our main assumption is that regardless of their origins, both $f$ and $h$ can be understood as posterior predictive models, which respectively describe the probability of the ID label $y$ given the input data $x$ for in distribution data $b=1$, and the probability that the data is in distribution given the input data $x$.


\begin{assumption}\label{assume:A2}
For all $(x,y)\in\mathcal{X}\times (\mathcal{Y}\cup\{K+1\})$:
\begin{itemize}[leftmargin=9em,noitemsep,topsep=0pt]
\item[\textbf{Assumption~\ref{assume:A2}A}] $p_s(y|x,b=1) = f(x)$, \quad \text{and }
\item[\textbf{Assumption~\ref{assume:A2}B}] $p_s(b=1|x) = h(x).$
\end{itemize}
\end{assumption}

We superimpose our assumptions onto the graphical model in Figure~\ref{fig:dag}. The validity of Assumption~\ref{assume:A2}A for practical classifiers is justified empirically in the experiments (\S\ref{subsec:ablation}) and the case when Assumption~\ref{assume:A2}B is not satisfied is discussed in \S\ref{subsec:rho-t-correction}.

\textbf{Practical Classifier Choices}: In practice, an ID classifier that satisfies the OSLS problem setup (Definition~\ref{def:osls-problem}) can be obtained by training a Neural Network classifier on the source domain dataset $\mathcal{D}^s$ through supervised learning. On the other hand, according to the OOD detection literature, the OOD classifier can be obtained based on the Neural Network ID classifier without ground truth OOD samples~\cite{odin, godin}. For example, OpenMax~\cite{openmax} fits image feature from each ID class with a Weibull distribution. The ID/OOD samples are distinguished based on the likelihood of the test samples for each ID class distribution. In this way, the ID/OOD classifier can be obtained with $\mathcal{D}^s$ given in Definition~\ref{def:osls-problem}. All the ID/OOD classifiers we consider in the experiments satisfy this requirement.

\section{Proposed Method}

\subsection{Method Overview}
Our proposed method focus mainly on the OSLS \emph{estimation} problem. The general idea of our model is to transform the OSLS problem into a mathematical form similar to the Closed Set Label Shift problem. Then we can leverage the advances of the CSLS methods to derive algorithms feasible for the OSLS problem. The main challenge lies in both problem transformation and algorithm derivation.

In our estimation model, we first provide a three-stage algorithm to estimate open set label shift, which requires the availability of a reference OOD dataset ${\mathcal{D}^{\textbf{o}}=\{x^o_i\}^{N^o}_{i=1}}$ at test time.
The \emph{estimation} model includes: 1) the source ID data ratio $\rho_s$ retrieval model; 2) the EM algorithm based target label distribution estimation model 
and 3) the target ID ratio $\rho_t$ correction model. The choice of reference OOD dataset is then discussed. 
Finally, an OSLS \emph{correction} model is also introduced. All the mathematical proofs can be found in Appendix.~\ref{Asec:proofs}.

\textbf{Remark on OOD Dataset $\mathcal{D}^{\textbf{o}}$}: Although our OSLS \emph{estimation} model is derived by assuming $\mathcal{D}^{\textbf{o}}$ contains ground truth OOD samples, this requirement can be relaxed (\S~\ref{subsec:ood-choice}) so that pseudo OOD samples can be used instead and the OSLS problem setup (Definition~\ref{def:osls-problem}) is satisfied. This use of pseudo OOD samples leads to our algorithms with demonstrated empirical performance benefits over existing models, using the same form of input training data, as shown in the experiment section. 

\subsection{Source ID/OOD Data Ratio retrieval} \label{subsec:rho-s-retrieval}

Several parameters have to be estimated before transforming the OSLS estimation problem into a CSLS estimation problem. We use the standard maximum likelihood estimator to estimate the source domain ID label distribution parameters $\vect{c}=p_s(y=\cdot)$, which amounts to computing the relative empirical frequencies of ID/OOD data on the source domain.
However, estimating the probability of ID $p_s(b=1)=\rho_s$ requires more careful attention.

This section aims to estimate the source domain ID data ratio $p_s(b=1)=\rho_s$ (Fig.~\ref{fig:dag}) under the OSLS problem setup, where only source domain ID dataset $\mathcal{D}^s$ and an OOD classifier $h(x)$ is available. In this work, we treat $h(x)$ as a classifier pre-trained on some unknown source domain dataset with both ID and OOD data, and estimate ${\rho_s}$ for that dataset with $h$, $\mathcal{D}^s$ and the reference OOD dataset $\mathcal{D}^\textbf{o}$.





We consider the following estimate of $\rho_s$:
\begin{equation}\label{eq:rho-s-estimator}
    \hat{\rho}_s = \frac{\hat{\mu}_0}{1 - \hat{\mu}_1 + \hat{\mu}_0},
\end{equation}
where 
\begin{equation}\label{eq:est-rho-s-sigma}
    \hat{\mu}_0 := \frac{1}{|\mathcal{D}^{\textbf{o}}|}\sum_{x\in\mathcal{D}^{\textbf{o}}}h(x) \quad \text{and} \quad \hat{\mu}_1 := \frac{1}{|\mathcal{D}^s|}\sum_{x\in\mathcal{D}^s}h(x).
\end{equation}
Utilizing concentration inequalities~\cite{vershynin2018high}, the 
error of estimate may be quantified.
\begin{restatable}{theorem}{estimaterho}
    \label{theorem:est-rho}
    (\textbf{Source ID/OOD ratio estimator}) 
    Under Assumption~\ref{assume:A2}B, given source ID dataset $\mathcal{D}^s$ and source OOD dataset $\mathcal{D}^{\textbf{o}}$, then for $\delta>0$, with probability of at least $1 - 2\delta$,
    \begin{equation}
        | \rho_s -  \hat{\rho}_s | \leq \frac{1}{1 - \mu_1 + \mu_0} \sqrt{\frac{\log 1/ \delta}{2\min(|\mathcal{D}^{\textbf{o}}|, |\mathcal{D}^s|)}} 
    \end{equation}
    where $\mu_0:=\mathbb{E}_{X_s|B_s=0}[h(x)]$, $\mu_1:=\mathbb{E}_{X_s|B_s=1}[h(x)]$. 
\end{restatable}
A variant of theorem~\ref{theorem:est-rho} can be shown and applied to the CSLS problem, multi-class case (see Appendix~\ref{Asubsec:est-rho-extension}), which enables retrieval of the source label distribution $\vect{c}$ under the absence of the source domain dataset $\mathcal{D}^s$.
We leave empirical investigation of the extended result for future work, as the focus of our work is on the OSLS problem.



\subsection{EM algorithm for OSLS Estimation} \label{subsec:em}
With the estimate of $\rho_s, \vect{c}$ obtained in the previous section, this section reformulate the OSLS estimation objective similar to a CSLS estimation objective and presents the EM algorithms to estimate the target label distribution $p_t(y=\cdot)=\vect{\pi}$ and target ID data ratio $p_t(b=1)=\rho_t$. 

\paragraph{Negative log likelihood} Based on Assumption~\ref{assume:A1}, \ref{assume:A2}, we can construct the negative log likelihood (NLL) of the target label distribution $p_t(y=\cdot)=\vect{\pi}$ and ID data ratio $p_t(b=1)=\rho_t$ given target unlabeled data $\mathcal{D}^t$.
\begin{restatable}{lemma}{lemmaobj}
\label{lemma:obj}
    Under Assumption~\ref{assume:A1},~\ref{assume:A2}, given $\mathcal{D}^t$, the negative log likelihood ${-\log L(\vect{\pi},\rho_t;\mathcal{D}^t)}$ can be written as:
    \begin{equation}\label{eq:nll}
    \begin{aligned}
        - \log L(\vect{\pi},\rho_t;\mathcal{D}^t) = - \sum^{N^t}_{i=1} \log \left(\sum^{K+1}_{j=1}\frac{\Tilde{\pi}_j}{\Tilde{c}_j} \Tilde{f}(x_i)_j \right) + C, \\
    \end{aligned}
    \end{equation}
    where $C$ does not depend on either $\vect{\pi}$ or $\rho_t$ and 
    \begin{equation}\label{eq:tilde-f}
    \Tilde{f}(x)_i := 
    \begin{cases}
    h(x)\cdot f(x)_i,  & i\in \mathcal{Y} \\
    1 - h(x), & i = K + 1,
    \end{cases},
\end{equation}
\begin{equation} \label{eq:tilde-pi-and-c}
\begin{aligned}
    \Tilde{\vect{\pi}} & := [\rho_t\cdot\pi_1,...,\rho_t\cdot\pi_K, 1 - \rho_t]^T \\
    \Tilde{\vect{c}} & := [\rho_s\cdot c_1, ...,\rho_s\cdot c_K, 1 - \rho_s]^T.
\end{aligned}
\end{equation}
\end{restatable}

\paragraph{Maximum likelihood estimation} 
That the negative log likelihood in~\eqref{eq:nll} has the same form of the NLL of a Closed Set Label Shift estimation problem with $K+1$ classes (see Appendix.~\ref{Asec:related-works} or \citealt{alexandari2020maximum}).
Observing this similarity, we can minimize the NLL in~\eqref{eq:nll} by viewing $\Tilde{f}$ as the closed set $K+1$ class classifier and $\Tilde{\vect{c}},\Tilde{\vect{\pi}}$ as parameters of the closed set source and target label distribution. The MLE of target label distribution parameters $\vect{\pi}^{\text{MLE}},\rho_t^{\text{MLE}}$ can be obtained via: 

\begin{equation}\label{eq:MLE-obj}
    \vect{\pi}^{\text{MLE}},\rho_t^{\text{MLE}} \in \underset{\Tilde{\pi}\in\Delta^K}{\arg\min} - \log L(\vect{\pi},\rho_t;\mathcal{D}^t).
\end{equation}
Although the MLE objective~\eqref{eq:MLE-obj} is not convex in $(\vect{\pi}, \rho_t)$, 
we may still derive an EM-algorithm which converges to the global minimum.
Inspired by \citealt{saerens2002adjusting}, we 
compute a reparameterised MLE of $\Tilde{\vect{\pi}}$ in ~\eqref{eq:tilde-pi-and-c}. As MLE is invariant under reparameterisation~\cite{murphy2012machine}, this MLE can be mapped back to the MLE of $\vect{\pi}$ and $\rho_t$. 
Details are described in Theorem~\ref{theorem:em-mle}.
\begin{restatable}{theorem}{propemmle}\textbf{(MLE)}
\label{theorem:em-mle}
    Under Assumption~\ref{assume:A1},~\ref{assume:A2}, the the NLL~\eqref{eq:nll} is convex in $\Tilde{\vect{\pi}}$ (and convex in $\rho_t$), and the EM algorithm MLE-OLS (Alg.~\ref{alg:MAPOLS}) converges to $\vect{\pi}^{\text{MLE}},\rho_t^{\text{MLE}}$~\eqref{eq:MLE-obj}.
\end{restatable}

\paragraph{Maximum a-posteriori estimation}: We may also attempt to compute MAP estimates $\vect{\pi}^{\text{MAP}},\rho_t^{\text{MAP}}$, when prior information about the two parameters are available. 
However the MAP is not invariant under reparameterisations, and the posterior probability density is nonconvex in $(\vect{\pi}, \rho_t)$. 
This makes it difficult to compute MAP estimates, and in this sense the MLE is favourable. 
We detail the use of a Dirichlet prior over ${\vect{\pi}\sim\text{Dir}(K,\vect{\alpha}^{\textbf{in}})}$ and a Beta prior over $\rho_t\sim\text{Beta}(\alpha_1^{\textbf{out}}, \alpha_2^{\textbf{out}})$ in Appendix~\ref{Asubsec:map}.


Alg.~\ref{alg:MAPOLS} summarises the EM algorithm for MLE and MAP estimation ($\mathbb{R}^K_{> 1}:=\{x\in\mathbb{R}^K|x_i> 1, i=1,...,K\}$).
\setlength{\textfloatsepsave}{\textfloatsep} 
\setlength{\textfloatsep}{0pt}
\begin{algorithm}[h]
\caption{MLE/MAP-OLS}
	\begin{algorithmic}
	    \label{alg:MAPOLS}
	    \STATE \textbf{Input: }$\mathcal{D}^t=\{x^t_i\}^{N^t}_{i=1}, \vect{c}, \rho_s, h(x), f(x)$,
        \begin{itemize}
            \item MLE-OLS: $\vect{\alpha}^{\textbf{in}}=\vect{1}$, $\alpha_1^{\textbf{out}}, \alpha_2^{\textbf{out}}=\vect{1}$.
            \item MAP-OLS: $\vect{\alpha}^{\textbf{in}}\in\mathbb{R}^{K}_{> 1}$, $\alpha_1^{\textbf{out}}, \alpha_2^{\textbf{out}}\in\mathbb{R}_{> 1}$.
        \end{itemize}
	    \STATE \textbf{Initialize:} $\vect{\pi}^{(0)}\in \Delta^{K-1}_{>0}, \rho^{(0)}_t\in (0, 1)$.
        \STATE \textbf{Construct:} $\Tilde{f},\Tilde{c}$ based on~\eqref{eq:tilde-f},\eqref{eq:tilde-pi-and-c}.
		\FOR{$m=0$ to $M$}
        \STATE \textbf{Construct: } $\Tilde{\vect{\pi}}^{(m)}$ based on $\vect{\pi}^{(m)},\rho^{(m)}_t$ and ~\eqref{eq:tilde-pi-and-c}.
        \STATE \textbf{E-step:} For $j\in\mathcal{Y}\cup\{K+1\}$, evaluate
        \STATE\begin{equation}
                g_{ij}^{(m)} = \frac{\Tilde{\pi}^{(m)}_j/ \Tilde{c}_j \cdot \Tilde{f}(x^t_i)_j}{\sum^K_{l=1}\Tilde{\pi}^{(m)}_l/ \Tilde{c}_l \cdot \Tilde{f}(x^t_i)_l}.
            \end{equation}
        \STATE  \textbf{M-step:}  For $j\in\mathcal{Y}$, evaluate 
        \STATE \begin{equation}\label{alg-eq:M}
                \left\{
                \begin{aligned}  
                    \pi_j^{(m+1)} & = \frac{\sum^{N^t}_{i=1} g_{ij}^{(m)} + \alpha_j^{\textbf{in}} - 1}{N^t  - \sum^{N^t}_{i=1}g_{iK+1}^{(m)} + \sum^K_{l=1}(\alpha_l^{\textbf{in}} - 1)} \\
                    \rho_t^{(m+1)} &  = \frac{N^t  - \sum^{N^t}_{i=1}g_{iK+1}^{(m)} + \alpha_1^{\textbf{out}} - 1}{N^t + \alpha_1^{\textbf{out}} + \alpha_2^{\textbf{out}} - 2}.
                \end{aligned}
                \right.
                \end{equation}
		\ENDFOR
		\STATE \textbf{Output: } $p_t(y=\cdot)=\vect{\pi}^{(M+1)}, p_t(b=1)=\rho_t^{(M+1)}$.
	\end{algorithmic}  
\end{algorithm}  
\setlength{\textfloatsep}{\textfloatsepsave}

\subsection{Target ID/OOD Data Ratio Correction}\label{subsec:rho-t-correction}
In Assumption~\ref{assume:A2}, we describe the conditional distribution $p(b=1|x)$ with an OOD classifier $h(x)$. 
In practice, however, the OOD classifiers can yield unsatisfactory performance due to the challenging OOD detection problem setup~\cite{zhang2023openood}. 
Deploying such a classifier in the OSLS algorithms can result in high estimation error. 

This section provides a correction model to mitigate the possible estimation error on $\rho_t$. 
We find that $\rho_t$ can still be estimated with a practical OOD classifier $h'$ that doesn't satisfies~\ref{assume:A2}B, if $h'(x)$ has different expected response to ID and OOD samples but identical response to samples in different ID classes: 
\begin{equation}\label{eq:rho-t-correction-condition}
\begin{aligned}
    & \mathbb{E}_{X_s|Y_s=i} [h'(x)] \neq \mathbb{E}_{X_s|Y_s=K+1} [h'(x)] \quad \text{and} \\
    & \mathbb{E}_{X_s|Y_s=i}[h'(x)]  = \mathbb{E}_{X_s|Y_s=j}[h'(x)] \text{ for all } i,j \in\mathcal{Y}.
\end{aligned}
\end{equation}

\begin{restatable}{theorem}{rhocali}\label{theorem:estrho-linear}(\textbf{Target ID/OOD ratio correction})
    Under Assumption~\ref{assume:A1}, \ref{assume:A2}A (without \ref{assume:A2}B), for a classifier ${h':\mathcal{X}\rightarrow[0,1]}$ that satisfies \eqref{eq:rho-t-correction-condition}, 
    given source ID dataset $\mathcal{D}^{s}$, OOD dataset $\mathcal{D}^{o}$, target dataset $\mathcal{D}^t$, then for $\delta>0$, with probability of at least $1-2\delta$ we have:
    \begin{equation}
    \begin{aligned}
        \vert \rho_t - \hat{\rho}^*_t \vert \leq \frac{1}{\vert \mu_1' - \mu_0'\vert} \sqrt{\frac{2 \log 1/ \delta}{\min(|\mathcal{D}^s|, |\mathcal{D}^{\textbf{o}}|,|\mathcal{D}^t|)}},
    \end{aligned}
    \end{equation}
    where
    \begin{equation} \label{eq:rho-correction}
        \hat{\rho}^*_t = \frac{\hat{\rho}' - \hat{\mu}_0'}{\hat{\mu}_1' - \hat{\mu}_0'}, \quad \text{and} \quad \hat{\rho}'  := \frac{1}{|\mathcal{D}^t|}\sum_{x_i\in\mathcal{D}^t} h'(x_i),
    \end{equation}
    with $\hat{\mu}_1',\hat{\mu}_0'$ and ${\mu_1'},{\mu_0'}$ defined in the same way as~\eqref{eq:est-rho-s-sigma} and Theorem.~\ref{theorem:est-rho} but substitute $h$ with $h'$.
\end{restatable}



The condition \eqref{eq:rho-t-correction-condition} in Theorem~\ref{theorem:estrho-linear} is a reasonable assumption because the first equation holds when $h'$ can roughly separate ID/OOD samples in the output space $[0,1]$. The second equation is likely to hold when $h'$ is trained/constructed based on a class-uniform ID dataset. 



Based on \eqref{eq:rho-correction} in Theorem~\ref{theorem:estrho-linear}, we propose a correction model for the $\rho_t^{\text{MLE}}$ and $\rho_t^{\text{MAP}}$ obtained in Alg.~\ref{alg:MAPOLS} via:
\begin{equation}\label{eq:rho-t-correction}
    \rho_t^{\text{MLE*}} = \frac{\rho_t^{\text{MLE}} - \hat{\mu}_0}{\hat{\mu}_1 - \hat{\mu}_0} \quad \text{and}\quad  \rho_t^{\text{MAP*}} = \frac{\rho_t^{\text{MAP}} - \hat{\mu}_0}{\hat{\mu}_1 - \hat{\mu}_0}.
\end{equation}

Due to the space limit, further discussion about \eqref{eq:rho-t-correction} are provided in Appendix~\ref{Asubsec:rho-t-correction-discussion} and empirical analysis in Fig.~\ref{fig:rho-t-ablation}.


\subsection{Choice of OOD Reference Dataset}\label{subsec:ood-choice}


In our OSLS estimation model, only $\hat{\mu}_0$ directly depends on the OOD reference dataset $\mathcal{D}^{\textbf{o}}$ (\eqref{eq:est-rho-s-sigma} and \eqref{eq:rho-t-correction}). Thus as long as the expectation of $h(x)$ on the distribution that generates $\mathcal{D}^{\textbf{o}}$ equals to the expectation of $h(x)$ on the ground truth OOD distribution, our model can yield desired estimates. 
Here we generate the OOD reference dataset by a linear combination of Gaussian noise and source domain ID samples in $\mathcal{D}^s$. With $\gamma\in(0,1)$ we have:
\begin{equation}\label{eq:ood-dataset-gen}
    \mathcal{D}^{\textbf{o}}_{\gamma}= \{(1 - \gamma) \cdot x_i + \gamma\cdot \epsilon\vert x_i\in\mathcal{D}^s, \epsilon\sim\mathcal{N}(0,1)\},
\end{equation}

We choose $\gamma$ to be close to 0 so that samples of $\mathcal{D}^{\textbf{o}}_{\gamma}$ will be close to the ground truth ID samples in $\mathcal{D}^s$. 
In this case, the $\hat{\mu}_0$ the computed by $\mathcal{D}^{\textbf{o}}_{\gamma}$ could be higher than that obtained with actual OOD samples, therefore we introduce another re-weight factor $T$ so that:
\begin{equation}\label{eq:sigma-0-hat}
    \hat{\mu}_0^* = \frac{1}{\vert \mathcal{D}^{\textbf{o}}_{\gamma} \vert T} \sum_{x_i\in\mathcal{D}^{\textbf{o}}_{\gamma}} h(x_i),
\end{equation}
which is then used as $\hat{\mu}_0$ in \eqref{eq:est-rho-s-sigma} and \eqref{eq:rho-t-correction} in our model.



\subsection{OSLS correction method}
The OSLS correction model can be implemented based on the Closed Set Label Shift correction model of $K+1$ classes~\cite{mapls}. 
With estimates of the parameters $\vect{c}, \rho_s$ of the source label distribution and parameters $\vect{\pi},\rho_t$ of the target label distribution, we can construct the source and target label distribution of all classes with $\Tilde{c},\Tilde{\pi}$ based on~\eqref{eq:tilde-pi-and-c} and the source domain classifier $\Tilde{f}$ in~\eqref{eq:tilde-f} for $K+1$ classes. 
The target domain classifier can be constructed via:
\begin{equation}\label{eq:lsc}
    \Tilde{g}(x) = \frac{ \frac{\Tilde{\pi}_j}{\Tilde{c}_j}  \Tilde{f}(x)_j}{\sum^{K+1}_{l=1} \frac{\Tilde{\pi}_l}{\Tilde{c}_l} \Tilde{f}(x)_l}.
\end{equation}

\subsection{Overall Framework}
Our practical OSLS-EM model estimate and correction open set label shift follows the procedure in Alg.~\ref{alg:OSLS-EM}.
\setlength{\textfloatsepsave}{\textfloatsep} 
\setlength{\textfloatsep}{0pt}
\begin{algorithm}[h]
\caption{OSLS-MLE/MAP Framework}
	\begin{algorithmic}
	    \label{alg:OSLS-EM}
	    \STATE \textbf{Input: }$\mathcal{D}^t,\mathcal{D}^s, h(x), f(x)$, hyper-params $\gamma, T$.
        \STATE \textbf{Optional: (MAP prior)} $\vect{\alpha}^{\textbf{in}}\in\mathbb{R}^{K}_{> 1}$, $\alpha_1^{\textbf{out}}, \alpha_2^{\textbf{out}}\in\mathbb{R}_{> 1}$.
        \STATE \textbf{OOD Dataset}: Generate OOD dataset $\mathcal{D}^{\textbf{o}}$ with ~\eqref{eq:ood-dataset-gen}.
		\STATE Estimate $\vect{c}$ with ground truth labels in $\mathcal{D}^s$.
        \STATE \textbf{OSLS estimation:}
        \begin{enumerate}
            \item \textbf{Source $\rho_s$ Retrieval}: Obtain $\hat{\rho}_s$ in ~\eqref{eq:rho-s-estimator} with $\hat{\mu}_0$ in ~\eqref{eq:est-rho-s-sigma}, $\hat{\mu}_0^*$ in ~\eqref{eq:sigma-0-hat}.
            \item \textbf{EM algorithm Estimation}: Obtain $\hat{\rho}_t, \hat{\vect{\pi}}$ in terms of MLE/MAP with Alg.~\ref{alg:MAPOLS}.
            \item \textbf{Target $\rho_t$ Correction}: Obtain corresponding $\hat{\rho}_t^{*}$ via ~\eqref{eq:rho-correction}.
        \end{enumerate}
        \STATE \textbf{OSLS correction:} Obtain $g(x)$ with ~\eqref{eq:lsc} based on the estimates $\hat{\rho}_t^{*}, \hat{\vect{\pi}}$.
	\end{algorithmic}  
\end{algorithm}  
\setlength{\textfloatsep}{\textfloatsepsave}
\section{Experiments}
\label{sec:experiments}

\subsection{Experimental Setups}
\textbf{Datasets}: Following the experimental setup in the OOD detection literature~\cite{zhang2023openood}, we evaluate our model with CIFAR10, CIFAR100~\cite{CIFAR} and ImageNet~\cite{imagenet} dataset as ID datasets and with SVHN~\cite{svhn}, Places~\cite{Places}, OpenImage-O~\cite{openimageo}, NINCO~\cite{ninco}, subset of TinyImageNet~\cite{tinyimagenet}, subset of iNaturalist~\cite{inaturalist}, subset of Species (SSB)~\cite{ssb} datasets as OOD datasets. The OOD datasets are split into near OOD and far OOD groups depending on their similarity to the ID dataset. Details of the dataset setup are provided in Tab.~\ref{tab:dataset-setup}, further details are available in Appendix.~\ref{Asubsec:dataset-details}.

\begin{table}[ht]
    \centering
    \footnotesize
    \begin{tabular}{c|l |c } \hline\hline
       ID dataset  &  \multicolumn{2}{c}{OOD dataset} \\ \hline
        \multirow{2}{*}{CIFAR10}  & Near &CIFAR100, TinyImageNet \\
        & Far & MNIST, SVHN, Texture, Places,\\ \hline
        \multirow{2}{*}{CIFAR100} & Near & CIFAR10, TinyImageNet, \\
         & Far & MNIST, SVHN, Texture, Places\\ \hline
        \multirow{2}{*}{ImageNet-200} & Near & SSB, NINCO, \\
        & Far & iNaturalist, Texture, OpenImage-O \\
         \hline\hline
    \end{tabular}
    \vspace{-1em}
    \caption{\textbf{Dataset setup in our experiment.} For each ID dataset, different OOD datasets are tested to justify the performance our our OSLS estimation and correction model.}
    \vspace{-1em}
    \label{tab:dataset-setup}
\end{table}

Our model is tested with different types of label shift, including the Dirichlet shift and the Ordered Long-Tailed (LT) shift commonly used in closed set label shift literature~\cite{BBSE,alexandari2020maximum}. The Dirichlet shift adjust ground truth $\vect{\pi}$ by sampling from a Dirichlet distribution with parameter $\alpha$ and the Ordered LT shift adjust $\vect{\pi}$ based on a Long-Tailed distribution with different imbalance factor and ``Forward" or ``Backward" order~\cite{mapls}. Under the open set setting, we also sub-sample the OOD datasets so that the test datasets have different OOD over ID ratios ($r={(1-\rho_t)/\rho_t}$). Details can be found in Tab.~\ref{tab:labelshift-setup}.

\begin{table}[ht]
    \centering
    \footnotesize
    \begin{tabular}{c|c c } \hline\hline
        Label Shift & Shift Parameters & OOD/ID data ratio $r$\\ \hline
        Dirichlet & $\alpha=1.0, 10.0$ & $r=1, 0.1, 0.01$ \\ \hline
        \multirow{2}{*}{Ordered LT} & $100, 50, 10$ &  \multirow{2}{*}{$r=1, 0.1, 0.01$} \\
        & ``Forward/Backward" & \\
        \hline\hline
    \end{tabular}
    \vspace{-1em}
    \caption{\textbf{Types of label shift in our experiment,} including Dirichlet shift with different shift parameter $\alpha$ and Ordered Long-Tailed (LT) shift with different imbalance factors under forward and backward order.
    }
    \label{tab:labelshift-setup}
\end{table}

\textbf{Model Setup}: The Neural Network ID classifiers are implemented using PyTorch~\cite{pytorch}. Following the convention in the OOD literature~\cite{zhang2023openood}, we train a ResNet18~\cite{RESNET} on CIFAR10/100 and ImageNet-200 datasets as multi-class ID classifiers. We test our model with different OOD classifiers, including OpenMax~\cite{openmax}, Ash~\cite{ash}, MLS~\cite{mls}, ReAct~\cite{sun2021react} and KNN~\cite{knn}, with the implementations provided by the OpenOOD project~\cite{zhang2023openood}. These ID/OOD classifiers are obtained without ground truth OOD samples available and therefore the OSLS problem setup (Definition~\ref{def:osls-problem}) is satisfied. The output of these classifiers are re-scaled to $[0,1]$ range to satisfy the requirement of our model $h:\mathcal{X}\rightarrow[0,1]$. Details are available in Appendix~\ref{Asubsec:id-classifier-details},\ref{Asubsec:ood-classifier-details}.

In the estimation model, We follow the MAPLS~\cite{mapls} setup to initialize the EM algorithms MLE/MAP-OLS with $\vect{\pi}=\vect{c}$ and $\rho_s=\rho_t$ and run for $100$ iterations to ensure convergence. We also provide estimation performance of SOTA closed set label shift estimation models BBSE~\cite{BBSE}, RLLS~\cite{RLLS}, MLLS~\cite{saerens2002adjusting} and MAPLS~\cite{mapls}. More details are given in Appendix.~\ref{Asubsec:csls}, \ref{Asubsec:em}.

\textbf{Evaluation Metrics}:
We evaluate our model mainly on the label shift estimation error $(w-\hat{w})^2/K$~\cite{BBSE} over ID classes. The label shift estimation error is the MSE between the ground truth target over source ID label distribution ratio $w=\vect{\pi}/\vect{c}$ and $\hat{w}$ is the one that was obtained with the estimator of $\vect{\pi}$. The comparison of ground truth ID data ratio $\rho_t$ and estimate $\hat{\rho}_t^{*}$ are also provided. We also test our model in terms of the Top1 Accuracy, where the results are provided in the Appendix~\ref{Asec:detailed-acc} due to the space limit.

\textbf{Reproducibility and Code Release}: To ensure the reproducibility of our model, the detailed experimental and hyperparameter setup of the ID classifier $f$ and the ID/OOD classifier $h$ follows the OpenOOD project publicly available at \url{https://github.com/Jingkang50/OpenOOD} (details in Appendix~\ref{Asubsec:id-classifier-details} and \ref{Asubsec:ood-classifier-details}). Our code is also publicly available with link in the abstract. 

\subsection{State-of-the-art Comparison}
We report performance of our OSLS-MAP model. As the open set label shift problem has been studied only recently, we mainly compare the performance of our model with state-of-the-art (SOTA) closed set label shift estimation models MLLS~\cite{saerens2002adjusting}, BBSE~\cite{BBSE}, RLLS~\cite{RLLS}, MAPLS~\cite{mapls}, and a baseline model. 
The baseline model considers the situation when no OSLS estimation model is available. In this case, it is natural to assume the target domain has an uniform ID label distribution $\vect{\pi}=\vect{1}/K$ (used in the closed set label shift model~\cite{mapls}) and same amount of ID/OOD data $r=1$ (used in the OOD detection model~\cite{meinke2022provably}).
The model proposed by \citealt{garg2022domain} is not compared because they adopt a domain adaptation approach and requires retraining the OOD and ID classifier for each experiment setup, which is time consuming especially in large scale datasets like ImageNet-200. Further, they have not report their performance on the estimation error $(w-\hat{w})^2/K$. More experimental results are provided in Appendix~\ref{Asec:detailed-wmse},~\ref{Asec:detailed-acc}.

\begin{table}[]
    \centering
    \footnotesize
    \begin{tabular}{c|c c } \hline\hline
        Dataset &  LT shift & Dirichlet shift \\ \hline
        CIFAR10 &  100\% (36/36) & 83.3\% (10/12)\\ 
        CIFAR100 & 91.7\% (33/36) & 83.3\% (10/12)\\
        ImageNet-200 & 100\% (36/36) & 83.3\% (10/12) \\
         \hline\hline
    \end{tabular}
    \vspace{-1em}
    \caption{\textbf{OSLS estimation error performance summary.} Percentage of the OSLS experiment settings (Tab.~\ref{tab:labelshift-setup}) that we are ahead of the baseline and all closed set estimation methods. }
    \label{tab:wmse-summary}
\end{table}

\setlength\tabcolsep{0.8pt} 
\begin{table*}[ht]
    \centering
    \tiny
    \renewcommand{\arraystretch}{1.1}
    \begin{tabular}{c | c | c||c c c | c c c | c c c | c c c } \hline\hline
    \multicolumn{3}{c||}{Dataset} & \multicolumn{6}{c|}{CIFAR100} & \multicolumn{6}{c}{ImageNet-200} \\ \hline
       \multicolumn{3}{c||}{ID label Shift param}  & \multicolumn{3}{c|}{LT10 Forward} & \multicolumn{3}{c|}{LT100 Forward} & \multicolumn{3}{c|}{LT10 Forward} & \multicolumn{3}{c}{LT100 Forward} \\ \hline
      \multicolumn{3}{c||}{OOD label shift param $r$} & $1.0$ & $0.1$ & $0.01$  & $1.0$ & $0.1$ & $0.01$  & $1.0$ & $0.1$ & $0.01$  & $1.0$ & $0.1$ & $0.01$ \\ \hline \hline

\multicolumn{15}{c}{Closed Set Label Shift estimation models} \\ \hline\hline 
\multicolumn{2}{c|}{\multirow{2}{*}{BBSE}}  & Near & $0.529_{\scaleto{\pm 0.017}{3pt}}$ & $0.131_{\scaleto{\pm 0.030}{3pt}}$ & $0.097_{\scaleto{\pm 0.021}{3pt}}$ & $0.850_{\scaleto{\pm 0.052}{3pt}}$ & $0.173_{\scaleto{\pm 0.041}{3pt}}$ & $0.167_{\scaleto{\pm 0.054}{3pt}}$ & $0.564_{\scaleto{\pm 0.014}{3pt}}$ & $0.119_{\scaleto{\pm 0.015}{3pt}}$ & $0.107_{\scaleto{\pm 0.012}{3pt}}$ & $0.735_{\scaleto{\pm 0.040}{3pt}}$ & $0.132_{\scaleto{\pm 0.009}{3pt}}$ & $0.112_{\scaleto{\pm 0.017}{3pt}}$\\ 

\multicolumn{2}{c|}{} & Far &  $4.118_{\scaleto{\pm 0.263}{3pt}}$ & $0.250_{\scaleto{\pm 0.037}{3pt}}$ & $0.099_{\scaleto{\pm 0.021}{3pt}}$ & $4.489_{\scaleto{\pm 0.238}{3pt}}$ & $0.294_{\scaleto{\pm 0.039}{3pt}}$ & $0.168_{\scaleto{\pm 0.053}{3pt}}$ &  $1.148_{\scaleto{\pm 0.042}{3pt}}$ & $0.134_{\scaleto{\pm 0.018}{3pt}}$ & $0.108_{\scaleto{\pm 0.012}{3pt}}$ & $1.389_{\scaleto{\pm 0.039}{3pt}}$ & $0.146_{\scaleto{\pm 0.012}{3pt}}$ & $0.112_{\scaleto{\pm 0.016}{3pt}}$\\ \hline 

\multicolumn{2}{c|}{\multirow{2}{*}{MLLS}}  & Near & $0.870_{\scaleto{\pm 0.069}{3pt}}$ & $0.116_{\scaleto{\pm 0.019}{3pt}}$ & $0.080_{\scaleto{\pm 0.022}{3pt}}$ & $1.100_{\scaleto{\pm 0.098}{3pt}}$ & $0.132_{\scaleto{\pm 0.034}{3pt}}$ & $0.113_{\scaleto{\pm 0.039}{3pt}}$ & $1.152_{\scaleto{\pm 0.101}{3pt}}$ & $0.131_{\scaleto{\pm 0.019}{3pt}}$ & $0.099_{\scaleto{\pm 0.017}{3pt}}$ & $1.272_{\scaleto{\pm 0.128}{3pt}}$ & $0.146_{\scaleto{\pm 0.025}{3pt}}$ & $0.116_{\scaleto{\pm 0.027}{3pt}}$\\ 

\multicolumn{2}{c|}{} & Far &  $9.656_{\scaleto{\pm 1.747}{3pt}}$ & $0.328_{\scaleto{\pm 0.034}{3pt}}$ & $0.083_{\scaleto{\pm 0.023}{3pt}}$ & $9.862_{\scaleto{\pm 1.469}{3pt}}$ & $0.353_{\scaleto{\pm 0.044}{3pt}}$ & $0.117_{\scaleto{\pm 0.040}{3pt}}$ &  $4.095_{\scaleto{\pm 0.078}{3pt}}$ & $0.167_{\scaleto{\pm 0.031}{3pt}}$ & $0.101_{\scaleto{\pm 0.017}{3pt}}$ & $4.436_{\scaleto{\pm 0.263}{3pt}}$ & $0.189_{\scaleto{\pm 0.031}{3pt}}$ & $0.117_{\scaleto{\pm 0.026}{3pt}}$\\ \hline

\multicolumn{2}{c|}{\multirow{2}{*}{RLLS}}  & Near & $0.426_{\scaleto{\pm 0.000}{3pt}}$ & $0.425_{\scaleto{\pm 0.000}{3pt}}$ & $0.425_{\scaleto{\pm 0.000}{3pt}}$ & $1.404_{\scaleto{\pm 0.000}{3pt}}$ & $1.402_{\scaleto{\pm 0.000}{3pt}}$ & $1.402_{\scaleto{\pm 0.000}{3pt}}$ & $0.432_{\scaleto{\pm 0.000}{3pt}}$ & $0.432_{\scaleto{\pm 0.000}{3pt}}$ & $0.432_{\scaleto{\pm 0.000}{3pt}}$ & $1.397_{\scaleto{\pm 0.000}{3pt}}$ & $1.396_{\scaleto{\pm 0.000}{3pt}}$ & $1.396_{\scaleto{\pm 0.000}{3pt}}$\\ 

\multicolumn{2}{c|}{} & Far &  $0.426_{\scaleto{\pm 0.000}{3pt}}$ & $0.425_{\scaleto{\pm 0.000}{3pt}}$ & $0.425_{\scaleto{\pm 0.000}{3pt}}$ & $1.404_{\scaleto{\pm 0.000}{3pt}}$ & $1.403_{\scaleto{\pm 0.000}{3pt}}$ & $1.402_{\scaleto{\pm 0.000}{3pt}}$ &  $0.433_{\scaleto{\pm 0.000}{3pt}}$ & $0.432_{\scaleto{\pm 0.000}{3pt}}$ & $0.432_{\scaleto{\pm 0.000}{3pt}}$ & $1.397_{\scaleto{\pm 0.000}{3pt}}$ & $1.396_{\scaleto{\pm 0.000}{3pt}}$ & $1.396_{\scaleto{\pm 0.000}{3pt}}$\\ \hline 

\multicolumn{2}{c|}{\multirow{2}{*}{MAPLS}}  & Near & $0.672_{\scaleto{\pm 0.040}{3pt}}$ & $0.116_{\scaleto{\pm 0.014}{3pt}}$ & $0.085_{\scaleto{\pm 0.014}{3pt}}$ & $0.965_{\scaleto{\pm 0.060}{3pt}}$ & $0.164_{\scaleto{\pm 0.024}{3pt}}$ & $0.134_{\scaleto{\pm 0.026}{3pt}}$ & $0.877_{\scaleto{\pm 0.069}{3pt}}$ & $0.114_{\scaleto{\pm 0.016}{3pt}}$ & $0.085_{\scaleto{\pm 0.014}{3pt}}$ & $1.046_{\scaleto{\pm 0.094}{3pt}}$ & $0.134_{\scaleto{\pm 0.021}{3pt}}$ & $0.095_{\scaleto{\pm 0.020}{3pt}}$\\ 

\multicolumn{2}{c|}{} & Far &  $7.481_{\scaleto{\pm 1.351}{3pt}}$ & $0.275_{\scaleto{\pm 0.024}{3pt}}$ & $0.087_{\scaleto{\pm 0.014}{3pt}}$ & $7.763_{\scaleto{\pm 1.137}{3pt}}$ & $0.336_{\scaleto{\pm 0.027}{3pt}}$ & $0.137_{\scaleto{\pm 0.027}{3pt}}$ &  $3.004_{\scaleto{\pm 0.055}{3pt}}$ & $0.139_{\scaleto{\pm 0.024}{3pt}}$ & $0.086_{\scaleto{\pm 0.014}{3pt}}$ & $3.328_{\scaleto{\pm 0.168}{3pt}}$ & $0.164_{\scaleto{\pm 0.025}{3pt}}$ & $0.097_{\scaleto{\pm 0.020}{3pt}}$\\ \hline 
\hline 

\multicolumn{15}{c}{Open Set Label Shift estimation models} \\ \hline\hline 

\multicolumn{3}{c||}{Baseline} & $0.426_{\scaleto{\pm 0.000}{3pt}}$ & $0.426_{\scaleto{\pm 0.000}{3pt}}$ & $0.426_{\scaleto{\pm 0.000}{3pt}}$ & $1.405_{\scaleto{\pm 0.000}{3pt}}$ & $1.405_{\scaleto{\pm 0.000}{3pt}}$ & $1.405_{\scaleto{\pm 0.000}{3pt}}$ & $0.436_{\scaleto{\pm 0.000}{3pt}}$ & $0.436_{\scaleto{\pm 0.000}{3pt}}$ & $0.436_{\scaleto{\pm 0.000}{3pt}}$ & $1.405_{\scaleto{\pm 0.000}{3pt}}$ & $1.405_{\scaleto{\pm 0.000}{3pt}}$ & $1.405_{\scaleto{\pm 0.000}{3pt}}$\\ \hline 

\multirow{10}{*}{\textbf{ours}} & \multirow{2}{*}{OpenMax}  &  Near &  \cc $\mathbf{0.387}_{\scaleto{\pm 0.030}{3pt}}$ & \cc $\mathbf{0.043}_{\scaleto{\pm 0.006}{3pt}}$ & \cc $\mathbf{0.046}_{\scaleto{\pm 0.009}{3pt}}$ & \cc $\mathbf{0.511}_{\scaleto{\pm 0.038}{3pt}}$ & \cc $\mathbf{0.081}_{\scaleto{\pm 0.012}{3pt}}$ & \cc $\mathbf{0.077}_{\scaleto{\pm 0.005}{3pt}}$ &  $0.699_{\scaleto{\pm 0.010}{3pt}}$ & \cc $\mathbf{0.035}_{\scaleto{\pm 0.002}{3pt}}$ & \cc $\mathbf{0.022}_{\scaleto{\pm 0.002}{3pt}}$ & \cc $0.820_{\scaleto{\pm 0.009}{3pt}}$ & \cc $\mathbf{0.035}_{\scaleto{\pm 0.002}{3pt}}$ & \cc $\mathbf{0.019}_{\scaleto{\pm 0.000}{3pt}}$\\ 

& & Far & $2.223_{\scaleto{\pm 0.233}{3pt}}$ & \cc $\mathbf{0.087}_{\scaleto{\pm 0.002}{3pt}}$ & \cc $\mathbf{0.046}_{\scaleto{\pm 0.010}{3pt}}$ & $2.341_{\scaleto{\pm 0.498}{3pt}}$ & \cc $\mathbf{0.118}_{\scaleto{\pm 0.015}{3pt}}$ & \cc $\mathbf{0.078}_{\scaleto{\pm 0.005}{3pt}}$ & $2.500_{\scaleto{\pm 0.153}{3pt}}$ & \cc $\mathbf{0.048}_{\scaleto{\pm 0.002}{3pt}}$ & \cc $\mathbf{0.022}_{\scaleto{\pm 0.003}{3pt}}$ & $2.739_{\scaleto{\pm 0.114}{3pt}}$ & \cc $\mathbf{0.047}_{\scaleto{\pm 0.001}{3pt}}$ & \cc $\mathbf{0.019}_{\scaleto{\pm 0.000}{3pt}}$\\ \cline{2-15}

& \multirow{2}{*}{MLS}  & Near & \cc $\mathbf{0.323}_{\scaleto{\pm 0.020}{3pt}}$ & \cc $\mathbf{0.078}_{\scaleto{\pm 0.004}{3pt}}$ & \cc $\mathbf{0.078}_{\scaleto{\pm 0.005}{3pt}}$ & \cc $\mathbf{0.415}_{\scaleto{\pm 0.017}{3pt}}$ & \cc $\mathbf{0.126}_{\scaleto{\pm 0.014}{3pt}}$ & \cc $\mathbf{0.120}_{\scaleto{\pm 0.013}{3pt}}$ & \cc $\mathbf{0.194}_{\scaleto{\pm 0.011}{3pt}}$ & \cc $\mathbf{0.069}_{\scaleto{\pm 0.005}{3pt}}$ & \cc $\mathbf{0.069}_{\scaleto{\pm 0.012}{3pt}}$ & \cc $\mathbf{0.217}_{\scaleto{\pm 0.002}{3pt}}$ & \cc $\mathbf{0.079}_{\scaleto{\pm 0.009}{3pt}}$ & \cc $\mathbf{0.081}_{\scaleto{\pm 0.006}{3pt}}$\\ 

& & Far & $1.289_{\scaleto{\pm 0.337}{3pt}}$ & \cc $\mathbf{0.099}_{\scaleto{\pm 0.008}{3pt}}$ & \cc $\mathbf{0.079}_{\scaleto{\pm 0.006}{3pt}}$ & \cc $\mathbf{1.366}_{\scaleto{\pm 0.313}{3pt}}$ & \cc $\mathbf{0.150}_{\scaleto{\pm 0.015}{3pt}}$ & \cc $\mathbf{0.120}_{\scaleto{\pm 0.013}{3pt}}$ & \cc $\mathbf{0.118}_{\scaleto{\pm 0.024}{3pt}}$ & \cc $\mathbf{0.069}_{\scaleto{\pm 0.005}{3pt}}$ & \cc $\mathbf{0.069}_{\scaleto{\pm 0.012}{3pt}}$ & \cc $\mathbf{0.126}_{\scaleto{\pm 0.007}{3pt}}$ & \cc $\mathbf{0.081}_{\scaleto{\pm 0.009}{3pt}}$ & \cc $\mathbf{0.082}_{\scaleto{\pm 0.006}{3pt}}$\\ \cline{2-15}

& \multirow{2}{*}{ReAct}  &  Near & \cc $\mathbf{0.331}_{\scaleto{\pm 0.030}{3pt}}$ & \cc $\mathbf{0.076}_{\scaleto{\pm 0.006}{3pt}}$ & \cc $\mathbf{0.075}_{\scaleto{\pm 0.005}{3pt}}$ & \cc $\mathbf{0.396}_{\scaleto{\pm 0.023}{3pt}}$ & \cc $\mathbf{0.110}_{\scaleto{\pm 0.012}{3pt}}$ & \cc $0.124_{\scaleto{\pm 0.010}{3pt}}$ & \cc $\mathbf{0.251}_{\scaleto{\pm 0.050}{3pt}}$ & \cc $\mathbf{0.092}_{\scaleto{\pm 0.028}{3pt}}$ & \cc $0.094_{\scaleto{\pm 0.022}{3pt}}$ & \cc $\mathbf{0.279}_{\scaleto{\pm 0.052}{3pt}}$ & \cc $\mathbf{0.110}_{\scaleto{\pm 0.030}{3pt}}$ & \cc $0.111_{\scaleto{\pm 0.026}{3pt}}$\\ 

& & Far & $1.138_{\scaleto{\pm 0.251}{3pt}}$ & \cc $\mathbf{0.096}_{\scaleto{\pm 0.004}{3pt}}$ & \cc $\mathbf{0.075}_{\scaleto{\pm 0.005}{3pt}}$ & \cc $\mathbf{1.202}_{\scaleto{\pm 0.281}{3pt}}$ & \cc $\mathbf{0.127}_{\scaleto{\pm 0.008}{3pt}}$ & \cc $0.124_{\scaleto{\pm 0.010}{3pt}}$ & \cc $\mathbf{0.103}_{\scaleto{\pm 0.015}{3pt}}$ & \cc $\mathbf{0.092}_{\scaleto{\pm 0.026}{3pt}}$ & \cc $0.094_{\scaleto{\pm 0.022}{3pt}}$ & \cc $\mathbf{0.128}_{\scaleto{\pm 0.022}{3pt}}$ & \cc $\mathbf{0.113}_{\scaleto{\pm 0.031}{3pt}}$ & \cc $0.112_{\scaleto{\pm 0.027}{3pt}}$\\ \cline{2-15}

& \multirow{2}{*}{KNN}  &  Near &  $0.736_{\scaleto{\pm 0.026}{3pt}}$ & \cc $0.141_{\scaleto{\pm 0.008}{3pt}}$ & \cc $0.139_{\scaleto{\pm 0.002}{3pt}}$ & \cc $\mathbf{0.817}_{\scaleto{\pm 0.034}{3pt}}$ & \cc $0.235_{\scaleto{\pm 0.018}{3pt}}$ & \cc $0.228_{\scaleto{\pm 0.023}{3pt}}$ &  \cc $\mathbf{0.309}_{\scaleto{\pm 0.008}{3pt}}$ & \cc $0.116_{\scaleto{\pm 0.007}{3pt}}$ & \cc $0.115_{\scaleto{\pm 0.005}{3pt}}$ & \cc $\mathbf{0.325}_{\scaleto{\pm 0.010}{3pt}}$ & \cc $0.133_{\scaleto{\pm 0.010}{3pt}}$ & \cc $0.127_{\scaleto{\pm 0.004}{3pt}}$\\ 

& & Far & $1.188_{\scaleto{\pm 0.173}{3pt}}$ & \cc $\mathbf{0.152}_{\scaleto{\pm 0.006}{3pt}}$ & \cc $0.140_{\scaleto{\pm 0.002}{3pt}}$ & \cc $\mathbf{1.287}_{\scaleto{\pm 0.126}{3pt}}$ & \cc $\mathbf{0.246}_{\scaleto{\pm 0.016}{3pt}}$ & \cc $0.229_{\scaleto{\pm 0.023}{3pt}}$ & \cc $\mathbf{0.158}_{\scaleto{\pm 0.013}{3pt}}$ & \cc $\mathbf{0.115}_{\scaleto{\pm 0.007}{3pt}}$ & \cc $0.115_{\scaleto{\pm 0.005}{3pt}}$ & \cc $\mathbf{0.167}_{\scaleto{\pm 0.012}{3pt}}$ & \cc $\mathbf{0.132}_{\scaleto{\pm 0.010}{3pt}}$ & \cc $0.127_{\scaleto{\pm 0.003}{3pt}}$\\ \cline{2-15}

& \multirow{2}{*}{Ash}  &  Near & \cc $\mathbf{0.358}_{\scaleto{\pm 0.050}{3pt}}$ & \cc $\mathbf{0.111}_{\scaleto{\pm 0.005}{3pt}}$ & \cc $0.101_{\scaleto{\pm 0.015}{3pt}}$ & \cc $\mathbf{0.541}_{\scaleto{\pm 0.051}{3pt}}$ & \cc $0.217_{\scaleto{\pm 0.023}{3pt}}$ & \cc $0.198_{\scaleto{\pm 0.035}{3pt}}$ & \cc $\mathbf{0.262}_{\scaleto{\pm 0.026}{3pt}}$ & \cc $\mathbf{0.110}_{\scaleto{\pm 0.016}{3pt}}$ & \cc $0.108_{\scaleto{\pm 0.013}{3pt}}$ & \cc $\mathbf{0.299}_{\scaleto{\pm 0.028}{3pt}}$ & \cc $\mathbf{0.132}_{\scaleto{\pm 0.015}{3pt}}$ & \cc $0.135_{\scaleto{\pm 0.018}{3pt}}$\\ 

& & Far & $1.015_{\scaleto{\pm 0.121}{3pt}}$ & \cc $\mathbf{0.119}_{\scaleto{\pm 0.002}{3pt}}$ & \cc $0.101_{\scaleto{\pm 0.014}{3pt}}$ & \cc $\mathbf{1.183}_{\scaleto{\pm 0.120}{3pt}}$ & \cc $\mathbf{0.229}_{\scaleto{\pm 0.021}{3pt}}$ & \cc $0.198_{\scaleto{\pm 0.035}{3pt}}$ & \cc $\mathbf{0.110}_{\scaleto{\pm 0.009}{3pt}}$ & \cc $\mathbf{0.110}_{\scaleto{\pm 0.016}{3pt}}$ & \cc $0.108_{\scaleto{\pm 0.013}{3pt}}$ & \cc $\mathbf{0.140}_{\scaleto{\pm 0.015}{3pt}}$ & \cc $\mathbf{0.133}_{\scaleto{\pm 0.014}{3pt}}$ & \cc $0.136_{\scaleto{\pm 0.018}{3pt}}$\\

\hline
    \end{tabular}
    \vspace{-1em}
    \caption{\textbf{Estimation Error $(w-\hat{w})^2/K(\downarrow)$ of our OSLS estimation and correction model on CIFAR100 and ImageNet-200 dataset with Near OOD datasets and Far OOD datasets comparison under different ID and OOD label shift. Outperforming results are in bold face and settings that outperform the baseline are colored in gray.} Our model outperforms baseline under most label shift settings. Each metric is averaged among corresponding OOD test set (Tab.~\ref{tab:dataset-setup}) and over three independent ID classifiers.}
    \vspace{-1em}
    \label{tab:wmse-cifar100}
\end{table*}
\setlength\tabcolsep{6pt} 

\begin{figure*}
    \centering
    \includegraphics[width=0.9\linewidth]{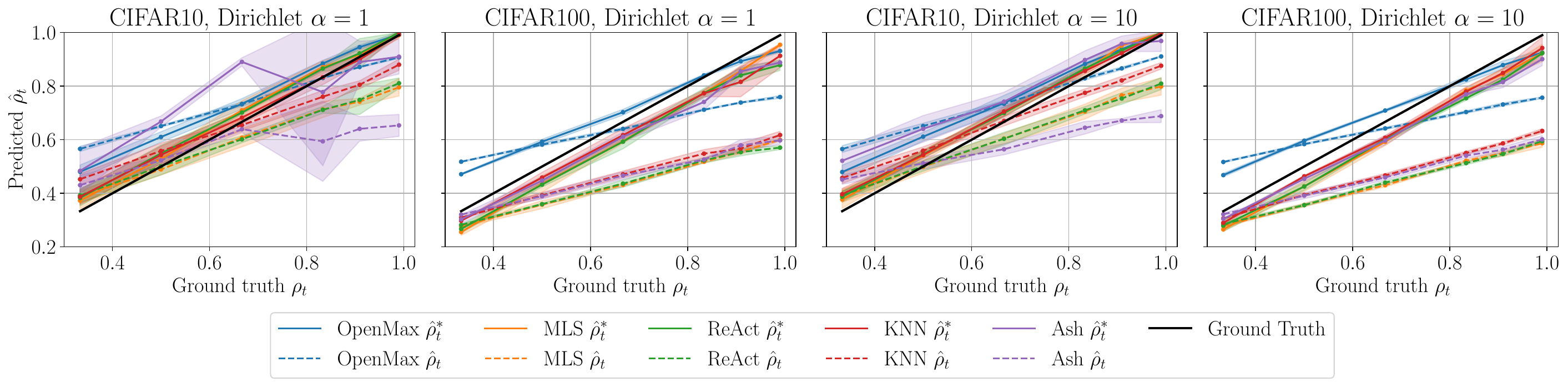}
    \vspace{-1em}
    \caption{\textbf{Estimation result comparison of $\hat{\rho}_t^{*}$ by our model (Solid lines), $\hat{\rho_t}$ by our model but without $\rho_t$ correction (\S~\ref{subsec:rho-t-correction}) (Dashed lines) based on different OOD classifiers and the Ground truth $\rho_t$ (Black, Solid line), on CIFAR10/100 dataset with Dirichlet shift and Near OOD dataset (Tab.~\ref{tab:dataset-setup}).} The estimation result exhibit a linear correlation with the ground truth, which is explained by our analysis in Theorem~\ref{theorem:estrho-linear} 
    Moreover, our $\rho_t$ correction model is able to adjust the predicted $\hat{\rho}_t$ to $\hat{\rho}_t^{*}$ that is closer to the ground truth. Shaded area are $\pm$ one standard deviation over three independent ID classifiers.}
    \vspace{-1em}
    \label{fig:rho-t-ablation}
\end{figure*}



\textbf{Estimation Error}:
As seen in Tab.~\ref{tab:wmse-cifar100}, our estimation model effectively estimate ID label shift in the open set settings on CIFAR100, ImageNet-200 datasets and outperform the open set base line in most of the settings. Moreover, although the closed set models performance increases when target domain has less OOD sample (small $r$), our model take OOD data into account and still outperform all existing SOTA closed set models in the reported Open set settings. Similar to the Top1 Accuracy result, in terms of the estimation error, OpenMax fits better with our Assumption~\ref{assume:A2}B and thus also performs the best among the OOD classifiers in most of the OSLS settings in the table.. 

\textbf{Target $\rho_t$ Estimation}:
Fig.~\ref{fig:rho-t-ablation} justifies our $\rho_t$ correction model (\S~\ref{subsec:rho-t-correction}) with results on CIFAR10/100 dataset under Dirichlet ID shift. As seen in the figure, most estimate $\hat{\rho}_t^*$ of our model matches better with the ground truth $\rho_t$ than $\hat{\rho}_t$ obtained without our $\rho_t$ correction model. Such result implies that the tested OOD classifiers roughly satisfies the requirement of Theorem~\ref{theorem:estrho-linear} in \eqref{eq:rho-t-correction-condition}. This is probably because these OOD classifiers are usually designed based on ID classifier's maximal output (e.g. max logit) and such output tends to be identical among ID classes when source domain ID dataset that the ID classifier trained on is class-uniform. More visualization can be found in Appendix~\ref{Asec:more-ablation}.

\subsection{Ablation Study}\label{subsec:ablation}
\textbf{Assumption Analysis}: Assumption~\ref{assume:A2}A is a common assumption for Neural Network classifiers, which has been used in previous label shift estimation problem~\cite{mapls} and other classification tasks like calibration~\cite{odin} and Long-Tailed Recognition~\cite{xu2021towards}. We justify Assumption~\ref{assume:A2}A for the practical classifiers used in our OSLS estimation and correction model with empirical evidences.

As discussed in the previous works~\cite{garg2020unified}, if Assumption~\ref{assume:A2}A is satisfied, classifier $f$ is a perfectly calibrated classifier on the source domain. The calibration performance of the classifier is commonly evaluated via the Expected Calibration Error (ECE)~\cite{guo2017calibration, liu2023model, liu2024self}, where a well calibrated classifier will have ECE close to 0. In this work, we provide the calibration performance of the practical classifiers $f$ that are used in our model.

\begin{table}[]
    \centering
    \footnotesize
    \begin{tabular}{c|c c c } \hline\hline
        Dataset & Classifier 1  & Classifier 2  & Classifier 3  \\ \hline
        CIFAR10 & 0.0277 & 0.0281 & 0.0242 \\
        CIFAR100 & 0.0628 & 0.0600 & 0.0621 \\
        ImageNet-200 & 0.0163 & 0.0133 & 0.0131 \\
        \hline\hline
    \end{tabular}
    \vspace{-1em}
    \caption{\textbf{Calibration Performance in terms of ECE ($\downarrow$) of the ID classifiers} (3 for each dataset) used in our model on the source domain validation set. The classifiers have good calibration performance, providing evidence that Assumption~\ref{assume:A2} is a useful assumption for building practical models.
    }
    \label{tab:ece}
\end{table}
As shown in Tab.~\ref{tab:ece}, the classifiers that we used in our model have good calibration performance. Hence Assumption~\ref{assume:A2}A is likely to be satisfied, which justifies the practical applicability of our model in the real world problems.

\textbf{Training Time}: The training time of our OSLS \emph{estimation} model on CIFAR10/100 dataset is less than 1 second and on ImageNet-200 dataset is less than 5 seconds. Since EM algorithm is scalable to large scale datasets~\cite{mapls}, our model can be easily applied to real world problems.

\section{Conclusion}

In this work, we analyze the problem of Open Set Label Shift and propose a model to estimate target label distribution of ID and OOD class and then adapt a source domain ID/OOD classifier to target domain without retraining. With reasonable assumptions and an OOD reference dataset, our overall estimate of the target label distribution is built on three estimates: 1) an estimate of source label distribution of ID/OOD class, 2) an estimate of target label distribution of ID/OOD class and 3) an estimate of target label distribution of OOD class when some assumptions on the OOD classifier is not satisfied. The source domain classifier is then adapted to the target domain based on the estimation results. We show that the requirement of an OOD reference dataset in our model can be relaxed and pseudo OOD samples generated from the ID samples can be used instead. Experiments on benchmark image datasets CIFAR10/100 and ImageNet-200 with different OSLS settings demonstrate the effectiveness of our model.

\clearpage

{
    \small
    \bibliographystyle{ieeenat_fullname}
    \bibliography{ref}
}


\clearpage
\appendix
\onecolumn

\etocdepthtag.toc{mtappendix}
\etocsettagdepth{mtchapter}{none}
\etocsettagdepth{mtappendix}{subsection}
\tableofcontents
\clearpage

\section{List of Symbols}
\adjust{
This section summarizes the list of symbols used in this paper:
\begin{table}[H]
    \centering
    \begin{tabular}{c l} \hline\hline
        Notations & \multicolumn{1}{c}{Explanation} \\ \hline
        \addlinespace
        $\mathbb{R}^d_{>1}$ & $d$-dimensional real value space with all values larger than $1$\\
        \addlinespace
        $\mathcal{X}\subseteq \mathbb{R}^d$ & Data space as subsset of the $d$-dimensional real value set\\ 
        \addlinespace
        $\mathcal{Y}=\{1,2,...,K\}$ & Label space of $K$ ID classes\\
        \addlinespace
        $\mathcal{Y}\cup\{K+1\}$ & Label space of $K$ ID classes and one $K+1$ OOD class\\
        \addlinespace
        $X_s, Y_s, B_s$ & Random Variable of the source domain image ($X_s$), label ($Y_s$) and ID data indicator ($B_s$)\\
        \addlinespace
        $X_t, Y_t, B_t$ & Random Variable of the target domain image ($X_t$), label ($Y_t$) and ID data indicator ($B_t$)\\
        \addlinespace
        $p_s(\cdot)$ & Source domain distribution (\eg $p_s(x)$ for $X_s$, $p_s(b=\cdot)$ for $B_s$)\\
        \addlinespace
        $p_t(\cdot)$ & Tource domain distribution (\eg $p_t(x)$ for $X_t$, $p_t(y=\cdot)$ for $Y_t$)\\
        \addlinespace
        $\mathcal{D}^s=\{(x^s_i,y^s_i)\}^{N_s}_{i=1}$ & Source domain labelled dataset with $N_s$ samples \\
        \addlinespace
        $\mathcal{D}^t=\{x^t_i\}^{N_t}_{i=1}$ & Target domain unlabelled dataset with $N_t$ samples \\
        \addlinespace
        $\mathcal{D}^{\textbf{o}}=\{x^o_i\}^{N_o}_{i=1}$ & OOD reference dataset with $N_o$ samples \\
        \addlinespace 
        $\vect{c}, \rho_s$ & Source domain ID label distribution $p_s(y=\cdot)=\vect{c}$ and ID data ratio $p_s(b=1)=\rho_s$ \\
        \addlinespace
        $\vect{\pi}, \rho_t$ & Target domain ID label distribution $p_t(y=\cdot)=\vect{\pi}$ and ID data ratio $p_t(b=1)=\rho_t$ \\
        \addlinespace
        $f:\mathcal{X}\rightarrow\Delta^{K-1}$ & Source domain ID classifier that output $K$ dimensional probability simplex\\
        \addlinespace
        $h:\mathcal{X}\rightarrow[0,1]$ & Source domain ID vs OOD classifier that output a scalar in $[0,1]$ \\
        \addlinespace
        $\gamma, T$ & Hyperparameters of our model when use Gaussian noise to generate pseudo OOD samples \\
        \addlinespace
        $|\mathcal{D}|$ & Returns cardinality of the dataset $\mathcal{D}$, \ie $|\mathcal{D}|=N$ if $\mathcal{D}=\{x_i\}^N_{i=1}$ \\
        \addlinespace
         \hline\hline
    \end{tabular}
    \label{Atab:los}
\end{table}
}

\clearpage

\section{Related Works}\label{Asec:related-works}

\subsection{MLLS} 
The Maximum Likelihood Label Shift method is a closed set label shift estimation model that was originally proposed by \citealt{saerens2002adjusting}. With unlabeled target domain data $\mathcal{D}^t=\{x^t_i\}^{N^t}_{i=1}$,  MLLS estimates the target label distribution $p_t(y=\cdot)=\vect{\pi}$ by maximizing the log likelihood: 
\[
\begin{aligned}
    \log L(\vect{\pi}; \mathcal{D}^t) := \log \left(\prod^{N^t}_{i=1}\sum^K_{j=1}\frac{\pi_j}{c_j}f(x_i)_j\right)
\end{aligned}
\]
using the EM algorithm, which is stated in Alg.~\ref{Aalg:MLLS}.
\setlength{\textfloatsepsave}{\textfloatsep} 
\setlength{\textfloatsep}{0pt}
\begin{algorithm}[ht]
\caption{MLLS}
	\begin{algorithmic}
	    \label{Aalg:MLLS}
	    \STATE \textbf{Input: } $\mathcal{D}^t=\{x^t_i\}^{N^t}_{i=1}, p_s(y=\cdot)=\vect{c}, f:\mathcal{X}\rightarrow \Delta^{K-1}$.
	    \STATE \textbf{Initialize: } $\vect{\pi}^{(0)}\in \Delta^{K-1}$.
		\FOR{$m=0$ to $M$}
        \STATE \textbf{E-step:} Evaluate
        \STATE\[
                g_{ij}^{m} = \frac{\frac{\pi_j}{c_j}f(x^t_i)_j}{\sum^K_{l=1}\frac{\pi_l}{c_l}f(x^t_i)_l}.
              \]
        \STATE  \textbf{M-step:} Evaluate
        \STATE \[
                        \pi_j^{(m+1)} = \frac{1}{N^t}\sum^{N^t}_{i=1} g_{ij}^m.
               \]
		\ENDFOR
		\STATE \textbf{Output: } $p_t(y=\cdot)=\vect{\pi}^{(M+1)}$.
	\end{algorithmic}  
\end{algorithm}  
\setlength{\textfloatsep}{\textfloatsepsave}

The iterative procedure of the EM algorithm is repeated until numerical convergence to obtain the MLE of the target label distribution $\vect{\pi}^{\text{MLE}}$, which satisfies:
\[
\vect{\pi}^{\text{MLE}} \in \underset{\vect{\pi}\in\Delta^{K-1}}{\arg\min} -\log L(\vect{\pi}; \mathcal{D}^t).
\]

In the following works, \citealt{alexandari2020maximum} proves that the NLL objective of MLLS is convex and empirically demonstrates that MLLS outperform other closed set label shift estimation methods in many image classification datasets. \citealt{garg2020unified} proved that MLLS is consistent when classifier $f$ is canonically calibrated.

\subsection{MAPLS}

The Maximum {\it a Posteriori} Label Shift (MAPLS) method is also a closed set label shift estimation model, which was recently proposed by \citealt{mapls}. By introducing a Dirichlet prior $\vect{\pi}\sim\text{Dir}(K,\vect{\alpha})$ over the target label distribution $\vect{\pi}$, MAPLS aims at optimizing the posterior:
\[
    p(\vect{\pi}|\mathcal{D}^t,\vect{\alpha}) = \frac{1}{Z} \prod^{K}_{i=1}\pi_i^{\alpha_i - 1}\prod^{N^t}_{i=1} \sum^K_{j=1}\frac{\pi_j}{c_j}f(x_i)_j,
\]
where $Z$ is the normalization constant. The EM algorithm of MAPLS can be written as Alg.~\ref{Aalg:MAPLS}.

\setlength{\textfloatsepsave}{\textfloatsep} 
\setlength{\textfloatsep}{0pt}
\begin{algorithm}[ht]
\caption{MAPLS}
	\begin{algorithmic}
	    \label{Aalg:MAPLS}
	    \STATE \textbf{Input: } $\mathcal{D}^t=\{x^t_i\}^{N^t}_{i=1}$, $p_s(y=\cdot)=\vect{c}$, $f:\mathcal{X}\rightarrow\Delta^{K-1}$, $\vect{\alpha}\in\mathbb{R}^K_{>1}$.
	    \STATE \textbf{Initialize: } $\vect{\pi}^{(0)}\in \Delta^{K-1}_{>0}$.
		\FOR{$m=0$ to $M$}
        \STATE \textbf{E-step} Evaluate:
        \STATE\begin{equation}
                g_{ij}^m = \frac{\frac{\pi^{(m)}_j}{c_j}f(x_i)_j}{\sum^K_{l=1}\frac{\pi^{(m)}_l}{c_l}f(x_i)_l}.
            \end{equation}
        \STATE  \textbf{M-step} Obtain $\vect{\pi}^{(m+1)}$ with:
        \STATE \begin{equation}
                        \pi_j^{(m+1)} = \frac{\sum^{N^t}_{i=1} g_{ij}^m + \alpha_j - 1}{N^t + \sum^K_{l=1}(\alpha_l - 1)}.
                \end{equation}
		\ENDFOR
		\STATE \textbf{Output: } $p_t(y=\cdot)=\vect{\pi}^{(M+1)}$.
	\end{algorithmic}  
\end{algorithm}  
\setlength{\textfloatsep}{\textfloatsepsave}

\citealt{mapls} further proved that the optimization objective of MAPLS algorithm is strictly convex and EM algorithm is guaranteed to converge to the MAP estimate $\vect{\pi}^{\text{MAP}}$ which satisfies:
\[
    \vect{\pi}^{\text{MAP}} = \underset{\vect{\pi}\in\Delta^{K-1}}{\arg\min} - \frac{1}{Z} \prod^{K}_{i=1}\pi_i^{\alpha_i - 1}\prod^{N^t}_{i=1} \sum^K_{j=1}\frac{\pi_j}{c_j}f(x_i)_j.
\]
The author of the MAPLS algorithm also empirically demonstrates that MAPLS outperforms other closed set label shift estimation models in large scale datasets like ImageNet, especially under large label shift settings.

\subsection{OOD detection} 
OOD detection has been widely studied in the Deep Learning regime. 
Existing approaches can be categorized into roughly three categories: post-hoc inference methods and training methods with or without OOD data.

The majority of the OOD methods are post-hoc inference methods, where the OOD classifier is constructed based on a pre-trained classifier over ID classes.  
OpenMax~\cite{openmax} proposed to construct the OOD classifier by modelling per-class features with a Weibull distribution. 
MSP~\cite{msp} utilize the maximal softmax score of the ID classifier prediction. 
ODIN~\cite{odin} observed that NN models respond to ID and OOD data differently under adversarial attacks~\cite{goodfellow2014explaining}. 
MDS~\cite{mds} also adopts the adversarial attack approach but detects OOD data with a Mahalanobis distance-based score. 
OpenGAN~\cite{opengan} trains an extra discriminator network to distinguish ID and OOD features. 
EBO~\cite{ebo} proposed an Energy-based score to detect OOD samples. 
GRAM~\cite{gram} establish their model with Gram matrices. 
ReAct~\cite{sun2021react} demonstrates that rectifying the penultimate layer features of the pre-trained classifier can help post-hoc OOD detection methods. 
MLS~\cite{mls} argues that maximal logit score is a better OOD indicator. 
VIM~\cite{vim} propose a three stage pipline to compute the OOD score by adjusting the features, logits and softmax probability of the ID classifier. 
\citealt{knn} introduces a k-Nearest Neighbor (KNN) based OOD classifier. 
Ash~\cite{ash} shows that pruning image features in the intermediate layers can help OOD detection.

Among training methods without OOD data, \citealt{RotPred} argues that training the classifier with an auxiliary self-supervised rotation loss is beneficial to OOD detection models. 
GODIN~\cite{godin} extends the ODIN model by introducing an extra linear layer that models the probability the data is not OOD given the image. 
CSI~\cite{csi} enhances a baseline OOD detector by training the classifier with a loss that contrasts ground truth samples with distribution shifted samples. 
APRL~\cite{aprl} encourages ID samples to move far away from a bounded space left for OOD data.

In the machine learning community, \citealt{vaze2021open,miller2021accuracy} argue that a good ID classifier implies a good OOD classifier. 
\citealt{hein2019relu} shows that for OOD sample, a ReLU network can predict its label as ID class with arbitrary high confidence. 
\citealt{meinke2019towards} propose a GMM based classifier approach to prevent the model from assigning OOD data with high confidence. 
\citealt{fang2022out} analyzes the conditions under which OOD detection is learnable.

\adjust{
\subsection{Open Set Domain Adaptation}
Compared with the Open Set Label Shift problem, the Open Set Domain Adaptation (OSDA) task considers a slightly different setup. Theoretically, the OSDA task does not require the Label Shift Assumption~\ref{assume:A1} to hold between the source and target domain. Empirically, OSDA models focus more on tackling the image distribution $p(x)$ shift rather than the label distribution $p(y)$ shift, where they are usually tested with source and target domain having identical ID label distribution~\cite{panareda2017open}. 

Similar to the relation between Closed Set Label Shift and Open Set Label Shift, the OSDA task extends the Closed Set Domain Adaptation (CSDA) task by allowing target domain having an extra class that contains all the new categories that not appear in the source domain~\cite{panareda2017open,saito2018open, liu2019separate, fang2020open, zhang2021learning,wang2023learning}. As the OSDA problem setup and the OSDA models are not used in this paper, we will not go into details of the OSDA problem. Therefore, we would like to refer readers who are interested in the OSDA problem to the cited literature for detailed discussions.
}

\clearpage

\onecolumn

\section{Mathematical Proofs}\label{Asec:proofs}

\subsection{Proof of Theorem~\ref{theorem:est-rho} (See page~\pageref{theorem:est-rho})}

\estimaterho*
\begin{proof}
Given the available information, for $p_s(b=1) = \rho_s$ we have:
\begin{equation}
\begin{aligned}
    p_s(b=1) & = \mathbb{E}_{X_s}[p(b=1|x)] = \mathbb{E}_{X_s}[h(x)] = \mathbb{E}_{B_s}[\mathbb{E}_{X_s|B_s}[h(x)]] \\
    & = (1 - p_s(b=1)) \cdot \mathbb{E}_{X_s|B_s=0}[h(x)] + p_s(b=1) \cdot \mathbb{E}_{X_s|B_s=1}[h(x)] 
\end{aligned}
\end{equation}
Rearranging the equation and we can get:
\begin{equation}
\begin{aligned}
    \rho_s & = \frac{\mathbb{E}_{X_s|B_s=0}[h(x)]}{1 - \mathbb{E}_{X_s|B_s=1}[h(x)] + \mathbb{E}_{X_s|B_s=0}[h(x)]}  \\
    & = \frac{\mu_0}{1 - \mu_1 + \mu_0},
\end{aligned}
\end{equation}
where $\mu_0:=\mathbb{E}_{X_s|B_s=0}[h(x)]$ and $\mu_1:= \mathbb{E}_{X_s|B_s=1}[h(x)]$.

The expectation terms can be approximated given OOD dataset $\mathcal{D}^\textbf{o}$ and source ID dataset $\mathcal{D}^s$:
\begin{equation}
\left\{
\begin{aligned}
    \mathbb{E}_{X_s|B_s=0}[h(x)] & \approx \frac{1}{\vert\mathcal{D}^{\textbf{o}}\vert}\sum_{x\in\mathcal{D}^{\textbf{o}}}h(x) \\
     \mathbb{E}_{X_s|B_s=1}[h(x)] & \approx \frac{1}{\vert\mathcal{D}^s\vert}\sum_{x\in\mathcal{D}^s}h(x), \\
\end{aligned}
\right.
\end{equation}
which yields the approximation $\hat{\rho}$:
\begin{equation}
    \hat{\rho} = \frac{\hat{\mu}_0}{1 - \hat{\mu}_1 + \hat{\mu}_0},
\end{equation}
where $\hat{\mu}_0 := \frac{1}{\vert\mathcal{D}^{\textbf{o}}\vert}\sum_{x\in\mathcal{D}^{\textbf{o}}}h(x)$ and $\hat{\mu}_1 :=  \frac{1}{\vert\mathcal{D}^s\vert}\sum_{x\in\mathcal{D}^s}h(x)$.

Note that since $h(x)\in[0,1]$, Hoeffding's inequality~\cite{vershynin2018high} guarantees for all $ \epsilon> 0$:
\begin{equation}\label{Aeq:binary-hoeffding}
\begin{aligned}
    p\left(\vert\mu_0  - \hat{\mu}_0\vert \geq \epsilon\right) & \leq 2e^{-2|\mathcal{D}^{\textbf{o}}|\epsilon^2} \\
    p\left(\vert\mu_1  - \hat{\mu}_1\vert \geq \epsilon\right) & \leq 2e^{-2|\mathcal{D}^s|\epsilon^2}.
\end{aligned}
\end{equation}

Therefore with high probability of at least $1 - 2e^{-2\min(|\mathcal{D}^{\textbf{o}}|,|\mathcal{D}^s|)\epsilon^2}$ we have:
\begin{equation}\label{Aeq:binary-bound}
\left\{
\begin{aligned}
      \rho - \hat{\rho}  \leq & \frac{\mu_0}{1 - \mu_1 + \mu_0}  - \frac{\mu_0 +\epsilon}{1 - (\mu_1 +\epsilon) + (\mu_0 +\epsilon)} =  \frac{\epsilon}{1 - \mu_1 + \mu_0} , \\
      \rho - \hat{\rho} \geq & \frac{\mu_0}{1 - \mu_1 + \mu_0}  - \frac{\mu_0 -\epsilon}{1 - (\mu_1 -\epsilon) + (\mu_0 -\epsilon)} =  \frac{-\epsilon}{1 - \mu_1 + \mu_0} , 
\end{aligned}
\right.
\end{equation}
for all $\delta\in[0, \max((1-\mu_0)/2, (1-\mu_1)/2)]$, which is equivalent to:
\begin{equation}\label{Aeq:}
    \vert \rho - \hat{\rho} \vert <  \frac{\epsilon}{1 - \mu_1 + \mu_0}.
\end{equation}

Letting $\delta:=e^{-2\min(|\mathcal{D}^{\textbf{o}}|,|\mathcal{D}^s|)\epsilon^2}$, rearrange the equations and we get the result.
\end{proof}

\subsection{Extension of Theorem~\ref{theorem:est-rho} to the Multi-Class setting (See page~\pageref{theorem:est-rho})}\label{Asubsec:est-rho-extension}
\textbf{Problem Setup: (General)} 
Given a blackbox model $h:\mathcal{X}\rightarrow\Delta^{K-1}$ that satisfies $h(x) = p(y|x)$ for distribution $p(x,y)$ and $K$ datasets $\mathcal{D}^1, \mathcal{D}^2, ...,\mathcal{D}^K$, with $\mathcal{D}^k$ containing samples drawn i.i.d. from $p(x|y=k)$, we want to estimate the label distribution $p(y=\cdot)=\vect{\rho}\in\Delta^{K-1}$. 

Similar to the binary case, we can write the label distribution as the sum of the conditional expectation:
\begin{equation}
\begin{aligned}
    \rho_j & = \mathbb{E}_{X}[p(y=j|x)] = \mathbb{E}_{X}[h(x)_j] = \mathbb{E}_{Y}[\mathbb{E}_{X|Y}[h(x)_j]] \\
    & = \sum^K_{k=1}p(y=k) \mathbb{E}_{X|Y}[h(x)_j] = (\mu \vect{\rho})_j, 
\end{aligned}
\end{equation}
where $\mu\in\mathbb{R}^{K\times K}$ with $\mu_{jk}:=\mathbb{E}_{X|Y=k}[h(x)_j]$.

\begin{restatable}{lemma}{estimatec}
    \label{lemma:est-c}
    (\textbf{Multi-Class}) If $p(y|x)=h(x)$ , given $\mathcal{D}^1, \mathcal{D}^2,...,\mathcal{D}^K$ containing samples $x$ drawn i.i.d.~from ${p(x|y=1)}, {p(x|y=2)},...,{p(x|y=K)}$, then for $p(y)=\vect{\rho}\in\Delta^{K-1}$ we have:
    \begin{equation}
         \underset{\vect{\rho}\in\Delta^{K-1}}{\arg\min} \Vert (\hat{\mu}-\mathbf{I})\vect{\rho}\Vert^2_2 \underset{a.s.}{\longrightarrow} \vect{\rho},
    \end{equation}
    where $\hat{\mu}\in\mathbb{R}^{K\times K}$ is a stochastic matrix with $\mu_{jk} = \frac{1}{|\mathcal{D}^k|}\sum_{x\in\mathcal{D}^k}h(x)_j$.
\end{restatable}


\begin{proof}
  Given the available information, let $p(y=j) = \rho_j$ for all $ j\in\mathcal{Y}=\{1,2,...,K\}$, then we have:
\begin{equation}
\begin{aligned}
    \rho_j & = \mathbb{E}_{X}[p(y=j|x)] = \mathbb{E}_{X}[h(x)_j] = \mathbb{E}_{Y}[\mathbb{E}_{X|Y}[h(x)_j]] \\
    & = \sum^K_{k=1}p(y=k) \mathbb{E}_{X|Y}[h(x)_j]\\
    & = (\mu \vect{\rho})_j, 
\end{aligned}
\end{equation}
where $\mu\in\mathbb{R}^{K\times K}$ with $\mu_{jk}:=\mathbb{E}_{X|Y=k}[h(x)_j]$.

The $\mu$ can be approximated via:
\begin{equation}
\begin{aligned}
    \mu_{jk} \approx \hat{\mu}_{jk}:= \frac{1}{|\mathcal{D}^k|}\sum_{x\in\mathcal{D}^k}h(x)_j.
\end{aligned}
\end{equation}

Hence we can approximate $\vect{\rho}$ with $\hat{\vect{\rho}}$ that is defined as:
\begin{equation}
\begin{aligned}
    \hat{\vect{\rho}} := \underset{\vect{\rho}\in\Delta^{K-1}}{\arg\min} \Vert (\hat{\mu}-\mathbf{I})\vect{\rho}\Vert^2_2. 
\end{aligned}
\end{equation}
\end{proof}

\subsection{Proof of Lemma~\ref{lemma:obj} (See page~\pageref{lemma:obj})}\label{Asubsec:nll-mlls}

\lemmaobj*

\begin{proof}
The label shift assumption can be written as:
\begin{equation}\label{Aeq:ls-assumption-ood}
\begin{aligned}
    p_s(x|y=i) & = p_t(x|y=i) \quad \text{for all} \quad i\in\mathcal{Y}\cup\{K+1\}\\
\end{aligned}
\end{equation}

On target domain, if we are given only unlabeled images $\mathcal{D}^t=\{x^t_i\}^{N^t}_{i=1}$, we can construct the likelihood:
\begin{equation}
\begin{aligned}
    L(\vect{\pi},\rho_t;\mathcal{D}^t)  = & \prod^{N^t}_{i=1} p_t(x;\vect{\pi}, \rho_t) = \prod^{N^t}_{i=1}  \left(\sum^2_{l=1}\sum^K_{j=1}p_t(x_i|y=j) p_t(y=j|b=l) p_t(b=l) \right).
\end{aligned}
\end{equation}

Note that $p_s(b=1)=\rho_s$, $p_t(b=1)=\rho_t$ and in Eq.~\ref{eq:label-distribution} for all $(x,j)\in\mathcal{X}\times(\mathcal{Y}\cup\{K+1\})$ we have:
\begin{equation}\label{Aeq:label-distribution}
\begin{aligned}
& p_s(y|b;\vect{c})=
\begin{cases}
  c_j, & \text{if } b=1, y\neq K+1\\
  1, & \text{if } b=0, y=K+1 \\
  0, & \text{otherwise}\\
\end{cases},
\quad 
& p_s(y|b;\vect{c})=
\begin{cases}
  \pi_j, & \text{if } b=1, y\neq K+1\\
  1, & \text{if } b=0, y=K+1 \\
  0, & \text{otherwise}\\
\end{cases}.
\end{aligned}
\end{equation}

Based on Eq.~\eqref{Aeq:label-distribution} and label shift Assumption~\ref{assume:A1} we have:
\begin{equation}\label{Aeq:nll}
\begin{aligned}
    L(\vect{\pi},\rho_t;\mathcal{D}^t) = & \prod^{N^t}_{i=1}  \left(\sum^2_{l=1}\sum^K_{j=1}p_t(x_i|y=j) p_t(y=j|b=l) p_t(b=l) \right) \\
    = &  \prod^{N^t}_{i=1}  \left(\sum^K_{j=1}p_t(x_i|y=j) p_t(y=j|b=1) p_t(b=1) +  p_t(x_i|y=K+1)p_t(y=K+1|b=0)p_t(b=0)\right) \\
    = &  \prod^{N^t}_{i=1}  \left(\sum^K_{j=1}p_s(x_i|y=j) p_t(y=j|b=1) p_t(b=1) +  p_s(x_i|y=K+1)p_t(y=K+1|b=0)p_t(b=0)\right) \\
    = & \prod^{N^t}_{i=1} \left(\sum^K_{j=1}\frac{p_s(y=j|x_i) }{p_s(y=j)}p_t(y=j|b=1) p_t(b=1) +  \frac{p_s(y=K+1|x_i)}{p_s(y=K+1)}p_t(y=K+1|b=0)p_t(b=0)\right) \\
    & \cdot Const,
\end{aligned}
\end{equation}
where $Const:=\prod^{N^t}_{i=1} p_s(x_i)$ is irrelevant to $\vect{\pi}$ or $\rho_t$.

Based on Eq.~\eqref{Aeq:label-distribution} and Assumption~\ref{assume:A2}: $p_s(y=\cdot|x,b=1)=f(x)$ and $p_s(b=1|x)=h(x)$ we have:
\begin{equation}
\begin{aligned}
    1 - h(x) & = p_s(b=0|x_i) \\
    & = \sum^{K+1}_{i=1} p_s(b=0|y=i) p(y=i|x_i) \\
    & =  \sum^{K+1}_{i=1} \frac{p_s(y=i|b=0)p_s(b=0)}{p_s(y=i)} p(y=i|x_i) \\
    & = \frac{1 \cdot p_s(b=0)}{\sum^2_{j=1}p_s(y=K+1|b=j)p_s(b=j)} p_s(y=K+1|x_i) \\
    & = \frac{p_s(b=0)}{p_s(b=0)} \cdot p_s(y=K+1|x_i) = p_s(y=K+1|x_i),
\end{aligned}
\end{equation}
and for $j\in\{1,2,...,K\}$ we have:
\begin{equation}
\begin{aligned}
    p_s(y=j|x_i) & = p_s(y=j|x_i,b=1) p_s(b=1|x_i) + p_s(y=j|x_i,b=0) p_s(b=0|x_i)\\
    & = f(x)_j \cdot \rho_s + 0 \cdot (1 - h(x)) = h(x) \cdot f(x_i)_j.
\end{aligned}
\end{equation}

Marginalize Eq.~\eqref{Aeq:label-distribution} we can also get:
\begin{equation}\label{Aeq:ls-equalities2}
\begin{aligned}
    p_s(y=j) = 
    \begin{cases}
        c_j \cdot \rho_s, & j \neq K+1 \\
        1 - \rho_s, & j = K+1 
    \end{cases},
    \quad 
    p_t(y=j) = 
    \begin{cases}
        \pi_j \cdot \rho_t, & j \neq K+1 \\
        1 - \rho_t, & j = K+1 
    \end{cases},
\end{aligned}
\end{equation}

Substituting Eq.~\eqref{Aeq:label-distribution} and Eq.~\eqref{Aeq:ls-equalities2} into the likelihood Eq.~\eqref{Aeq:nll} we get:
\begin{equation}\label{Aeq:nll-orign}
\begin{aligned}
    L(\vect{\pi},\rho_t;\mathcal{D}^t) 
    & = \prod^{N^t}_{i=1} \left(\sum^K_{j=1}\frac{h(x) \cdot f(x_i)_j }{\rho_s \cdot c_j}\pi_j \cdot \rho_t +  \frac{1 - h(x)}{1 - \rho_s}\cdot 1 \cdot (1 - \rho_t)\right) \cdot Const, \\
    & = \prod^{N^t}_{i=1} \left(\frac{\rho_t}{\rho_s}h(x_i) \cdot \sum^K_{j=1}\frac{\pi_j}{c_j} f(x_i)_j +  \frac{1 - \rho_t}{1 - \rho_s} \cdot (1- h(x_i))\right) \cdot Const. \\
\end{aligned}
\end{equation}

Further substitute Eq.~\eqref{eq:tilde-f} and Eq.~\eqref{eq:tilde-pi-and-c} into Eq.~\eqref{Aeq:nll-orign} and then we can get the result.

\end{proof}

\subsection{Proof of Theorem~\ref{theorem:em-mle} (See page~\pageref{theorem:em-mle})}
\propemmle*

\setlength{\textfloatsepsave}{\textfloatsep} 
\setlength{\textfloatsep}{0pt}
\begin{algorithm}[H]
\caption{MLE-OLS}\label{Aalg:MLE-OLS}
	\begin{algorithmic}
	    \STATE \textbf{Input: } $\mathcal{D}^t_f=\{x^t_i\}^{N^t}_{i=1}, \vect{c}, \rho_s, h(x), f(x)$.
	    \STATE \textbf{Initialize: } $\vect{\pi}^{(0)}\in\Delta^{K-1}_{>0}, \rho^{(0)}_t\in (0, 1)$
		\FOR{$m=0$ to $M$}
        \STATE \textbf{Construct: } $\Tilde{\vect{\pi}}^{(m)}$ based on $\vect{\pi}^{(m)},\rho^{(m)}_t$ and Eq.~\eqref{eq:tilde-pi-and-c}.
        \STATE \textbf{E-step:} For $j\in(\mathcal{Y}\cup\{K+1\})$, evaluate
        \STATE\begin{equation}
                g_{ij}^{(m)} = \frac{\Tilde{\pi}^{(m)}_j/ \Tilde{c}_j \cdot \Tilde{f}(x^t_i)_j}{\sum^{K+1}_{l=1}\Tilde{\pi}^{(m)}_l/ \Tilde{c}_l \cdot \Tilde{f}(x^t_i)_l}.
            \end{equation}
        \STATE  \textbf{M-step:}  For $j\in\mathcal{Y}$, evaluate 
        \STATE \begin{equation}
                \left\{
                \begin{aligned}  
                    \pi_j^{(m+1)} & = \frac{\sum^{N^t}_{i=1} g_{ij}^{(m)}}{N^t - \sum^{N^t}_{i=1}g_{iK+1}^{(m)}} \\
                    \rho_t^{(m+1)} &  = \frac{N^t - \sum^{N^t}_{i=1}g_{iK+1}^{(m)} }{N}
                \end{aligned}
                \right.
                \end{equation}
		\ENDFOR
		\STATE \textbf{Output: } $p_t(y=\cdot)=\vect{\pi}^{(M+1)}, p_t(b=1)=\rho_t^{(M+1)}$.
	\end{algorithmic}  
\end{algorithm}  
\setlength{\textfloatsep}{\textfloatsepsave}

\begin{proof}
\textbf{Convexity}:
As shown in Lemma~\ref{lemma:obj} Eq.~\eqref{Aeq:nll-orign}, the negative log likelihood of $\vect{\pi},\rho_t$ given Assumption~\ref{assume:A1}, \ref{assume:A2} and unlabeled target domain dataset $\mathcal{D}^t$ can be written as:
\begin{equation}
\begin{aligned}
    -\log L(\vect{\pi},\rho_t;\mathcal{D}^t) 
    & = -\sum^{N^t}_{i=1}\log \left(\frac{\rho_t}{\rho_s}h(x_i) \cdot \sum^K_{j=1}\frac{\pi_j}{c_j} f(x_i)_j +  \frac{1 - \rho_t}{1 - \rho_s} \cdot (1- h(x_i))\right) + C. \\
\end{aligned}
\end{equation}

As a function of $\rho_t$, the NLL can be rewritten as:
\begin{equation}
    -\log L(\vect{\pi},\rho_t;\mathcal{D}^t)  = -\sum^{N^t}_{i=1}\log (A\rho_t + B) + C,
\end{equation}
which is a convex function w.r.t.~$\rho_t$.

As a function of $\vect{\pi}$, the NLL can be rewritten as:
\begin{equation}
    -\log L(\vect{\pi},\rho_t;\mathcal{D}^t) =-\sum^{N^t}_{i=1}\log \left(A \sum^K_{j=1}\frac{\pi_j}{c_j} f(x_i)_j + B\right) + C,
\end{equation}
which is a convex function w.r.t $\vect{\pi}$.

Moreover, same as the close world setting~\cite{alexandari2020maximum}, the NLL is convex in the reparameterisation of $\Tilde{c}_t$.
\end{proof}

\begin{proof}
\textbf{EM algorithm}:
The NLL objective of MLE defined in Lemma~\ref{lemma:obj}, Eq.~\eqref{eq:nll} can be rewritten as:
\begin{equation}\label{Aeq:new-nll}
\begin{aligned}
    - \log L(\vect{\pi},\rho_t;\mathcal{D}^t) 
    & = - \sum^{N^t}_{i=1} \log \left(\sum^K_{j=1}\frac{\Tilde{\pi}_j}{\Tilde{c}_j} \Tilde{f}(x_i)_j \right) + C, \\
\end{aligned}
\end{equation}
which is reparametrised as the objective of the closed set label shift estimation model MLLS~\cite{saerens2002adjusting} algorithm (Appendix.~\ref{Asec:related-works}, Alg.~\ref{Aalg:MLLS}). 

As MLE is invariant under reparametrisation~\cite{murphy2012machine}, and MLLS has been proved to converge to a MLE estimate \cite{alexandari2020maximum}, thus EM algorithm~\ref{Aalg:MLE-OLS} converges to a $\Tilde{c}_t^{\text{MLE}}$ and will also converge to a $\vect{\pi}^{\text{MLE}},\rho_t^{\text{MLE}}$.

The MLE can be seen as a special case of MAP estimate with prior distribution being $1$. In this case, by setting  
$\vect{\alpha}^{\textbf{in}}=\vect{1}, \alpha_1^{\textbf{out}}=1,\alpha_2^{\textbf{out}})=1$. Proof of EM algorithm for MAP estimate can be found in Proof of Proposition~\ref{prop:em-map}.
\end{proof}

\subsection{MAP estimation of target label distribution parameters}\label{Asubsec:map}
\textbf{MAP estimate:} Moreover, if we employ a prior ${\vect{\pi}\sim p(\vect{\pi}|\vect{\alpha}^{\textbf{in}})}$ over the target label distribution $\vect{\pi}$, or a prior $\rho_t\sim p(\rho_t|\vect{\alpha}^{\textbf{out}})$ over the target ID data ratio $\rho_t$, we can construct the posterior of $\vect{\pi}$ and $\rho_t$ as:
\begin{equation}\label{eq:posterior}
\begin{aligned}
    - \log p(\vect{\pi},\rho_t|\mathcal{D}^t,\vect{\alpha})  = & - \log L(\vect{\pi},\rho_t;\mathcal{D}^t) - \log p(\vect{\pi}|\vect{\alpha}^{\textbf{in}})  \\
    & - \log p(\rho_t|\vect{\alpha}^{\textbf{out}}) + C.
\end{aligned}
\end{equation}

In this work, inspired by \citealt{mapls}, we show that with a Dirichlet prior over ${\vect{\pi}\sim\text{Dir}(K,\vect{\alpha}^{\textbf{in}})}$ or a Beta prior over $\rho_t\sim\text{Beta}(\alpha_1^{\textbf{out}}, \alpha_2^{\textbf{out}})$,  the MAP estimate $\Tilde{\vect{\pi}}^{\text{MAP}}$ can be obtained via another EM algorithm over the objective:
\begin{equation} \label{eq:MAP-obj}
\vect{\pi}^{\text{MAP}},\rho_t^{\text{MAP}} \in \underset{\Tilde{\pi}\in\Delta^K}{\arg\min} - \log p(\vect{\pi},\rho_t|\mathcal{D}^t,\vect{\alpha}),
\end{equation}
where the details are also provided in Proposition~\ref{prop:em-map}.

\begin{restatable}{proposition}{propemmap}\textbf{(MAP)}
\label{prop:em-map}
    Under Assumption~\ref{assume:A1},~\ref{assume:A2}, if ${\vect{\pi}\sim\text{Dir}(K,\vect{\alpha}^{\textbf{in}})}$ with $\vect{\alpha}^{\textbf{in}}\in\mathbb{R}^{K}_{>1}$ and $\rho_t\sim\text{Beta}(\alpha_1^{\textbf{out}}, \alpha_2^{\textbf{out}})$ with $\alpha_1^{\textbf{out}}, \alpha_2^{\textbf{out}}\in\mathbb{R}_{>1}$, then 
    \begin{itemize}
        \item The posterior in Eq.~\eqref{eq:posterior} is strictly convex in $\vect{\pi}$ and strictly convex in $\rho_t$.
        \item EM algorithm~\ref{alg:MAPOLS} converge to the $\Tilde{\vect{\pi}}^{\text{MAP}}$ in Eq.~\eqref{eq:MAP-obj}.
    \end{itemize}
\end{restatable}


\setlength{\textfloatsepsave}{\textfloatsep} 
\setlength{\textfloatsep}{0pt}
\begin{algorithm}[ht]
\caption{MAP-OLS}
	\begin{algorithmic}
	    \STATE \textbf{Input: }$\mathcal{D}^t_f=\{x^t_i\}^{N^t}_{i=1}, \vect{c}, \rho_s, h(x), f(x)$, $\vect{\alpha}^{\textbf{in}}, \alpha_1^{\textbf{out}}, \alpha_2^{\textbf{out}}$.
        \STATE \textbf{Require:} $\vect{\alpha}^{\textbf{in}}\in\mathbb{R}^{K}_{> 1}$, $\alpha_1^{\textbf{out}},\alpha_2^{\textbf{out}}\in\mathbb{R}_{> 1}$.
	    \STATE \textbf{Initialize:} $\vect{\pi}^{(0)}\in\Delta^{K-1}_{>0}, \rho^{(0)}_t\in (0, 1)$.
        \STATE \textbf{Construct:} $\Tilde{f}$ based on Eq.~\eqref{eq:tilde-f}.
		\FOR{$m=0$ to $M$}
        \STATE \textbf{Construct: } $\Tilde{\vect{\pi}}^{(m)}$ based on $\vect{\pi}^{(m)},\rho^{(m)}_t$ and Eq.~\eqref{eq:tilde-pi-and-c}.
        \STATE \textbf{E-step:} For $j\in\mathcal{Y}\cup\{K+1\}$, evaluate
        \STATE\begin{equation}
                g_{ij}^{(m)} = \frac{\Tilde{\pi}^{(m)}_j/ \Tilde{c}_j \cdot \Tilde{f}(x^t_i)_j}{\sum^K_{l=1}\Tilde{\pi}^{(m)}_l/ \Tilde{c}_l \cdot \Tilde{f}(x^t_i)_l}.
            \end{equation}
        \STATE  \textbf{M-step:}  For $j\in\mathcal{Y}$, evaluate 
        \STATE \begin{equation}
                \left\{
                \begin{aligned}  
                    \pi_j^{(m+1)} & = \frac{\sum^{N^t}_{i=1} g_{ij}^{(m)} + \alpha_j^{\textbf{in}} - 1}{N^t - \sum^{N^t}_{i=1}g_{iK+1}^{(m)} + \sum^K_{l=1}(\alpha_l^{\textbf{in}} - 1)} \\
                    \rho_t^{(m+1)} &  = \frac{N^t - \sum^{N^t}_{i=1}g_{iK+1}^{(m)} + \alpha_1^{\textbf{out}} - 1}{N^t + \alpha_1^{\textbf{out}} + \alpha_2^{\textbf{out}} - 2}.
                \end{aligned}
                \right.
                \end{equation}
		\ENDFOR
		\STATE \textbf{Output: } $p_t(y=\cdot)=\vect{\pi}^{(M+1)}, p_t(b=1)=\rho_t^{(M+1)}$.
	\end{algorithmic}  
\end{algorithm}  
\setlength{\textfloatsep}{\textfloatsepsave}

\begin{proof}
\textbf{Convexity}: As shown in the Proof Proposition~\ref{theorem:em-mle}, the MLE objective given in Lemma~\ref{lemma:obj} is convex on $\vect{\pi},\rho_t$.

Since Dirichlet prior $\vect{\pi}\sim \text{Dir}(K,\vect{\alpha}^{\textbf{in}})$ with $\vect{\alpha}^{\textbf{in}}>\vect{1}$ is strictly convex on $\vect{\pi}$. And Beta prior $\rho_t\sim \text{Beta}(\alpha^{\textbf{out}}_1, \alpha^{\textbf{out}}_2)$ with $\alpha^{\textbf{out}}_1, \alpha^{\textbf{out}}_2 > 1$ is strictly convex on $\rho_t$, the overall posterior:
\begin{equation}
    - \log p(\vect{\pi},\rho_t|\mathcal{D}^t,\vect{\alpha})  = - \log L(\vect{\pi},\rho_t;\mathcal{D}^t) - \log p(\vect{\pi}|\vect{\alpha}^{\textbf{in}})  - \log p(\rho_t|\vect{\alpha}^{\textbf{out}}) + C
\end{equation}
is strictly convex on $\vect{\pi}$ and $\rho_t$

\end{proof}

\begin{proof}
\textbf{EM algorithm: } 

To be concise, we will use the notation:
\begin{equation}\label{Aeq:tilde-f}
\begin{aligned}
    \Tilde{f}(x)_i & = 
    \begin{cases}
    h(x)\cdot f(x)_i,  & i\in \mathcal{Y} \\
    1 - h(x), & i = K + 1,
    \end{cases} \\
    \Tilde{\vect{\pi}} & = [\rho_t\cdot\pi_1,...,\rho_t\cdot\pi_K, 1 - \rho_t]^T \\
    \Tilde{\vect{c}} & = [\rho_s\cdot c_1,...,\rho_t\cdot c_K, 1 - \rho_s]^T
\end{aligned}
\end{equation}

\textbf{Remark:} We prove the case with the model having both prior $\vect{\pi}\sim\text{Dir}(K,\vect{\alpha}^{\textbf{in}})$ and $\rho_t\sim\text{Beta}(\alpha_1^{\textbf{out}}, \alpha_2^{\textbf{out}})$, where $\vect{\alpha}^{\textbf{in}}\in\mathbb{R}^{K}_{>1}$ and $\alpha_1^{\textbf{out}}, \alpha_2^{\textbf{out}}\in\mathbb{R}_{>1}$. EM algorithms for other cases can be derived similarly by setting $\vect{\alpha}^{\textbf{in}}=\vect{1}$ or $\alpha_1^{\textbf{out}}=1, \alpha_2^{\textbf{out}}=1$ or both.

The proof consists of three stages: 
\begin{enumerate}
    \item Identify the latent variable, derive the complete posterior;
    \item Construct the $Q(\vect{\pi},\rho_t|\vect{\pi}^{(m)},\rho_t^{(m)})$ and obtain \textbf{E-Step};
    \item Optimize $Q(\vect{\pi},\rho_t|\vect{\pi}^{(m)},\rho_t^{(m)})$ w.r.t $\vect{\pi},\rho_t$ and obtain \textbf{M-Step}.
\end{enumerate}

\textbf{Step 1:} As discussed in the main paper (Eq.~\eqref{eq:nll}), we can construct the latent variable $\Tilde{Y}_s\sim\text{Cat}(K+1,\Tilde{\vect{c}})$ and $\Tilde{Y}_t\sim\text{Cat}(K+1,\Tilde{\vect{\pi}})$. With $\Tilde{Y}_t$ as latent variable, let $\Tilde{\mathbb{Y}}=\{\Tilde{y}^t_i\}^N_{i=1}$ with $\Tilde{y}^t_i\in\mathcal{Y}\cup\{K+1\}$, the complete posterior $p(\vect{\pi}|\mathcal{D}^t,\Tilde{\mathbb{Y}},\vect{\alpha}^{\textbf{in}}, \alpha_1^{\textbf{out}}, \alpha_2^{\textbf{out}})$ can be written as:
\begin{equation}
\begin{aligned}
    p(\vect{\pi},\rho_t|\mathcal{D}^t_f,\Tilde{\mathbb{Y}},\vect{\alpha}^{\textbf{in}}, \alpha_1^{\textbf{out}}, \alpha_2^{\textbf{out}}) & = \frac{1}{C} p(\vect{\pi}|\vect{\alpha}^{\textbf{in}})p(\rho_t|\alpha_1^{\textbf{out}}, \alpha_2^{\textbf{out}})\prod^{N}_{i=1} \prod^{K+1}_{j=1} p_t(x^t_i,\Tilde{y}^t_i=j;\Tilde{\vect{\pi}}) \\
    & = \frac{1}{C} p(\vect{\pi}|\vect{\alpha}^{\textbf{in}})p(\rho_t|\alpha_1^{\textbf{out}}, \alpha_2^{\textbf{out}})\prod^{N}_{i=1} \prod^{K+1}_{j=1} \frac{p_t(\Tilde{y}^t_i=j;\Tilde{\vect{\pi}})}{p_s(\Tilde{y}^t_i=j)}p_s(\Tilde{y}^t_i=j|x^t_i)  \\
    & =\frac{1}{C} \rho_t^{\alpha_1^{\textbf{out}}}(1 - \rho_t)^{\alpha_2^{\textbf{out}}}\prod^K_{l=1}\pi_l^{\alpha^{\textbf{in}}_l-1} \prod^{N}_{i=1} \prod^{K+1}_{j=1}\left(\frac{\Tilde{\pi}_j}{\Tilde{c}_j}\right)^{\mathbb{I}(\Tilde{y}^t_i=j)}\Tilde{f}(x^t_i)_j,
\end{aligned}
\end{equation}
where $C$ includes all the terms that are constant w.r.t $\vect{\pi},\rho_t$.

\textbf{Step 2:} Given the complete posterior $p(\vect{\pi}|\mathcal{D}^t_f,\Tilde{\mathbb{Y}},\vect{\alpha}^{\textbf{in}}, \alpha_1^{\textbf{out}}, \alpha_2^{\textbf{out}})$, we can construct the $Q(\vect{\pi},\rho_t|\vect{\pi}^{(m)},\rho_t^{(m)})$ in the \textbf{E-Step} as:
\begin{equation}
\begin{aligned}
Q(\vect{\pi},\rho_t|\vect{\pi}^{(m)},\rho_t^{(m)}) = & \underset{\Tilde{\mathbb{Y}}|\mathcal{D}^t,\vect{\pi}^{(m)},\rho_t^{(m)}}{\mathbb{E}}\left[\log p(\vect{\pi},\rho_t|\mathcal{D}^t_f,\Tilde{\mathbb{Y}},\vect{\alpha}^{\textbf{in}}, \alpha_1^{\textbf{out}}, \alpha_2^{\textbf{out}})\right]\\
    = & \underset{\Tilde{\mathbb{Y}}|\mathcal{D}^t,\vect{\pi}^{(m)},\rho_t^{(m)}}{\mathbb{E}}\bigg[ \sum^{N}_{i=1} \sum^{K+1}_{j=1}\mathbb{I}(\Tilde{y}^t_i=j) \log\Tilde{\pi}_j + \sum^K_{l=1} (\alpha_l-1)\log\pi_l \\
     & \quad + \alpha_1^{\textbf{out}} \cdot \log\rho_t + \alpha_2^{\textbf{out}} \cdot \log (1 - \rho_t) + C \bigg] \\
    = & \sum^{N}_{i=1} \sum^{K+1}_{j=1}p_t(\Tilde{y}^t_i=j|x^t_i;\Tilde{\vect{\pi}}^{(m)}) \log\Tilde{\pi}_j + \sum^K_{l=1} (\alpha_l-1)\log\pi_l \\
    & \quad + \alpha_1^{\textbf{out}} \cdot \log\rho_t + \alpha_2^{\textbf{out}} \cdot \log (1 - \rho_t)  + C  \\
    = & \sum^{N}_{i=1} \sum^{K+1}_{j=1}g_{ij}^{(m)}\log\Tilde{\pi}_j  + \sum^K_{l=1} (\alpha_l-1)\log\pi_l \\
    & \quad + \alpha_1^{\textbf{out}} \cdot \log\rho_t + \alpha_2^{\textbf{out}} \cdot \log (1 - \rho_t)  + C,
\end{aligned}  
\end{equation}
where the likelihood $g_{ij}^{(m)} := p_t(\Tilde{y}^t_i=j|x^t_i;\vect{\pi}^{(m)},\rho^{(m)})$ can be simply obtained via:
\begin{equation}\label{Aeq:gij-2}
\begin{aligned}
    g_{ij}^{(m)} = \frac{\frac{\Tilde{\pi}^{(m)}_j}{\Tilde{c}^{(m)}_j}\Tilde{f}(x_i)_j}{\sum^{K+1}_{l=1}\frac{\Tilde{\pi}^{(m)}_l}{\Tilde{c}^{(m)}_l}\Tilde{f}(x_i)_l} \quad \text{for all} \quad j\in\mathcal{Y}\cup\{K+1\}.
\end{aligned}
\end{equation}

\textbf{Step 3:} In the \textbf{M-step}, with available $Q(\vect{\pi},\rho_t|\vect{\pi}^{(m)},\rho_t^{(m)})$, we solve the optimization objective with respect to $\vect{\pi}$ by fixing $\rho_t$ and vise versa:
\begin{equation}
\begin{aligned}
    \vect{\pi}^{(m+1)},\rho_t^{(m+1)} & =\underset{\vect{\pi}\in\Delta^{K-1},\rho_t\in[0,1]}{\arg\max}Q(\vect{\pi},\rho_t|\vect{\pi}^{(m)},\rho_t^{(m)}) 
\end{aligned}
\end{equation}

By substitution, the objective can be rewritten as:
\begin{equation}\label{Aeq:q-obj}
\left\{
    \begin{aligned}
         \min_{\vect{\pi}} & - \sum^{N^t}_{i=1} \sum^{K+1}_{j=1} g_{ij}^{(m)} \log \Tilde{\pi}_j - \sum^K_{l=1} (\alpha^{\textbf{in}}_l-1)\log\pi_l - \alpha_1^{\textbf{out}} \cdot \log\rho_t - \alpha_2^{\textbf{out}} \cdot \log (1 - \rho_t)\\
        \text{s.t: } & \sum^K_{j=1}\pi_j = 1,  \Tilde{\vect{\pi}} = [\rho_t\cdot\pi_1,...,\rho_t\cdot\pi_K, 1 - \rho_t]^T, \\
        & \pi_i\geq 0 \text{ for } i\in[1,2,...K], \rho_t\in[0,1].
    \end{aligned}
\right.
\end{equation}

\textbf{Convexity} Eq.~\eqref{Aeq:q-obj} is just a linear combination of $\log\pi_i$, which is a concave function w.r.t.~$\vect{\pi}$. Knowing that the constraints define a convex set on $\mathbb{R}^{K}$, therefore Eq.~\eqref{Aeq:q-obj} is convex w.r.t $\vect{\pi}$ and every local minima is a global minima. Similarly, it is also easy to show that Eq.~\eqref{Aeq:q-obj} is also convex w.r.t.~$\rho_t$ for $\rho_t\in[0,1]$.

\textbf{Optimization without inequality constraints} 
With only equality constraints, standard the Lagrangian Multiplier method can be applied. The Lagrangian can be written as:
\begin{equation}\label{Aeq:langrangian}
\begin{aligned}
    \mathcal{L}(\vect{\pi},\rho_t,\lambda) = & \sum^{N^t}_{i=1} \sum^K_{j=1} g_{ij}^{(m)} \log (\rho_t \cdot \pi_j) + \sum^{N^t}_{i=1} g_{iK+1}^{(m)} \log (1 - \rho_t) \\
    & + \sum^K_{j=1} (\alpha_j^{\textbf{in}}-1)\log\pi_j + (\alpha_1^{\textbf{out}} - 1 ) \cdot \log\rho_t +  (\alpha_2^{\textbf{out}} - 1) \cdot \log (1 - \rho_t) \\
    & + \lambda\left(1 - \sum^K_{j=1}\pi_j\right).
\end{aligned}
\end{equation}

The optimal $\vect{\pi},\rho_t$ can be found by taking all the partial derivative of $\mathcal{L}(\vect{\pi},\rho_t,\lambda)$ w.r.t $\pi_j, \rho_t$ and $\lambda$ to $0$:
\begin{equation}
\left\{
\begin{aligned}
    \frac{\partial \mathcal{L}}{\partial \pi_j} & = \frac{\sum^{N^t}_{i=1} g_{ij}^{(m)}}{\pi_j} +  \frac{\alpha_j^{\textbf{in}}-1}{\pi_j} - \lambda = 0 \\
    \frac{\partial \mathcal{L}}{\partial \rho_t} & = \frac{\sum^{N^t}_{i=1} \sum^K_{j=1} g_{ij}^{(m)} + (\alpha_1^{\textbf{out}} - 1 )}{\rho_t}  - \frac{\sum^{N^t}_{i=1} g_{iK+1}^{(m)} + (\alpha_2^{\textbf{out}} - 1)}{1 - \rho_t} = 0\\
    \frac{\partial \mathcal{L}}{\partial \lambda} & = \sum^K_{i=1}\pi_i - 1 = 0.
\end{aligned}
\right.
\end{equation}

The solution to the above equation set can be written as:
\begin{equation}
\left\{
\begin{aligned}
    \pi_j & =  \frac{\sum^{N^t}_{i=1} g_{ij}^{(m)} + \alpha_j^{\textbf{in}} - 1}{\lambda} \\\
    \rho_t & =  \frac{\sum^{N^t}_{i=1}\sum^{K}_{j=1}g_{ij}^{(m)} + \alpha_1^{\textbf{out}} - 1}{N^t + \alpha_1^{\textbf{out}} + \alpha_2^{\textbf{out}} - 2} \\
    \lambda & = \sum^{N^t}_{i=1}\sum^{K}_{j=1}g_{ij}^{(m)} + \sum^K_{l=1}(\alpha_l^{\textbf{in}} - 1).
\end{aligned}
\right.
\end{equation}

Therefore optimal $\vect{\pi}, \rho_t$ for $Q(\vect{\pi},\rho_t|\vect{\pi}^{(m)},\rho_t^{(m)})$ without inequality constraints is given by:
\begin{equation}\label{Aeq:equality-solution}
\begin{aligned}
    \pi_j & = \frac{\sum^{N^t}_{i=1} g_{ij}^{(m)} + \alpha_j^{\textbf{in}} - 1}{\sum^{N^t}_{i=1}\sum^{K}_{j=1}g_{ij}^{(m)} + \sum^K_{l=1}(\alpha_l^{\textbf{in}} - 1)}, \\
    \rho_t & =  \frac{\sum^{N^t}_{i=1}\sum^{K}_{j=1}g_{ij}^{(m)} + \alpha_1^{\textbf{out}} - 1}{N^t + \alpha_1^{\textbf{out}} + \alpha_2^{\textbf{out}} - 2}
\end{aligned}
\end{equation}

\textbf{Proof that the solution satisfies inequality constraints} 
Note that we have:
\begin{itemize}
    \item $g_{ij}^{(m)}$ in Eq.~\eqref{Aeq:gij-2} is non-negative
    \item $\Tilde{c}_i>0, i=1,2...K$ is non-negative
    \item $\alpha^{\textbf{in}}_i-1>0, i=1,2...K$ and $\alpha^{\textbf{out}}_1-1>0, \alpha^{\textbf{out}}_2-1>0$
\end{itemize}
Therefore we have $\vect{\pi}^{(t)}> 0 \Rightarrow\vect{\pi}^{(m+1)}>0$. Because the optimization problem is convex, when $\pi_j^{(t)}>0, j=1,2,...K$, Eq.~\eqref{Aeq:equality-solution} gives the global optimal $\vect{\pi}^{(m+1)},\rho_t^{(m+1)}$ for the optimization problem in Eq.~\eqref{Aeq:q-obj}:
\begin{equation}
\left\{
\begin{aligned}
    \pi_j^{(m+1)} & = \frac{\sum^{N^t}_{i=1} g_{ij}^{(m)} + \alpha_j^{\textbf{in}} - 1}{N^t - \sum^{N^t}_{i=1} g_{iK+1}^{(m)} + \sum^K_{l=1}(\alpha_l^{\textbf{in}} - 1)}  \quad \text{for all} \quad  i\in\mathcal{Y}\\
    \rho_t^{(m+1)} & =  \frac{N^t - \sum^{N^t}_{i=1} g_{iK+1}^{(m)} + \alpha_1^{\textbf{out}} - 1}{N^t + \alpha_1^{\textbf{out}} + \alpha_2^{\textbf{out}} - 2},
\end{aligned}
\right.
\end{equation}
given the fact that $\sum^{N^t}_{i=1}\sum^{K}_{j=1}g_{ij}^{(m)} = N^t - \sum^{N^t}_{i=1} g_{iK+1}^{(m)}$.

\end{proof}

\subsection{Proof of Theorem~\ref{theorem:estrho-linear} (See page~\pageref{theorem:estrho-linear})}

\rhocali*

\begin{proof}
For target domain dataset $\mathcal{D}^t$, if we are given ID/OOD label:  $\mathcal{D}^t=\mathcal{D}^{\textbf{ti}}\cup\mathcal{D}^{\textbf{to}}$, for a practical classifier $h'(x)$ we can write:
\begin{equation}\label{Aeq:expectedh}
\begin{aligned}
    \rho':=\mathbb{E}_{X_t} [h'(x)] & = \mathbb{E}_{X_t|B_t=1} [h'(x)] \cdot p_t(b=1) + \mathbb{E}_{X_t|B_t=0} [h'(x)] \cdot p_t(b=0) \\
    & = \rho_t \cdot  \mathbb{E}_{X_t|B_t=1} [h'(x)] + (1- \rho_t) \cdot \mathbb{E}_{X_t|B_t=0} [h'(x)] \\
    & = \mu^t_1 \cdot \rho_t + \mu^t_0 \cdot (1 - \rho_t),
\end{aligned}
\end{equation}
where:
\begin{equation}
    \mu^t_1 := \mathbb{E}_{X_t|B_t=1} [h'(x)] \quad \text{and} \quad  \mu^t_0 := \mathbb{E}_{X_t|B_t=0} [h'(x)].
\end{equation}
Rearranging Eq.~\eqref{Aeq:expectedh}, we have that:
\begin{equation}\label{Aeq:est-cali}
    \rho_t =\frac{1}{\mu^t_1 - \mu^t_0} \rho' - \frac{\mu^t_0}{\mu^t_1 - \mu^t_0},
\end{equation}
where the equation holds when $\mu^t_1 \neq \mu^t_0$:
\begin{equation}
    \mu^t_1 = \mathbb{E}_{X_t|B_t=1} [h'(x)] \neq \mathbb{E}_{X_t|B_t=0} [h'(x)] = \mu^t_0.
\end{equation}

\noindent \textbf{Option 1 \eqref{eq:rho-t-correction-condition}}: Under Assumption~\ref{assume:A1}, the condition \eqref{eq:rho-t-correction-condition} holds implies:
\begin{equation}
        \mathbb{E}_{X_s|Y_s=i}[h'(x)] = \mathbb{E}_{X_t|Y_t=j}[h'(x)] \quad \text{for all} \quad i,j\in\mathcal{Y}, 
\end{equation}
then according Eq.~\eqref{eq:label-distribution} we have:
\begin{equation}
\begin{aligned}
    \mu^t_1 = \mathbb{E}_{X_t|B_t=1} [h'(x)] & = \sum^K_{i=1}\mathbb{E}_{X_t|Y_t=i} [p_t(y=i|b=1)\cdot h'(x)] \\
    & = \sum^K_{i=1}\mathbb{E}_{X_t|Y_t=i} [\pi_i\cdot h'(x)]  = \sum^K_{i=1}\pi_i\cdot \mathbb{E}_{X_t|Y_t=1} [h'(x)] \\
    & = \mathbb{E}_{X_t|Y_t=1} [h'(x)]  = \mathbb{E}_{X_s|Y_s=1} [h'(x)] \\
    & = \sum^K_{i=1} c_i \mathbb{E}_{X_s|Y_s=1} [h'(x)] = \sum^K_{i=1}\mathbb{E}_{X_s|Y_s=i} [c_i\cdot h'(x)]  \\
    & = \sum^K_{i=1}\mathbb{E}_{X_s|Y_s=i} [p_s(y=i|b=1)\cdot h'(x)]  \\
    & = \mathbb{E}_{X_t|B_t=1} [h'(x)] = \mu'_1.
\end{aligned}
\end{equation}

\noindent \textbf{Option 2 $\vect{\pi}=\vect{c}$} Condition Eq.~\eqref{eq:label-distribution} can actually be replace with $\vect{\pi}=\vect{c}$ with the results still holds:
\begin{equation}
\begin{aligned}
    \mu^t_1 = \mathbb{E}_{X_t|B_t=1} [h'(x)] & = \sum^K_{i=1}\mathbb{E}_{X_t|Y_t=i} [p_t(y=i|b=1)\cdot h'(x)] \\
    & = \sum^K_{i=1}\mathbb{E}_{X_t|Y_t=i} [\pi_i\cdot h'(x)]  = \sum^K_{i=1}\mathbb{E}_{X_s|Y_s=i} [c_i\cdot h'(x)]  \\
    & = \sum^K_{i=1}\mathbb{E}_{X_s|Y_s=i} [p_s(y=i|b=1)\cdot h'(x)]  = \mathbb{E}_{X_t|B_t=1} [h'(x)] = \mu'_1,
\end{aligned}
\end{equation}
where $\mu'_1:=\mathbb{E}_{X_s|B_s=1} [h'(x)]$.

For both options we have:
\begin{equation}
    \begin{aligned}
    \mu^t_0 = \mathbb{E}_{X_t|B_t=0} [h'(x)] & = \mathbb{E}_{X_t|Y_t=K+1} [p_t(y=K+1|b=0)\cdot h'(x)] \\
    & = \mathbb{E}_{X_s|Y_s=K+1} [p_s(y=K+1|b=0)\cdot h'(x)]= \mathbb{E}_{X_s|B_s=0} [h'(x)] = \mu'_0.
\end{aligned}
\end{equation}
where $\mu'_0:=\mathbb{E}_{X_s|B_s=0} [h'(x)]$ are defined in the same way as $\mu_1, \mu_0$ defined in Theorem~\ref{theorem:est-rho} but substitute $h$ as $h'$.

The expectations can be approximated by $\hat{\mu}_1', \hat{\mu}_1'$ with source domain ID dataset $\mathcal{D}^s$ and OOD dataset $\mathcal{D}^{\textbf{o}}$ (Eq.~\eqref{eq:est-rho-s-sigma}). Moreover, $\mathbb{E}_{X_t} [h(x)]=\rho$ can be estimated with $\hat{\rho}$ given target dataset $\mathcal{D}^t$:
\begin{equation}
\begin{aligned}
    \hat{\mu}_1' :=  \frac{1}{|\mathcal{D}^{s}|} \sum_{x_i\in\mathcal{D}^{s}} h'(x_i), \quad \hat{\mu}_0':= \frac{1}{|\mathcal{D}^{\textbf{o}}|} \sum_{x_i\in\mathcal{D}^{\textbf{o}}} h'(x_i) \quad \text{and} \quad \hat{\rho}' :=  \frac{1}{|\mathcal{D}^{t}|} \sum_{x_i\in\mathcal{D}^{t}} h'(x_i).
\end{aligned}
\end{equation}

Therefore as long as \eqref{eq:rho-t-correction-condition} holds, we can use $\mathcal{D}^s$ and $\mathcal{D}^{\textbf{o}}$ to estimate $\rho_t$ with Eq.~\eqref{Aeq:est-cali}:
\begin{equation}
\begin{aligned}
    \rho_t 
    & \approx \hat{\rho_t}^* := \frac{1}{\hat{\mu}_1' - \hat{\mu}_0'} \hat{\rho}' - \frac{\hat{\mu}_0'}{\hat{\mu}_1' - \hat{\mu}_0'},
\end{aligned}
\end{equation}

Note that since $h'(x)\in[0,1]$, Hoeffding's inequality guarantees for some small $\epsilon> 0$:
\begin{equation}\label{Aeq:binary-hoeffding2}
\begin{aligned}
    p\left(\vert\mu_0' - \hat{\mu}_0'\vert \geq \epsilon\right) & \leq 2e^{-2|\mathcal{D}^{\textbf{so}}|\epsilon^2} \\
    p\left(\vert\mu_1'  - \hat{\mu}_1'\vert \geq \epsilon\right) & \leq 2e^{-2|\mathcal{D}^{\textbf{si}}|\epsilon^2} \\
    p\left(\vert \rho' - \hat{\rho}' \vert \geq \epsilon\right) & \leq 2e^{-2|\mathcal{D}^t|\epsilon^2},
\end{aligned}
\end{equation}

Therefore with high probability of at least $1 - 2e^{-2\min(|\mathcal{D}^{\textbf{si}}|,|\mathcal{D}^{\textbf{so}}|, |\mathcal{D}^t|)\epsilon^2}$ we have:
\begin{equation}
\left\{
\begin{aligned}
    \rho_t - \hat{\rho}^*_t & \leq \frac{\vert \rho' - \mu_0'\vert}{\vert \mu_1' - \mu_0' \vert} - \frac{\vert \rho' - \epsilon- \mu_0' -\epsilon\vert}{\vert\mu_1' + \epsilon- \mu_0' -\epsilon\vert} = \frac{2\epsilon}{\vert\mu_1' - \mu_0'\vert} \\
    \rho_t - \hat{\rho}^*_t & \geq \frac{-\vert\rho' - \mu_0'\vert}{\vert\mu_1' - \mu_0'\vert} - \frac{-\vert\rho' + \epsilon- \mu_0' +\epsilon\vert}{\vert\mu_1' - \epsilon- \mu_0' +\epsilon\vert} = \frac{-2\epsilon}{\vert\mu_1' - \mu_0'\vert},
\end{aligned}
\right.
\end{equation}
which is equivalent to:
\begin{equation}
    \vert \rho_t - \hat{\rho}^*_t \vert \leq \frac{2\epsilon}{\vert\mu_1' - \mu_0'\vert}.
\end{equation}

Letting $\delta:=e^{-2\min(|\mathcal{D}^{\textbf{si}}|,|\mathcal{D}^{\textbf{so}}|, |\mathcal{D}^t|)\epsilon^2}$, rearrange the equations and we get the result.

\end{proof}


\subsection{Further Discussion on $\rho_t$ correction model}\label{Asubsec:rho-t-correction-discussion}

This section further discusses the $\rho_t$ correction model Eq.~\eqref{eq:rho-t-correction} proposed in \S\ref{subsec:rho-t-correction} in our main paper. The model adjusts $\rho_t^{\text{MLE}}$ and $\rho_t^{\text{MAP}}$ obtained in Alg.~\ref{alg:MAPOLS} with Eq.~\eqref{eq:rho-t-correction}, which is based on Theorem~\ref{theorem:estrho-linear}. 

We will show that for a special case, the MLE $\rho_t^{\text{MLE}}$ defined in MLE objective Eq.~\eqref{eq:MLE-obj} will have a closed-form solution, which is simply averaging the response of $h(x)$ on target dataset $\mathcal{D}^t$:

\begin{restatable}{lemma}{mlesimple}\label{lemma:rho-t-mle}
    Under Assumption~\ref{assume:A1},\ref{assume:A2}, if $\vect{\pi}=\vect{c}$ and $h:\mathcal{X}\rightarrow\{0,1\}$, then the $\rho_t^{\text{MLE}}$ defined in Eq.~\eqref{eq:MLE-obj} can be obtained given target dataset $\mathcal{D}^t$ via:
    \begin{equation}
        \rho_t^{\text{MLE}} = \frac{1}{N^t}\sum^{N^t}_{i=1}h(x_i).
    \end{equation}
\end{restatable}

\begin{proof}

When Assumption~\ref{assume:A2}B is satisfied, given the information available, substituting:
\begin{equation}
    p_s(b=1|x)=h(x) \in\{0,1\} \quad \text{and}\quad \vect{\pi}=\vect{c}
\end{equation}
into the NLL in Eq.~\eqref{eq:nll} and we have:
\begin{equation}
\begin{aligned}
    - \log L(\rho_t;\mathcal{D}^t) &= - \sum^{N^t}_{i=1} \log \Bigg(\frac{\rho_t}{\rho_s}h(x_i) \cdot \sum^K_{j=1}\frac{\pi_j}{c_j} f(x_i)_j  +  \frac{1 - \rho_t}{1 - \rho_s} \cdot (1- h(x_i))\Bigg) + C \\
    & = - \sum^{N^t}_{i=1} \log \Bigg(\frac{\rho_t}{\rho_s}h(x_i) \cdot 1 +  \frac{1 - \rho_t}{1 - \rho_s} \cdot (1- h(x_i))\Bigg) + C \\
    & = - \sum^{N^t}_{i=1} \mathbb{I}_1(h(x_i))\cdot \log \Bigg(\frac{\rho_t}{\rho_s}\Bigg)  - \sum^{N^t}_{i=1} (1 - \mathbb{I}_1(h(x_i))) \cdot \log \Bigg(\frac{1 - \rho_t}{1 - \rho_s} \Bigg) + C
\end{aligned}
\end{equation}

Let the derivative w.r.t.~$\rho_t$ equals $0$ and we have:
\begin{equation}
\begin{aligned}
    \frac{d \left(- \log L(\rho_t;\mathcal{D}^t)\right)}{d\rho_t} & =  - \sum^{N^t}_{i=1} \mathbb{I}_1(h(x_i))\cdot \frac{\rho_s}{\rho_t} \cdot \frac{1}{\rho_s}  - \sum^{N^t}_{i=1} (1 - \mathbb{I}_1(h(x_i)))\cdot \frac{1 - \rho_s}{1 - \rho_t} \cdot \frac{-1}{1-\rho_s} \\
    & = - \sum^{N^t}_{i=1} \mathbb{I}_1(h(x_i))\cdot \frac{1}{\rho_t}   + \sum^{N^t}_{i=1} (1 - \mathbb{I}_1(h(x_i)))\cdot \frac{1}{1 - \rho_t}  = 0
\end{aligned}
\end{equation}
Solve the above equation for $\rho_t$ and we get:
\begin{equation}
    \rho_t = \frac{1}{N^t}\sum^{N^t}_{i=1}h(x_i),
\end{equation}
which is the closed-form solution to the MLE objective Eq.~\eqref{eq:nll} under the special setting of no ID label shift ($\vect{\pi}=\vect{c}$) and a discrete ID/OOD classifier ($h:\mathcal{X}\rightarrow\{0,1\}$). \\
\end{proof}

As shown in Lemma~\ref{lemma:rho-t-mle}, when $\vect{\pi}=\vect{c}$ and $h:\mathcal{X}\rightarrow\{0,1\}$, $\rho_t^{\text{MLE}}$ can be obtained by averaging $h(x)=p_s(b=1|x)$ over the target dataset $\mathcal{D}^t$ based on Assumption~\ref{assume:A2}B, i.e. $h(x)\neq p_s(b=1|x)$ when the assumption is not satisfied, according the proof of Theorem~\ref{theorem:estrho-linear}, the condition $\vect{\pi}=\vect{c}$ enable us to use:

\begin{equation} \label{Aeq:rho-correction}
    \hat{\rho}^*_t = \frac{\hat{\rho} - \hat{\mu}_0'}{\hat{\mu}_1' - \hat{\mu}_0'}, \quad \text{and} \quad \hat{\rho}'  := \frac{1}{|\mathcal{D}^t|}\sum_{x_i\in\mathcal{D}^t} h(x_i),
\end{equation}
to obtain the estimate of ground truth $\rho_t$.

\clearpage

\section{Experimental Setup}\label{Asec:experment-setup}
\subsection{ID Classifier Details} \label{Asubsec:id-classifier-details}
Our code for training the ID classifier and constructing the OOD classifier is mainly based on the open source project OpenOOD~\cite{yang2022openood,zhang2023openood} on OOD detection. The project is publicly available at \url{https://github.com/Jingkang50/OpenOOD}. 

We follow the basic setup in OpenOOD to train the ID classifier $f$, where we train a ResNet18 model for the CIFAR10/100 and ImageNet-200 datasets. Each model is trained 3 times with different random seeds.

\begin{table}[ht]
    \centering
    \footnotesize
    \begin{tabular}{c| c c c c c c} \hline\hline
         Dataset & Model & Setup & optimizer & lr & weight decay & epoch  \\ \hline
         CIFAR10 & ResNet18 &Train from Scratch & SGD & $0.1$ & $5e^{-4}$ & 100   \\
         CIFAR100 & ResNet18 &Train from Scratch & SGD  & $0.1$ & $5e^{-4}$ & 100   \\
         ImageNet-200 & ResNet18 & Train from Scratch & SGD  & $0.1$ & $5e^{-4}$ & 90 \\
         \hline\hline
    \end{tabular}
    \vspace{-1em}
    \caption{Source domain ID classifier $f$ setup used in our model. The setup follows exactly the OpenOOD project implementation (retrieved on May 2024). }
    \vspace{-1em}
    \label{Atab:model-implementation}
\end{table}

\subsection{OOD Classifier Details}\label{Asubsec:ood-classifier-details}
We use the implementation provided in OpenOOD project to construct the OOD detection binary classifiers $h$ proposed by OpenMax~\cite{openmax}, Ash~\cite{ash}, MLS~\cite{mls}, ReAct~\cite{sun2021react} and KNN~\cite{knn}. All the OOD detection models are post-hoc inference models based on the ID classifier $f$. The detailed hyper-parameter setups of each OOD detector are listed in Tab.~\ref{Atab:ood-classifier-setup}, where:
\begin{itemize}
    \item OpenMax has no official implementations, we follow the OpenOOD implementation with the same hyperparameter provided by the code.
    \item KNN follows the exact hyperparameter setup of the original paper K=50, which is also adopted in the OpenOOD.
    \item MLS does not require any hyperparameter.
    \item ASH has one hyperparameter "percentile", which is obtained with a parameter search among [65, 70, 75, 80, 85, 90, 95] over a validation set (subset of the source domain dataset $\mathcal{D}^s$ ) provided by OpenOOD, the original paper simply choose 65.
    \item ReAct has one hyperparameter "percentile", which is also obtained with a parameter search among [85, 90, 95, 99] over a subset of $\mathcal{D}^s$ provided by OpenOOD, the original simply chose 90.
\end{itemize}

\begin{table}[ht]
    \centering
    \footnotesize
    \begin{tabular}{c|| c | c} \hline
        Model Name & Source Code & Date of Retrieval  \\ \hline \hline
        OpenOOD~\cite{yang2022openood} & \url{https://github.com/Jingkang50/OpenOOD}  & May 2024\\ \hline
        OpenMax \cite{openmax}  & \url{https://github.com/Jingkang50/OpenOOD} & May 2024 \\ \hline
        KNN \cite{knn} &  \url{https://github.com/deeplearning-wisc/knn-ood} & May 2024 \\  \hline
        MLS \cite{mls} & \url{https://github.com/Jingkang50/OpenOOD}  & May 2004 \\ \hline
        Ash \cite{ash} & \url{https://github.com/andrijazz/ash} & May 2024 \\ \hline
        ReAct~\cite{sun2021react} & \url{https://github.com/deeplearning-wisc/react} & May 2024\\\hline
        \hline 
        \hline
    \end{tabular}
    \caption{Source code details of reproduced OOD detection models. The code for OpenMax, KNN, MLS, Ash and ReAct have been collected in the OpenOOD project and can be directly tested within the project.}
    \label{Atab:ood-reproduce}
\end{table}

\begin{table}[ht]
    \centering
    \begin{tabular}{c|c} \hline\hline
        OOD classifier & hyper-parameters \\ \hline
        OpenMax & Weibull fitting: alpha=3, threshold=0.9, tail=20; coreset\_sampling\_ratio=0.01; \\
        KNN & \# of nearest neighbor $K=50$ \\
        MLS & - \\
        Ash & parameter search on percentile=[65, 70, 75, 80, 85, 90, 95] \\
        ReAct & parameter search on percentile=[85, 90, 95, 99] \\
         \hline\hline
    \end{tabular}
        \vspace{-1em}
    \caption{Detailed hyper-parameter setups of the OOD detectors used in our work. All the hyper-parameter setup are following the default setups provided by the OpenOOD project~\cite{yang2022openood} (retrieved on May 2024).}
    \label{Atab:ood-classifier-setup}
\end{table}

\noindent\textbf{Output Re-scaling:} Existing OOD classifiers focus more on ID/OOD separation and hence usually output a real valued scalar instead of $[0,1]$ confidence. For example, the MLS~\cite{mls} model actually outputs the max logit of the ID classifier's prediction. To use these OOD classifiers in our OSLS estimation model, we need to re-scale the output of these OOD classifier in the binary range $[0,1]$.

In this work, we re-scale an OOD classifier $h':\mathcal{X}\rightarrow\mathbb{R}$ to a binary classifier $h:\mathcal{X}\rightarrow [0,1]$ with two approaches: \textbf{logistic regression} and \textbf{thresholding}. The logistic regression model $h_0:\mathbb{R}_+\rightarrow[0,1]$ is trained based on the source domain ID dataset $\mathcal{D}^s$ and reference OOD dataset $\mathcal{D}^{\textbf{o}}$ (see Fig.~\ref{fig:model-structure}). On the other hand, the thresholding approach obtains the threshold by computing the median values of the output of OOD classifier $h'$ given ID dataset $\mathcal{D}^s$ and OOD dataset $\mathcal{D}^{\textbf{o}}$. The threshold is picked as the average of the two median values. Details or the two re-scaling models are described in Tab.~\ref{Atab:rescale-model}.
\begin{table}[ht]
    \centering
    \footnotesize
    \begin{tabular}{c|c | c} \hline\hline
     Dataset & Re-scaling Model & Model Setup \\ \hline
      \multirow{2}{*}{CIFAR10/100} & \multirow{2}{*}{Logistic Regression} & epoch 100; optimizer: SGD; batch\_size: 512; lr 0.05; lr\_scheduler: Cosine; loss: BCE; \\
        & & $h(x) = 1/(1 + e^{-w\cdot h'(x) + b})$\\\hline
        ImageNet-200 & Thresholding & $h(x) = \begin{cases}
            1, & h'(x) > (\text{median}(h'(\mathcal{D}^s)) + \text{median}(h'(\mathcal{D}^\textbf{o}))) / 2\\
            0, & \text{Otherwise}
        \end{cases}$. \\
         \hline\hline
    \end{tabular}
    \vspace{-1em}
    \caption{Re-scaling model setup that normalize the output of a OOD classifier into $[0,1]$.}
    \label{Atab:rescale-model}
\end{table}

Although thresholding approach only outputs $\{0,1\}$ instead of a continuous confidence, it is suitable for our $\rho_t$ correction model (Sec.~\ref{subsec:rho-t-correction}) because the linear correction approach has theoretical guarantees when the ID/OOD classifier $h(x)$ outputs binary values (Theorem~\ref{theorem:estrho-linear}). Moreover, OOD detectors on large-scale datasets are more likely to violate Assumption~\ref{assume:A2}, thus the $\rho_t$ correction model might become more necessary.

\subsection{OOD reference Dataset details}

As discussed in the main paper, the reference OOD dataset $\mathcal{D}^{\textbf{o}}$ is generated by linear combination of Gaussian noise and ground truth samples in source domain ID dataset $\mathcal{D}^s$. The hyper-parameters used $\gamma, T$ used in the OOD dataset generation process and $\hat{\mu}_0$ rescaling are: CIFAR10: $\gamma=0.2, T=2$, CIFAR100: $\gamma=0.1, T=2$, ImageNet-200: $\gamma=0.2, T=2$.

\subsection{Datasets Details}\label{Asubsec:dataset-details}
\textbf{Train Datasets}: We use the standard CIFAR10/100 and ImageNet-200 datasets as ID datasets, with the detailed information provided in Tab.~\ref{Atab:dataset-info}.

\begin{table}[ht]
    \centering
    \footnotesize
    \begin{tabular}{c| c c c c c} \hline\hline
        Dataset & Train \# samples & Val \# samples & Test \# samples & \# of Classes \\ \hline
        CIFAR10  & 50k & 9000 & 1000 & 10  \\
        CIFAR100  & 50k & 9000 & 1000 & 100 \\
        ImageNet-200  & 260k & 1000 & 9000 & 200  \\ 
        \hline \hline
    \end{tabular}
    \vspace{-1em}
    \caption{Detailed information of ID datasets.}
    \label{Atab:dataset-info}
\end{table}

\textbf{Test Datasets:} We use the OOD datasets setup provided in the OpenOOD project, where validation sets of CIFAR10 and CIFAR100 are used as OOD datasets, with each contains 9000 samples. For the other OOD datatsets, TinyImageNet has 7793 samples, MNIST has 70000 samples, SVHN has 26032 samples, Texture has 5640 samples, Places has 35195 samples, SSB has 49000 samples, NINCO has 5879 sampels, iNaturalist has 10000 samples, OpenImage-O has 15869 samples. Many of these OOD datasets are actually subsampled from the original datasets to avoid overlapping in classes.

\textbf{Test label shifts:} We follow the MAPLS~\cite{mapls} official code to adjust the test datasets for ordered Long-Tail and Dirichlet label shift (retrieved in May 2024). More detailed information for different test set shifts can be found in the following Tab.~\ref{Atab:labelshift-setup}.

\begin{table}[ht]
    \centering
    \footnotesize
    \begin{tabular}{c|c c } \hline\hline
        Label Shift & Shift Parameters & OOD/ID data ratio $r$\\ \hline
        Dirichlet & $\alpha=1.0, 10.0$ (2500 samples) & $r=1, 0.1, 0.01$ \\ \hline
        \multirow{2}{*}{Ordered LT} & $100, 50, 10$ &  \multirow{2}{*}{$r=1, 0.1, 0.01$} \\
        & ``Forward/Backward" & \\
        \hline\hline
    \end{tabular}
    \vspace{-1em}
    \caption{\textbf{Types of label shift in our experiment,} including Dirichlet shift with different shift parameter $\alpha$ and Ordered Long-Tailed (LT) shift with different imbalance factors under forward and backward order.
    }
    \label{Atab:labelshift-setup}
\end{table}

\subsection{Closed Set Label Shift Estimation Model details} \label{Asubsec:csls}

We test closed set label shift estimation model BBSE~\cite{BBSE}, MLLS~\cite{saerens2002adjusting}, RLLS~\cite{RLLS} and MAPLS~\cite{mapls} based on the official implementation of MAPLS provided in \url{https://github.com/ChangkunYe/MAPLS} retrieved on June 2024. These models are used to test on the open set label shift dataset without any adjustment of hyper-parameters or other setups.

According to MAPLS, MLLS code is provided by Alexandari \etal~\cite{alexandari2020maximum} which has included the source code of RLLS \cite{RLLS} and BBSE \cite{BBSE} with their original github page provided in the Tab.~\ref{Atab:csls-reproduce}. Only RLLS has a hyperparameter in their model. We follow Alexandari \etal~\cite{alexandari2020maximum} and the RLLS original implementation to set the hyperparameter to be $\alpha=0.01$. 
\begin{table}[ht]
    \centering
    \begin{tabular}{c|| c | c} \hline
        Model Name & Source Code & Date of Retrieval  \\ \hline \hline
        \multirow{2}{*}{MLLS \cite{saerens2002adjusting, alexandari2020maximum}}  & \href{https://github.com/kundajelab/labelshiftexperiments}{https://github.com/kundajelab/labelshiftexperiments} & Aug 2022 \\
        & \href{https://github.com/kundajelab/abstention}{https://github.com/kundajelab/abstention} & Aug 2022 \\ \hline
        BBSE \cite{BBSE} &  \href{https://github.com/flaviovdf/label-shift}{https://github.com/flaviovdf/label-shift} & Aug 2022 \\  \hline
        RLLS \cite{RLLS} & \href{https://github.com/Angie-Liu/labelshift}{https://github.com/Angie-Liu/labelshift} & Aug 2022 \\
        \hline 
        \hline
    \end{tabular}
    \caption{Source Code details of reproduced existing label shift estimation models.}
    \label{Atab:csls-reproduce}
\end{table}

\subsection{EM algorithm} \label{Asubsec:em}

We use the same EM algorithm running procedure as MAPLS~\cite{mapls} proposed in the closed set label shift problem. Specifically, the procedure is as follows: 1) Initialize the target label distribution to be the same as source label distribution, i.e. $\vect{\pi}^{(0)}=\hat{\vect{c}}$ and $\rho_t^{(0)} = \hat{\rho}_s$, 2) Run EM algorithm~\ref{alg:MAPOLS} for 100 epoch to ensure convergence and 3) Output $\vect{\pi}^{(101)}$ and $\vect{\pi}^{(101)}$.

For the MAP estimate, we use the Adaptive Prior Learning (APL) model proposed by MAPLS~\cite{mapls} to determine parameter $\vect{\alpha}^{\textbf{in}}\in\mathbb{R}^{K}_{> 1}$ in the Dirichlet prior for ID classes and use no Bernoulli prior ($\alpha_1^{\textbf{out}}, \alpha_2^{\textbf{out}}=\vect{1}$). 



\clearpage
\section{More Visualizations and Ablation studies}\label{Asec:more-ablation}

\subsection{Target $\rho_t$ Estimation}
\begin{figure}[ht]
    \centering
    \includegraphics[width=1\linewidth]{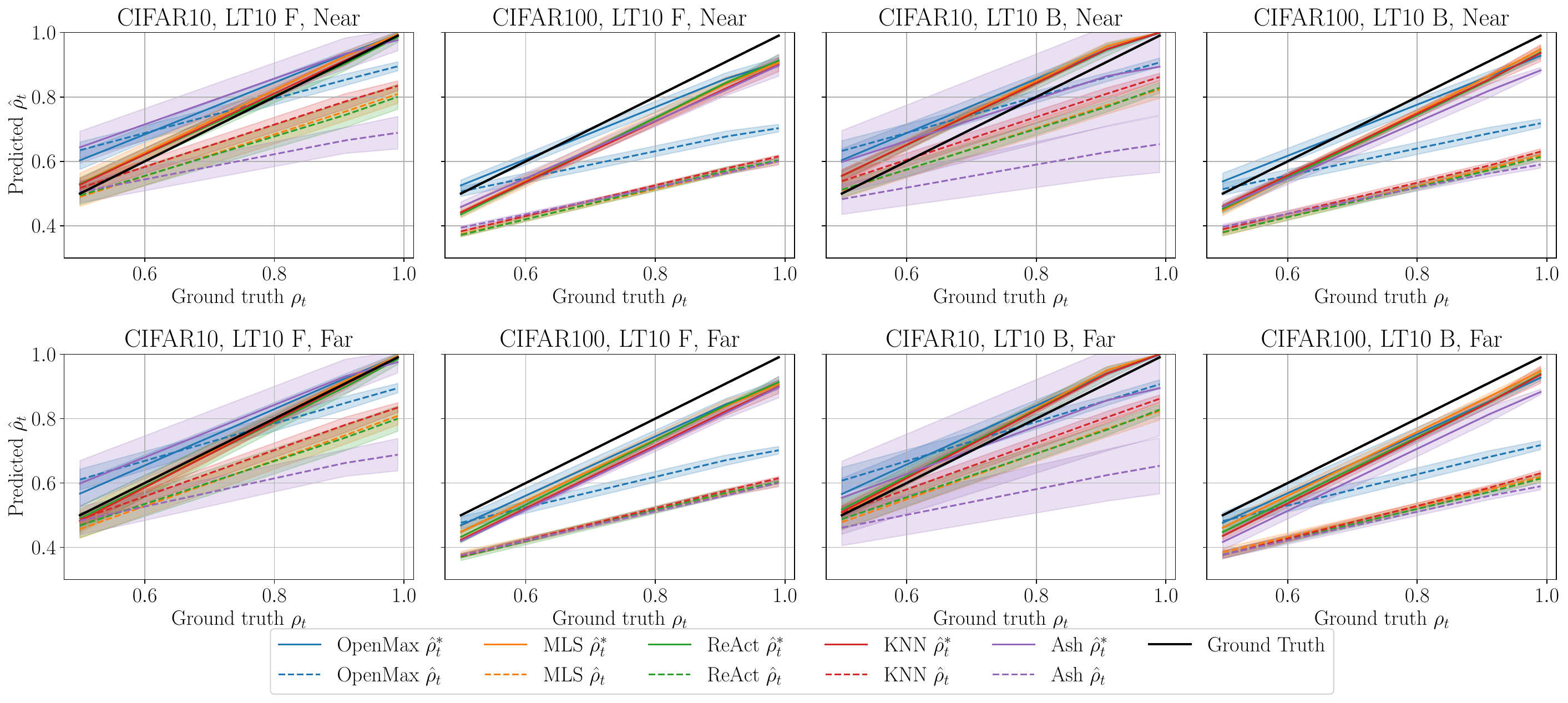}
        \vspace{-1em}
    \caption{\textbf{Estimation result comparison of $\hat{\rho}_t^{*}$ by our model (Solid lines), $\hat{\rho_t}$ by our model but without $\rho_t$ correction (Sec.~\ref{subsec:rho-t-correction}) (Dashed lines) based on different OOD classifiers and the Ground truth $\rho_t$ (Black, Solid line), on CIFAR10/100 dataset with LT10 shift  (``F" for Forward and ``B" for Backward) and Near + Far OOD dataset (Tab.~\ref{tab:dataset-setup}).} Shaded area are $\pm$ one standard deviation over corresponding OOD datasets and three independent ID classifiers.}
    \label{Afig:rho-t-cifar-LT10}
\end{figure}

\begin{figure}[ht]
    \centering
    \includegraphics[width=1\linewidth]{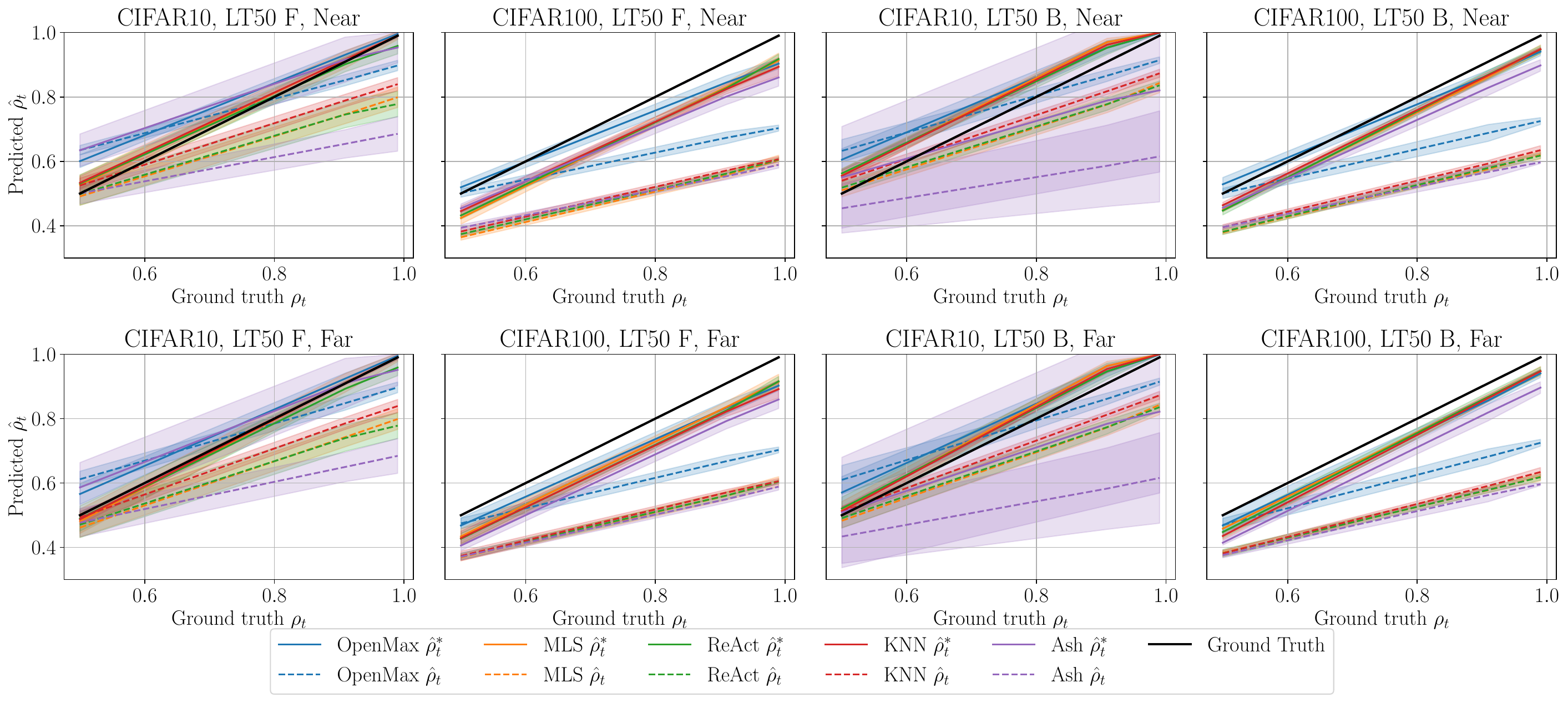}
        \vspace{-1em}
    \caption{\textbf{Estimation result comparison of $\hat{\rho}_t^{*}$ by our model (Solid lines), $\hat{\rho_t}$ by our model but without $\rho_t$ correction (Sec.~\ref{subsec:rho-t-correction}) (Dashed lines) based on different OOD classifiers and the Ground truth $\rho_t$ (Black, Solid line), on the CIFAR10/100 dataset with LT50 shift  (``F" for Forward and ``B" for Backward) and Near + Far OOD dataset (Tab.~\ref{tab:dataset-setup}).} Shaded areas are $\pm$ one standard deviation over corresponding OOD datasets and three independent ID classifiers.}
    \label{Afig:rho-t-cifar-LT50}
\end{figure}

\begin{figure}[ht]
    \centering
    \includegraphics[width=1\linewidth]{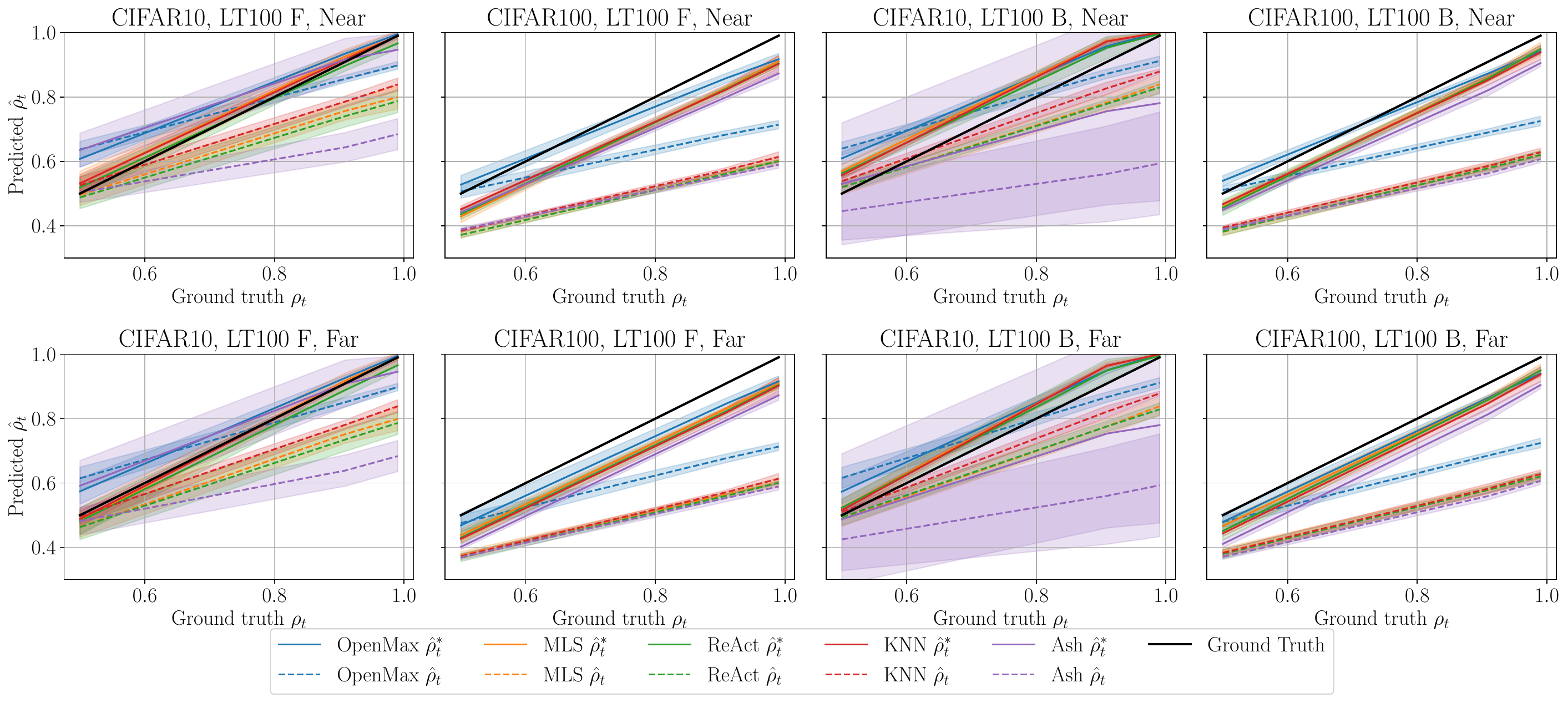}
    \vspace{-1em}
    \caption{\textbf{Estimation result comparison of $\hat{\rho}_t^{*}$ by our model (Solid lines), $\hat{\rho_t}$ by our model but without $\rho_t$ correction (Sec.~\ref{subsec:rho-t-correction}) (Dashed lines) based on different OOD classifiers and the Ground truth $\rho_t$ (Black, Solid line), on the CIFAR10/100 dataset with LT100 shift (``F" for Forward and ``B" for Backward) and Near + Far OOD dataset (Tab.~\ref{tab:dataset-setup}).} Shaded area are $\pm$ one standard deviation over corresponding OOD datasets and three independent ID classifiers.}
    \label{Afig:rho-t-cifar-LT100}
\end{figure}

\begin{figure}[ht]
    \centering
    \includegraphics[width=1\linewidth]{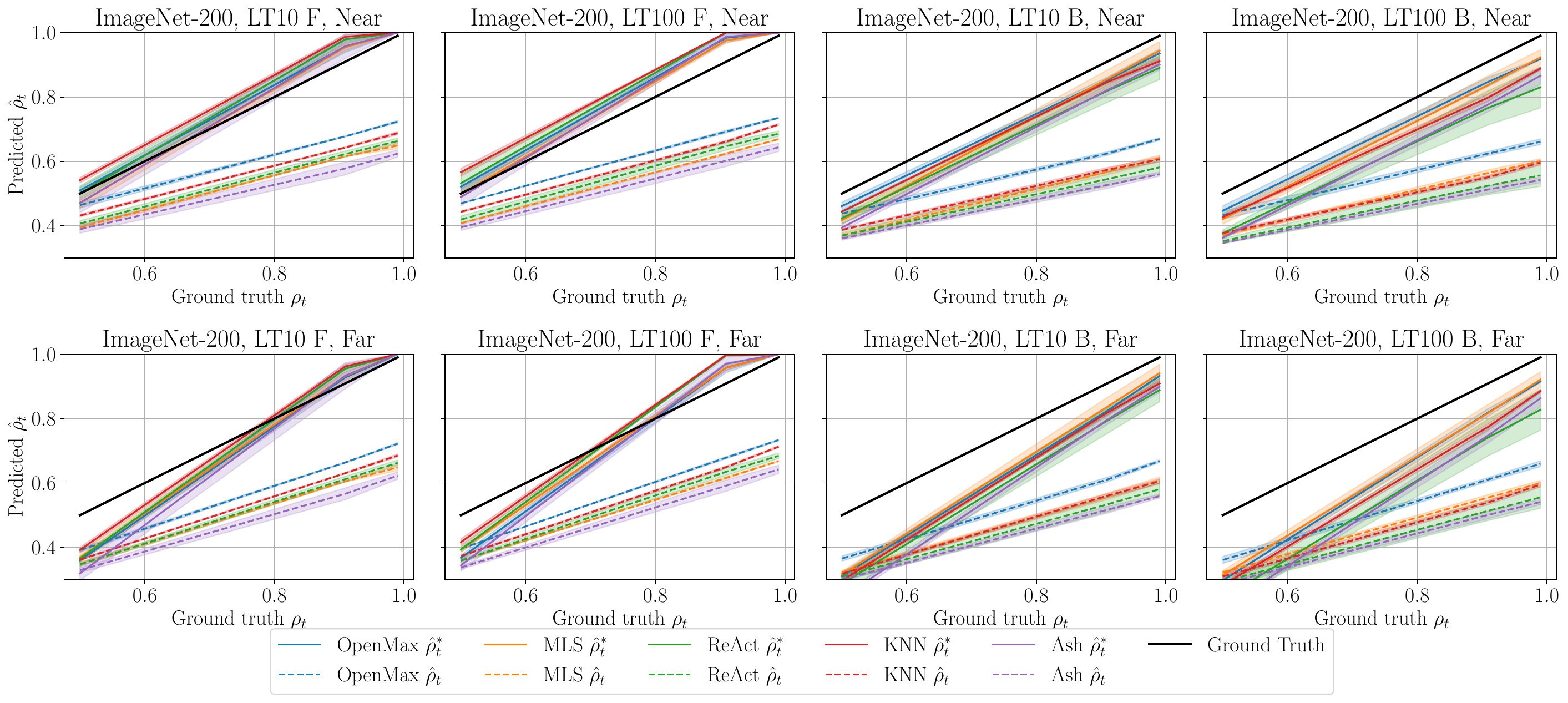}
    \vspace{-1em}
    \caption{\textbf{Estimation result comparison of $\hat{\rho}_t^{*}$ by our model (Solid lines), $\hat{\rho_t}$ by our model but without $\rho_t$ correction (Sec.~\ref{subsec:rho-t-correction}) (Dashed lines) based on different OOD classifiers and the Ground truth $\rho_t$ (Black, Solid line), on the ImageNet-200 dataset with LT10/LT100 shift (``F" for Forward and ``B" for Backward) and Near + Far OOD dataset (Tab.~\ref{tab:dataset-setup}).} Shaded area are $\pm$ one standard deviation over corresponding OOD datasets and three independent ID classifiers.}
    \label{Afig:rho-t-imagenet200-LT100}
\end{figure}

The full experiment visualizations of MAP estimate $\hat{\rho}_t^{*}$ with our model, $\hat{\rho}_t$ with our model but no $\rho_t$ correction in Sec.~\ref{subsec:rho-t-correction} and ground truth $\rho_t$ on different ID/OOD datasets with different label shift are listed in the following figures. 
\begin{itemize}
    \item Fig.~\ref{Afig:rho-t-cifar-dir} includes results on Dirichlet ID label shift on CIFAR10/100 datasets.
    \item Fig.~\ref{Afig:rho-t-cifar-LT10} includes results on LT10 ID label shift on CIFAR10/100 datasets.
    \item Fig.~\ref{Afig:rho-t-cifar-LT50} includes results on LT50 ID label shift on CIFAR10/1000 datasets.
    \item Fig.~\ref{Afig:rho-t-cifar-LT100} includes results on LT100 ID label shift on CIFAR10/1000 datasets.
    \item Fig.~\ref{Afig:rho-t-imagenet200-LT100} includes results on LT10/LT100 ID label shift on ImageNet-200 dataset.
\end{itemize}

As seen in the figures, the estimation result of our model exhibits a linear correlation with the ground truth, which is explained by our analysis in Theorem~\ref{theorem:estrho-linear}.
Moreover, our $\rho_t$ correction model is able to adjust the predicted $\hat{\rho}_t$ to $\hat{\rho}_t^{*}$ that is closer to the ground truth. 

\begin{figure}[ht]
    \centering
    \includegraphics[width=1\linewidth]{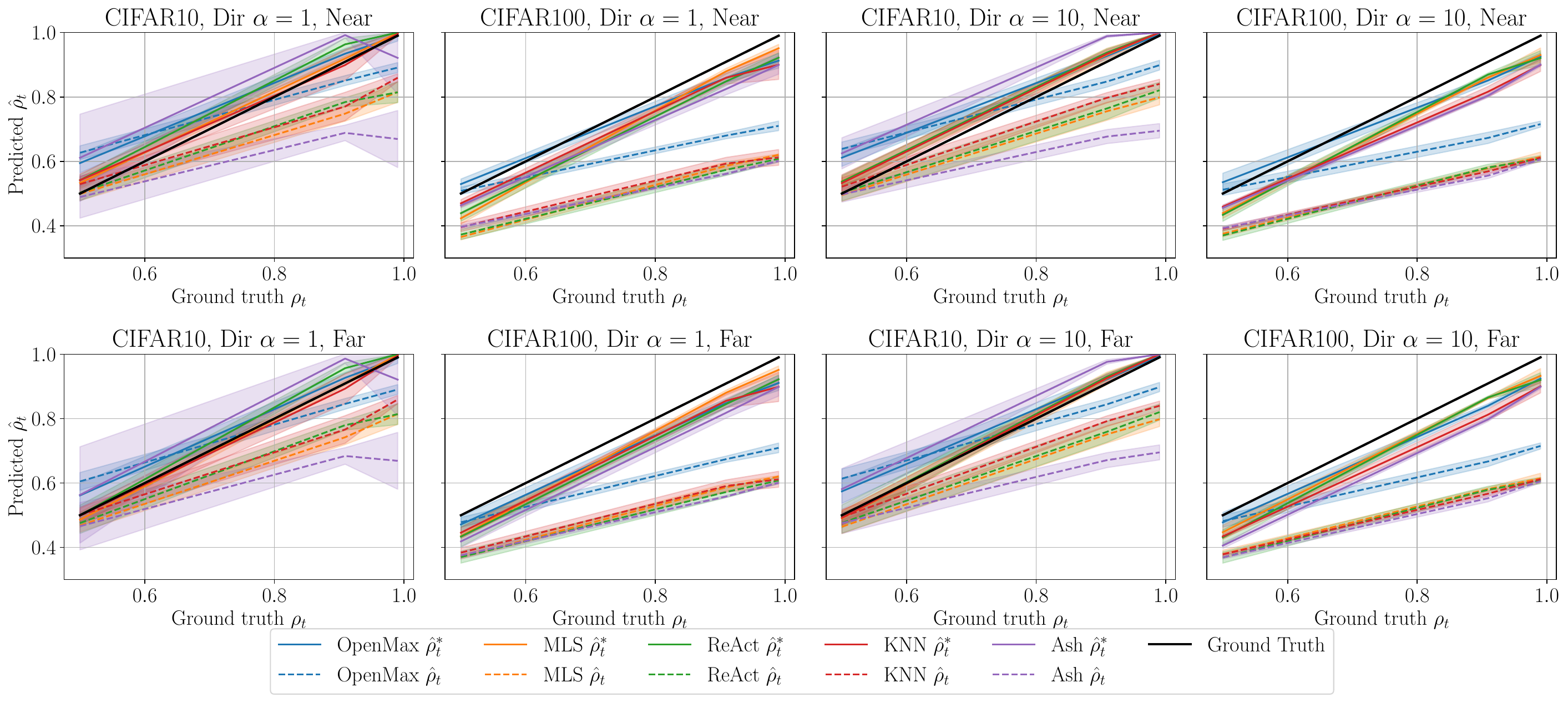}
        \vspace{-1em}
    \caption{\textbf{Estimation result comparison of $\hat{\rho}_t^{*}$ by our model (Solid lines), $\hat{\rho_t}$ by our model but without $\rho_t$ correction (Sec.~\ref{subsec:rho-t-correction}) (Dashed lines) based on different OOD classifiers and the Ground truth $\rho_t$ (Black, Solid line), on CIFAR10/100 dataset with Dirichlet shift and Near + Far OOD dataset (Tab.~\ref{tab:dataset-setup}).} Shaded area are $\pm$ one standard deviation over corresponding OOD datasets and three independent ID classifiers.}
    \label{Afig:rho-t-cifar-dir}
\end{figure}

\clearpage
\subsection{Hyperparameter sensitivity ablation}
This section provides the ablation study of the sensitivity of the hyperparameter $\gamma$ in Eq.~\eqref{eq:ood-dataset-gen} when generating pseudo OOD samples with Gaussian noise:
\begin{equation}
    \mathcal{D}^{\textbf{o}}_{\gamma}= \{(1 - \gamma) \cdot x_i + \gamma\cdot \epsilon\vert x_i\in\mathcal{D}^s, \epsilon\sim\mathcal{N}(0,1)\}.
\end{equation}

As shown in Tab.~\ref{Atab:wmse-cifar100-LT-gamma-ablation}, our model exhibits stable performance when $\gamma$ varies.

\setlength\tabcolsep{0.8pt} 
\begin{table*}[ht]
    \centering
    \scriptsize
    \renewcommand{\arraystretch}{1.1}
    \begin{tabular}{c | c | c||c c c | c c c | c c c } \hline\hline
    \multicolumn{3}{c||}{Dataset} & \multicolumn{9}{c}{CIFAR100} \\ \hline
       \multicolumn{3}{c||}{ID label Shift param}  & \multicolumn{3}{c|}{LT10 Forward} & \multicolumn{3}{c|}{LT50 Forward} & \multicolumn{3}{c}{LT100 Forward} \\ \hline
      \multicolumn{3}{c||}{OOD label shift param $r$} & $1.0$ & $0.1$ & $0.01$  & $1.0$ & $0.1$ & $0.01$  & $1.0$ & $0.1$ & $0.01$ \\ \hline \hline

\multicolumn{12}{c}{Closed Set Label Shift estimation models} \\ \hline\hline 
\multicolumn{2}{c|}{\multirow{2}{*}{BBSE}}  & Near & $0.560_{\scaleto{\pm 0.038}{3pt}}$ & $0.121_{\scaleto{\pm 0.027}{3pt}}$ & $0.107_{\scaleto{\pm 0.026}{3pt}}$ & $0.758_{\scaleto{\pm 0.057}{3pt}}$ & $0.171_{\scaleto{\pm 0.044}{3pt}}$ & $0.136_{\scaleto{\pm 0.031}{3pt}}$ & $0.841_{\scaleto{\pm 0.038}{3pt}}$ & $0.188_{\scaleto{\pm 0.048}{3pt}}$ & $0.151_{\scaleto{\pm 0.030}{3pt}}$\\ 
\multicolumn{2}{c|}{} & Far &  $4.128_{\scaleto{\pm 0.245}{3pt}}$ & $0.253_{\scaleto{\pm 0.028}{3pt}}$ & $0.109_{\scaleto{\pm 0.027}{3pt}}$ & $4.370_{\scaleto{\pm 0.301}{3pt}}$ & $0.291_{\scaleto{\pm 0.043}{3pt}}$ & $0.139_{\scaleto{\pm 0.034}{3pt}}$ & $4.431_{\scaleto{\pm 0.228}{3pt}}$ & $0.306_{\scaleto{\pm 0.054}{3pt}}$ & $0.153_{\scaleto{\pm 0.031}{3pt}}$\\ \hline 
\multicolumn{2}{c|}{\multirow{2}{*}{MLLS}}  & Near & $0.906_{\scaleto{\pm 0.061}{3pt}}$ & $0.114_{\scaleto{\pm 0.028}{3pt}}$ & $0.088_{\scaleto{\pm 0.028}{3pt}}$ & $1.029_{\scaleto{\pm 0.066}{3pt}}$ & $0.155_{\scaleto{\pm 0.044}{3pt}}$ & $0.105_{\scaleto{\pm 0.028}{3pt}}$ & $1.072_{\scaleto{\pm 0.067}{3pt}}$ & $0.150_{\scaleto{\pm 0.042}{3pt}}$ & $0.113_{\scaleto{\pm 0.034}{3pt}}$\\ 
\multicolumn{2}{c|}{} & Far &  $9.633_{\scaleto{\pm 1.442}{3pt}}$ & $0.348_{\scaleto{\pm 0.057}{3pt}}$ & $0.092_{\scaleto{\pm 0.028}{3pt}}$ & $9.910_{\scaleto{\pm 1.551}{3pt}}$ & $0.380_{\scaleto{\pm 0.044}{3pt}}$ & $0.108_{\scaleto{\pm 0.030}{3pt}}$ & $9.896_{\scaleto{\pm 1.523}{3pt}}$ & $0.373_{\scaleto{\pm 0.065}{3pt}}$ & $0.115_{\scaleto{\pm 0.035}{3pt}}$\\ \hline 
\multicolumn{2}{c|}{\multirow{2}{*}{RLLS}}  & Near & $0.426_{\scaleto{\pm 0.000}{3pt}}$ & $0.425_{\scaleto{\pm 0.000}{3pt}}$ & $0.425_{\scaleto{\pm 0.000}{3pt}}$ & $1.100_{\scaleto{\pm 0.000}{3pt}}$ & $1.099_{\scaleto{\pm 0.000}{3pt}}$ & $1.099_{\scaleto{\pm 0.000}{3pt}}$ & $1.404_{\scaleto{\pm 0.000}{3pt}}$ & $1.402_{\scaleto{\pm 0.000}{3pt}}$ & $1.402_{\scaleto{\pm 0.000}{3pt}}$\\ 
\multicolumn{2}{c|}{} & Far &  $0.426_{\scaleto{\pm 0.000}{3pt}}$ & $0.425_{\scaleto{\pm 0.000}{3pt}}$ & $0.425_{\scaleto{\pm 0.000}{3pt}}$ & $1.100_{\scaleto{\pm 0.000}{3pt}}$ & $1.099_{\scaleto{\pm 0.000}{3pt}}$ & $1.099_{\scaleto{\pm 0.000}{3pt}}$ & $1.404_{\scaleto{\pm 0.000}{3pt}}$ & $1.403_{\scaleto{\pm 0.000}{3pt}}$ & $1.402_{\scaleto{\pm 0.000}{3pt}}$\\ \hline 
\multicolumn{2}{c|}{\multirow{2}{*}{MAPLS}}  & Near & $0.700_{\scaleto{\pm 0.034}{3pt}}$ & $0.114_{\scaleto{\pm 0.018}{3pt}}$ & $0.091_{\scaleto{\pm 0.019}{3pt}}$ & $0.878_{\scaleto{\pm 0.037}{3pt}}$ & $0.164_{\scaleto{\pm 0.033}{3pt}}$ & $0.120_{\scaleto{\pm 0.019}{3pt}}$ & $0.946_{\scaleto{\pm 0.041}{3pt}}$ & $0.175_{\scaleto{\pm 0.031}{3pt}}$ & $0.135_{\scaleto{\pm 0.026}{3pt}}$\\ 
\multicolumn{2}{c|}{} & Far &  $7.469_{\scaleto{\pm 1.122}{3pt}}$ & $0.290_{\scaleto{\pm 0.040}{3pt}}$ & $0.094_{\scaleto{\pm 0.018}{3pt}}$ & $7.758_{\scaleto{\pm 1.196}{3pt}}$ & $0.340_{\scaleto{\pm 0.029}{3pt}}$ & $0.123_{\scaleto{\pm 0.020}{3pt}}$ & $7.779_{\scaleto{\pm 1.171}{3pt}}$ & $0.350_{\scaleto{\pm 0.046}{3pt}}$ & $0.138_{\scaleto{\pm 0.026}{3pt}}$\\ \hline 
\hline 
\multicolumn{12}{c}{Open Set Label Shift estimation models} \\ \hline\hline 
\multicolumn{3}{c||}{Baseline} & $0.426_{\scaleto{\pm 0.000}{3pt}}$ & $0.426_{\scaleto{\pm 0.000}{3pt}}$ & $0.426_{\scaleto{\pm 0.000}{3pt}}$ & $1.101_{\scaleto{\pm 0.000}{3pt}}$ & $1.101_{\scaleto{\pm 0.000}{3pt}}$ & $1.101_{\scaleto{\pm 0.000}{3pt}}$ & $1.405_{\scaleto{\pm 0.000}{3pt}}$ & $1.405_{\scaleto{\pm 0.000}{3pt}}$ & $1.405_{\scaleto{\pm 0.000}{3pt}}$\\ \hline 
\multirow{10}{*}{\textbf{ours}} & \multirow{2}{*}{$\gamma=0.1$}  &  Near &  \cc $\mathbf{0.396}_{\scaleto{\pm 0.018}{3pt}}$ & \cc $\mathbf{0.043}_{\scaleto{\pm 0.004}{3pt}}$ & \cc $\mathbf{0.046}_{\scaleto{\pm 0.009}{3pt}}$ & \cc $\mathbf{0.470}_{\scaleto{\pm 0.027}{3pt}}$ & \cc $\mathbf{0.068}_{\scaleto{\pm 0.013}{3pt}}$ & \cc $\mathbf{0.073}_{\scaleto{\pm 0.013}{3pt}}$ & \cc $\mathbf{0.508}_{\scaleto{\pm 0.005}{3pt}}$ & \cc $\mathbf{0.085}_{\scaleto{\pm 0.017}{3pt}}$ & \cc $\mathbf{0.078}_{\scaleto{\pm 0.015}{3pt}}$\\ 
& & Far & $2.152_{\scaleto{\pm 0.396}{3pt}}$ & \cc $\mathbf{0.082}_{\scaleto{\pm 0.008}{3pt}}$ & \cc $\mathbf{0.047}_{\scaleto{\pm 0.009}{3pt}}$ & $2.224_{\scaleto{\pm 0.323}{3pt}}$ & \cc $\mathbf{0.104}_{\scaleto{\pm 0.010}{3pt}}$ & \cc $\mathbf{0.074}_{\scaleto{\pm 0.013}{3pt}}$ & $2.426_{\scaleto{\pm 0.329}{3pt}}$ & \cc $\mathbf{0.118}_{\scaleto{\pm 0.012}{3pt}}$ & \cc $\mathbf{0.079}_{\scaleto{\pm 0.015}{3pt}}$\\ \cline{2-12}
 & \multirow{2}{*}{$\gamma=0.2$}  &  Near &  $0.473_{\scaleto{\pm 0.008}{3pt}}$ & \cc $\mathbf{0.039}_{\scaleto{\pm 0.005}{3pt}}$ & \cc $\mathbf{0.032}_{\scaleto{\pm 0.001}{3pt}}$ & \cc $\mathbf{0.573}_{\scaleto{\pm 0.023}{3pt}}$ & \cc $\mathbf{0.062}_{\scaleto{\pm 0.002}{3pt}}$ & \cc $\mathbf{0.051}_{\scaleto{\pm 0.004}{3pt}}$ & \cc $\mathbf{0.589}_{\scaleto{\pm 0.013}{3pt}}$ & \cc $\mathbf{0.063}_{\scaleto{\pm 0.008}{3pt}}$ & \cc $\mathbf{0.057}_{\scaleto{\pm 0.005}{3pt}}$\\ 
& & Far & $3.100_{\scaleto{\pm 0.115}{3pt}}$ & \cc $\mathbf{0.094}_{\scaleto{\pm 0.007}{3pt}}$ & \cc $\mathbf{0.034}_{\scaleto{\pm 0.001}{3pt}}$ & $3.077_{\scaleto{\pm 0.221}{3pt}}$ & \cc $\mathbf{0.111}_{\scaleto{\pm 0.011}{3pt}}$ & \cc $\mathbf{0.053}_{\scaleto{\pm 0.004}{3pt}}$ & $3.144_{\scaleto{\pm 0.221}{3pt}}$ & \cc $\mathbf{0.118}_{\scaleto{\pm 0.009}{3pt}}$ & \cc $\mathbf{0.058}_{\scaleto{\pm 0.005}{3pt}}$\\ \cline{2-12}
 & \multirow{2}{*}{$\gamma=0.3$}  &  Near &  $0.480_{\scaleto{\pm 0.031}{3pt}}$ & \cc $\mathbf{0.034}_{\scaleto{\pm 0.002}{3pt}}$ & \cc $\mathbf{0.034}_{\scaleto{\pm 0.002}{3pt}}$ & \cc $\mathbf{0.543}_{\scaleto{\pm 0.022}{3pt}}$ & \cc $\mathbf{0.062}_{\scaleto{\pm 0.008}{3pt}}$ & \cc $\mathbf{0.055}_{\scaleto{\pm 0.008}{3pt}}$ & \cc $\mathbf{0.601}_{\scaleto{\pm 0.016}{3pt}}$ & \cc $\mathbf{0.057}_{\scaleto{\pm 0.008}{3pt}}$ & \cc $\mathbf{0.056}_{\scaleto{\pm 0.011}{3pt}}$\\ 
& & Far & $3.069_{\scaleto{\pm 0.082}{3pt}}$ & \cc $\mathbf{0.089}_{\scaleto{\pm 0.004}{3pt}}$ & \cc $\mathbf{0.036}_{\scaleto{\pm 0.002}{3pt}}$ & $3.268_{\scaleto{\pm 0.238}{3pt}}$ & \cc $\mathbf{0.116}_{\scaleto{\pm 0.010}{3pt}}$ & \cc $\mathbf{0.057}_{\scaleto{\pm 0.008}{3pt}}$ & $3.236_{\scaleto{\pm 0.104}{3pt}}$ & \cc $\mathbf{0.114}_{\scaleto{\pm 0.008}{3pt}}$ & \cc $\mathbf{0.057}_{\scaleto{\pm 0.012}{3pt}}$\\ \cline{2-12}
& \multirow{2}{*}{$\gamma=0.4$}  &  Near &  $0.482_{\scaleto{\pm 0.039}{3pt}}$ & \cc $\mathbf{0.035}_{\scaleto{\pm 0.002}{3pt}}$ & \cc $\mathbf{0.032}_{\scaleto{\pm 0.004}{3pt}}$ & \cc $\mathbf{0.571}_{\scaleto{\pm 0.027}{3pt}}$ & \cc $\mathbf{0.067}_{\scaleto{\pm 0.012}{3pt}}$ & \cc $\mathbf{0.056}_{\scaleto{\pm 0.010}{3pt}}$ & \cc $\mathbf{0.609}_{\scaleto{\pm 0.027}{3pt}}$ & \cc $\mathbf{0.062}_{\scaleto{\pm 0.009}{3pt}}$ & \cc $\mathbf{0.053}_{\scaleto{\pm 0.003}{3pt}}$\\ 
& & Far & $3.093_{\scaleto{\pm 0.068}{3pt}}$ & \cc $\mathbf{0.100}_{\scaleto{\pm 0.007}{3pt}}$ & \cc $\mathbf{0.034}_{\scaleto{\pm 0.004}{3pt}}$ & $3.272_{\scaleto{\pm 0.117}{3pt}}$ & \cc $\mathbf{0.123}_{\scaleto{\pm 0.010}{3pt}}$ & \cc $\mathbf{0.057}_{\scaleto{\pm 0.010}{3pt}}$ & $3.314_{\scaleto{\pm 0.076}{3pt}}$ & \cc $\mathbf{0.123}_{\scaleto{\pm 0.002}{3pt}}$ & \cc $\mathbf{0.054}_{\scaleto{\pm 0.002}{3pt}}$\\ \cline{2-12}
 & \multirow{2}{*}{$\gamma=0.5$}  &  Near &  $0.486_{\scaleto{\pm 0.025}{3pt}}$ & \cc $\mathbf{0.041}_{\scaleto{\pm 0.003}{3pt}}$ & \cc $\mathbf{0.032}_{\scaleto{\pm 0.004}{3pt}}$ & \cc $\mathbf{0.578}_{\scaleto{\pm 0.017}{3pt}}$ & \cc $\mathbf{0.063}_{\scaleto{\pm 0.013}{3pt}}$ & \cc $\mathbf{0.055}_{\scaleto{\pm 0.007}{3pt}}$ & \cc $\mathbf{0.598}_{\scaleto{\pm 0.028}{3pt}}$ & \cc $\mathbf{0.060}_{\scaleto{\pm 0.007}{3pt}}$ & \cc $\mathbf{0.055}_{\scaleto{\pm 0.008}{3pt}}$\\ 
& & Far & $3.135_{\scaleto{\pm 0.155}{3pt}}$ & \cc $\mathbf{0.102}_{\scaleto{\pm 0.007}{3pt}}$ & \cc $\mathbf{0.033}_{\scaleto{\pm 0.003}{3pt}}$ & $3.209_{\scaleto{\pm 0.135}{3pt}}$ & \cc $\mathbf{0.131}_{\scaleto{\pm 0.010}{3pt}}$ & \cc $\mathbf{0.057}_{\scaleto{\pm 0.008}{3pt}}$ & $3.335_{\scaleto{\pm 0.123}{3pt}}$ & \cc $\mathbf{0.115}_{\scaleto{\pm 0.008}{3pt}}$ & \cc $\mathbf{0.056}_{\scaleto{\pm 0.008}{3pt}}$\\ \hline

         \hline
    \end{tabular} \vspace{-1em}
    \caption{\textbf{Ablation study of hyperparameter $\gamma$ when generating pseudo OOD samples.} Estimation Error $(w-\hat{w})^2/K(\downarrow)$ of our OSLS estimation model (OpenMax OOD detector) on CIFAR100 dataset with Near OOD datasets and Far OOD datasets comparison under Ordered-LT (Forward) ID and OOD label shift.}
    \label{Atab:wmse-cifar100-LT-gamma-ablation}
\vspace{-1em} \end{table*}
\setlength\tabcolsep{6pt}

\setlength\tabcolsep{0.8pt} 
\begin{table*}[ht]
    \centering
    \scriptsize
    \renewcommand{\arraystretch}{1.1}
    \begin{tabular}{c | c | c||c c c | c c c | c c c } \hline\hline
    \multicolumn{3}{c||}{Dataset} & \multicolumn{9}{c}{CIFAR100} \\ \hline
       \multicolumn{3}{c||}{ID label Shift param}  & \multicolumn{3}{c|}{LT10 Backward} & \multicolumn{3}{c|}{LT50 Backward} & \multicolumn{3}{c}{LT100 Backward} \\ \hline
      \multicolumn{3}{c||}{OOD label shift param $r$} & $1.0$ & $0.1$ & $0.01$  & $1.0$ & $0.1$ & $0.01$  & $1.0$ & $0.1$ & $0.01$ \\ \hline \hline

\multicolumn{12}{c}{Closed Set Label Shift estimation models} \\ \hline\hline 
\multicolumn{2}{c|}{\multirow{2}{*}{BBSE}}  & Near & $0.540_{\scaleto{\pm 0.029}{3pt}}$ & $0.152_{\scaleto{\pm 0.008}{3pt}}$ & $0.159_{\scaleto{\pm 0.025}{3pt}}$ & $0.732_{\scaleto{\pm 0.050}{3pt}}$ & $0.252_{\scaleto{\pm 0.027}{3pt}}$ & $0.264_{\scaleto{\pm 0.035}{3pt}}$ & $0.778_{\scaleto{\pm 0.026}{3pt}}$ & $0.281_{\scaleto{\pm 0.050}{3pt}}$ & $0.339_{\scaleto{\pm 0.073}{3pt}}$\\ 
\multicolumn{2}{c|}{} & Far &  $4.042_{\scaleto{\pm 0.273}{3pt}}$ & $0.276_{\scaleto{\pm 0.011}{3pt}}$ & $0.161_{\scaleto{\pm 0.023}{3pt}}$ & $4.075_{\scaleto{\pm 0.388}{3pt}}$ & $0.381_{\scaleto{\pm 0.049}{3pt}}$ & $0.262_{\scaleto{\pm 0.037}{3pt}}$ & $4.080_{\scaleto{\pm 0.223}{3pt}}$ & $0.387_{\scaleto{\pm 0.056}{3pt}}$ & $0.339_{\scaleto{\pm 0.077}{3pt}}$\\ \hline 
\multicolumn{2}{c|}{\multirow{2}{*}{MLLS}}  & Near & $0.912_{\scaleto{\pm 0.083}{3pt}}$ & $0.131_{\scaleto{\pm 0.013}{3pt}}$ & $0.119_{\scaleto{\pm 0.009}{3pt}}$ & $1.107_{\scaleto{\pm 0.085}{3pt}}$ & $0.203_{\scaleto{\pm 0.012}{3pt}}$ & $0.173_{\scaleto{\pm 0.017}{3pt}}$ & $1.152_{\scaleto{\pm 0.061}{3pt}}$ & $0.218_{\scaleto{\pm 0.017}{3pt}}$ & $0.203_{\scaleto{\pm 0.022}{3pt}}$\\ 
\multicolumn{2}{c|}{} & Far &  $9.500_{\scaleto{\pm 1.553}{3pt}}$ & $0.332_{\scaleto{\pm 0.036}{3pt}}$ & $0.118_{\scaleto{\pm 0.008}{3pt}}$ & $9.583_{\scaleto{\pm 1.578}{3pt}}$ & $0.404_{\scaleto{\pm 0.058}{3pt}}$ & $0.167_{\scaleto{\pm 0.017}{3pt}}$ & $9.494_{\scaleto{\pm 1.499}{3pt}}$ & $0.381_{\scaleto{\pm 0.039}{3pt}}$ & $0.201_{\scaleto{\pm 0.024}{3pt}}$\\ \hline 
\multicolumn{2}{c|}{\multirow{2}{*}{RLLS}}  & Near & $0.426_{\scaleto{\pm 0.000}{3pt}}$ & $0.425_{\scaleto{\pm 0.000}{3pt}}$ & $0.425_{\scaleto{\pm 0.000}{3pt}}$ & $1.100_{\scaleto{\pm 0.000}{3pt}}$ & $1.099_{\scaleto{\pm 0.000}{3pt}}$ & $1.099_{\scaleto{\pm 0.000}{3pt}}$ & $1.403_{\scaleto{\pm 0.000}{3pt}}$ & $1.402_{\scaleto{\pm 0.000}{3pt}}$ & $1.402_{\scaleto{\pm 0.000}{3pt}}$\\ 
\multicolumn{2}{c|}{} & Far &  $0.426_{\scaleto{\pm 0.000}{3pt}}$ & $0.425_{\scaleto{\pm 0.000}{3pt}}$ & $0.425_{\scaleto{\pm 0.000}{3pt}}$ & $1.100_{\scaleto{\pm 0.000}{3pt}}$ & $1.099_{\scaleto{\pm 0.000}{3pt}}$ & $1.099_{\scaleto{\pm 0.000}{3pt}}$ & $1.403_{\scaleto{\pm 0.000}{3pt}}$ & $1.402_{\scaleto{\pm 0.000}{3pt}}$ & $1.402_{\scaleto{\pm 0.000}{3pt}}$\\ \hline 
\multicolumn{2}{c|}{\multirow{2}{*}{MAPLS}}  & Near & $0.710_{\scaleto{\pm 0.052}{3pt}}$ & $0.119_{\scaleto{\pm 0.007}{3pt}}$ & $0.106_{\scaleto{\pm 0.003}{3pt}}$ & $0.941_{\scaleto{\pm 0.063}{3pt}}$ & $0.196_{\scaleto{\pm 0.008}{3pt}}$ & $0.159_{\scaleto{\pm 0.009}{3pt}}$ & $1.007_{\scaleto{\pm 0.044}{3pt}}$ & $0.218_{\scaleto{\pm 0.012}{3pt}}$ & $0.188_{\scaleto{\pm 0.012}{3pt}}$\\ 
\multicolumn{2}{c|}{} & Far &  $7.360_{\scaleto{\pm 1.206}{3pt}}$ & $0.268_{\scaleto{\pm 0.025}{3pt}}$ & $0.106_{\scaleto{\pm 0.002}{3pt}}$ & $7.476_{\scaleto{\pm 1.220}{3pt}}$ & $0.345_{\scaleto{\pm 0.039}{3pt}}$ & $0.155_{\scaleto{\pm 0.009}{3pt}}$ & $7.439_{\scaleto{\pm 1.153}{3pt}}$ & $0.339_{\scaleto{\pm 0.023}{3pt}}$ & $0.186_{\scaleto{\pm 0.012}{3pt}}$\\ \hline 
\hline 
\multicolumn{12}{c}{Open Set Label Shift estimation models} \\ \hline\hline 
\multicolumn{3}{c||}{Baseline} & $0.426_{\scaleto{\pm 0.000}{3pt}}$ & $0.426_{\scaleto{\pm 0.000}{3pt}}$ & $0.426_{\scaleto{\pm 0.000}{3pt}}$ & $1.101_{\scaleto{\pm 0.000}{3pt}}$ & $1.101_{\scaleto{\pm 0.000}{3pt}}$ & $1.101_{\scaleto{\pm 0.000}{3pt}}$ & $1.405_{\scaleto{\pm 0.000}{3pt}}$ & $1.405_{\scaleto{\pm 0.000}{3pt}}$ & $1.405_{\scaleto{\pm 0.000}{3pt}}$\\ \hline 
\multirow{10}{*}{\textbf{ours}} & \multirow{2}{*}{$\gamma=0.1$}  &  Near &  $0.428_{\scaleto{\pm 0.046}{3pt}}$ & \cc $\mathbf{0.041}_{\scaleto{\pm 0.002}{3pt}}$ & \cc $\mathbf{0.034}_{\scaleto{\pm 0.004}{3pt}}$ & \cc $\mathbf{0.565}_{\scaleto{\pm 0.076}{3pt}}$ & \cc $\mathbf{0.058}_{\scaleto{\pm 0.001}{3pt}}$ & \cc $\mathbf{0.056}_{\scaleto{\pm 0.006}{3pt}}$ & \cc $\mathbf{0.565}_{\scaleto{\pm 0.028}{3pt}}$ & \cc $\mathbf{0.063}_{\scaleto{\pm 0.001}{3pt}}$ & \cc $\mathbf{0.052}_{\scaleto{\pm 0.003}{3pt}}$\\ 
& & Far & $2.105_{\scaleto{\pm 0.489}{3pt}}$ & \cc $\mathbf{0.080}_{\scaleto{\pm 0.007}{3pt}}$ & \cc $\mathbf{0.035}_{\scaleto{\pm 0.003}{3pt}}$ & $2.244_{\scaleto{\pm 0.372}{3pt}}$ & \cc $\mathbf{0.093}_{\scaleto{\pm 0.004}{3pt}}$ & \cc $\mathbf{0.058}_{\scaleto{\pm 0.007}{3pt}}$ & $2.192_{\scaleto{\pm 0.320}{3pt}}$ & \cc $\mathbf{0.100}_{\scaleto{\pm 0.015}{3pt}}$ & \cc $\mathbf{0.054}_{\scaleto{\pm 0.003}{3pt}}$\\ \cline{2-12}
 & \multirow{2}{*}{$\gamma=0.2$}  &  Near &  $0.497_{\scaleto{\pm 0.025}{3pt}}$ & \cc $\mathbf{0.034}_{\scaleto{\pm 0.004}{3pt}}$ & \cc $\mathbf{0.028}_{\scaleto{\pm 0.001}{3pt}}$ & \cc $\mathbf{0.628}_{\scaleto{\pm 0.017}{3pt}}$ & \cc $\mathbf{0.059}_{\scaleto{\pm 0.008}{3pt}}$ & \cc $\mathbf{0.049}_{\scaleto{\pm 0.005}{3pt}}$ & \cc $\mathbf{0.664}_{\scaleto{\pm 0.025}{3pt}}$ & \cc $\mathbf{0.058}_{\scaleto{\pm 0.007}{3pt}}$ & \cc $\mathbf{0.046}_{\scaleto{\pm 0.005}{3pt}}$\\ 
& & Far & $2.879_{\scaleto{\pm 0.178}{3pt}}$ & \cc $\mathbf{0.093}_{\scaleto{\pm 0.007}{3pt}}$ & \cc $\mathbf{0.029}_{\scaleto{\pm 0.001}{3pt}}$ & $2.855_{\scaleto{\pm 0.192}{3pt}}$ & \cc $\mathbf{0.115}_{\scaleto{\pm 0.019}{3pt}}$ & \cc $\mathbf{0.050}_{\scaleto{\pm 0.004}{3pt}}$ & $2.879_{\scaleto{\pm 0.215}{3pt}}$ & \cc $\mathbf{0.115}_{\scaleto{\pm 0.010}{3pt}}$ & \cc $\mathbf{0.047}_{\scaleto{\pm 0.005}{3pt}}$\\ \cline{2-12}
 & \multirow{2}{*}{$\gamma=0.3$}  &  Near &  $0.539_{\scaleto{\pm 0.012}{3pt}}$ & \cc $\mathbf{0.034}_{\scaleto{\pm 0.004}{3pt}}$ & \cc $\mathbf{0.029}_{\scaleto{\pm 0.006}{3pt}}$ & \cc $\mathbf{0.662}_{\scaleto{\pm 0.014}{3pt}}$ & \cc $\mathbf{0.060}_{\scaleto{\pm 0.010}{3pt}}$ & \cc $\mathbf{0.048}_{\scaleto{\pm 0.003}{3pt}}$ & \cc $\mathbf{0.688}_{\scaleto{\pm 0.021}{3pt}}$ & \cc $\mathbf{0.066}_{\scaleto{\pm 0.007}{3pt}}$ & \cc $\mathbf{0.053}_{\scaleto{\pm 0.003}{3pt}}$\\ 
& & Far & $3.000_{\scaleto{\pm 0.185}{3pt}}$ & \cc $\mathbf{0.092}_{\scaleto{\pm 0.015}{3pt}}$ & \cc $\mathbf{0.030}_{\scaleto{\pm 0.006}{3pt}}$ & $2.987_{\scaleto{\pm 0.171}{3pt}}$ & \cc $\mathbf{0.113}_{\scaleto{\pm 0.006}{3pt}}$ & \cc $\mathbf{0.050}_{\scaleto{\pm 0.003}{3pt}}$ & $3.025_{\scaleto{\pm 0.235}{3pt}}$ & \cc $\mathbf{0.114}_{\scaleto{\pm 0.022}{3pt}}$ & \cc $\mathbf{0.055}_{\scaleto{\pm 0.003}{3pt}}$\\ \cline{2-12}
& \multirow{2}{*}{$\gamma=0.4$}  &  Near &  $0.534_{\scaleto{\pm 0.001}{3pt}}$ & \cc $\mathbf{0.037}_{\scaleto{\pm 0.003}{3pt}}$ & \cc $\mathbf{0.027}_{\scaleto{\pm 0.002}{3pt}}$ & \cc $\mathbf{0.669}_{\scaleto{\pm 0.024}{3pt}}$ & \cc $\mathbf{0.059}_{\scaleto{\pm 0.002}{3pt}}$ & \cc $\mathbf{0.049}_{\scaleto{\pm 0.001}{3pt}}$ & \cc $\mathbf{0.696}_{\scaleto{\pm 0.019}{3pt}}$ & \cc $\mathbf{0.063}_{\scaleto{\pm 0.009}{3pt}}$ & \cc $\mathbf{0.046}_{\scaleto{\pm 0.002}{3pt}}$\\ 
& & Far & $3.001_{\scaleto{\pm 0.144}{3pt}}$ & \cc $\mathbf{0.101}_{\scaleto{\pm 0.009}{3pt}}$ & \cc $\mathbf{0.028}_{\scaleto{\pm 0.003}{3pt}}$ & $3.117_{\scaleto{\pm 0.085}{3pt}}$ & \cc $\mathbf{0.113}_{\scaleto{\pm 0.014}{3pt}}$ & \cc $\mathbf{0.050}_{\scaleto{\pm 0.001}{3pt}}$ & $3.023_{\scaleto{\pm 0.134}{3pt}}$ & \cc $\mathbf{0.131}_{\scaleto{\pm 0.016}{3pt}}$ & \cc $\mathbf{0.048}_{\scaleto{\pm 0.003}{3pt}}$\\ \cline{2-12}
 & \multirow{2}{*}{$\gamma=0.5$}  &  Near &  $0.526_{\scaleto{\pm 0.018}{3pt}}$ & \cc $\mathbf{0.035}_{\scaleto{\pm 0.002}{3pt}}$ & \cc $\mathbf{0.030}_{\scaleto{\pm 0.004}{3pt}}$ & \cc $\mathbf{0.647}_{\scaleto{\pm 0.005}{3pt}}$ & \cc $\mathbf{0.052}_{\scaleto{\pm 0.004}{3pt}}$ & \cc $\mathbf{0.054}_{\scaleto{\pm 0.005}{3pt}}$ & \cc $\mathbf{0.693}_{\scaleto{\pm 0.017}{3pt}}$ & \cc $\mathbf{0.060}_{\scaleto{\pm 0.002}{3pt}}$ & \cc $\mathbf{0.044}_{\scaleto{\pm 0.004}{3pt}}$\\ 
& & Far & $3.078_{\scaleto{\pm 0.159}{3pt}}$ & \cc $\mathbf{0.092}_{\scaleto{\pm 0.017}{3pt}}$ & \cc $\mathbf{0.032}_{\scaleto{\pm 0.003}{3pt}}$ & $3.046_{\scaleto{\pm 0.153}{3pt}}$ & \cc $\mathbf{0.122}_{\scaleto{\pm 0.018}{3pt}}$ & \cc $\mathbf{0.056}_{\scaleto{\pm 0.006}{3pt}}$ & $3.021_{\scaleto{\pm 0.160}{3pt}}$ & \cc $\mathbf{0.118}_{\scaleto{\pm 0.021}{3pt}}$ & \cc $\mathbf{0.046}_{\scaleto{\pm 0.005}{3pt}}$\\ \hline

         \hline
    \end{tabular} \vspace{-1em}
    \caption{\textbf{Ablation study of hyperparameter $\gamma$ when generating pseudo OOD samples.} Estimation Error $(w-\hat{w})^2/K(\downarrow)$ of our OSLS estimation model (OpenMax OOD detector) on the CIFAR100 dataset with Near OOD datasets and Far OOD datasets comparison under Ordered-LT (Backward) ID and OOD label shift.}
    \label{Atab:wmse-cifar100-LTr-gamma-ablation}
\vspace{-1em} \end{table*}
\setlength\tabcolsep{6pt}
\clearpage
\section{More Estimation Error Results}\label{Asec:detailed-wmse}

We provide the full open set label shift estimation error on the CIFAR10, CIFAR100 datasets. The results show that our model consistently performs better than the open set label shift baseline and CSLS estimation models on most of the tested dataset setups.

\subsection{CIFAR10}

\setlength\tabcolsep{0.8pt} 
\begin{table*}[ht]
    \centering
    \scriptsize
    \renewcommand{\arraystretch}{1.1}
    \begin{tabular}{c | c | c||c c c | c c c | c c c } \hline\hline
    \multicolumn{3}{c||}{Dataset} & \multicolumn{9}{c}{CIFAR10} \\ \hline
       \multicolumn{3}{c||}{ID label Shift param}  & \multicolumn{3}{c|}{LT10 Forward} & \multicolumn{3}{c|}{LT50 Forward} & \multicolumn{3}{c}{LT100 Forward} \\ \hline
      \multicolumn{3}{c||}{OOD label shift param $r$} & $1.0$ & $0.1$ & $0.01$  & $1.0$ & $0.1$ & $0.01$  & $1.0$ & $0.1$ & $0.01$ \\ \hline \hline

\multicolumn{12}{c}{Closed Set Label Shift estimation models} \\ \hline\hline 
\multicolumn{2}{c|}{\multirow{2}{*}{BBSE}}  & Near & $0.172_{\scaleto{\pm 0.014}{3pt}}$ & $0.005_{\scaleto{\pm 0.001}{3pt}}$ & $0.003_{\scaleto{\pm 0.001}{3pt}}$ & $0.341_{\scaleto{\pm 0.025}{3pt}}$ & $0.008_{\scaleto{\pm 0.000}{3pt}}$ & $0.007_{\scaleto{\pm 0.003}{3pt}}$ & $0.418_{\scaleto{\pm 0.025}{3pt}}$ & $0.011_{\scaleto{\pm 0.001}{3pt}}$ & $0.010_{\scaleto{\pm 0.002}{3pt}}$\\ 
\multicolumn{2}{c|}{} & Far &  $0.400_{\scaleto{\pm 0.012}{3pt}}$ & $0.014_{\scaleto{\pm 0.002}{3pt}}$ & $0.004_{\scaleto{\pm 0.001}{3pt}}$ & $0.547_{\scaleto{\pm 0.009}{3pt}}$ & $0.016_{\scaleto{\pm 0.001}{3pt}}$ & $0.008_{\scaleto{\pm 0.003}{3pt}}$ & $0.629_{\scaleto{\pm 0.006}{3pt}}$ & $0.018_{\scaleto{\pm 0.002}{3pt}}$ & $0.010_{\scaleto{\pm 0.003}{3pt}}$\\ \hline 
\multicolumn{2}{c|}{\multirow{2}{*}{MLLS}}  & Near & $0.168_{\scaleto{\pm 0.016}{3pt}}$ & $0.005_{\scaleto{\pm 0.000}{3pt}}$ & $0.002_{\scaleto{\pm 0.000}{3pt}}$ & $0.318_{\scaleto{\pm 0.023}{3pt}}$ & $0.007_{\scaleto{\pm 0.001}{3pt}}$ & $0.002_{\scaleto{\pm 0.001}{3pt}}$ & $0.388_{\scaleto{\pm 0.026}{3pt}}$ & $0.010_{\scaleto{\pm 0.001}{3pt}}$ & $0.002_{\scaleto{\pm 0.001}{3pt}}$\\
\hline

\hline

    \end{tabular} \vspace{-1em}
    \caption{\textbf{Estimation Error $(w-\hat{w})^2/K(\downarrow)$ of our OSLS estimation model on the CIFAR10 dataset with Near OOD datasets and Far OOD datasets comparison under Ordered-LT (Forward) ID and OOD label shift. Settings in which our model outperforms baselines are colored in gray.} Our model outperforms baselines under most label shift settings. Each metric is averaged among corresponding the OOD test set (Tab.~\ref{tab:dataset-setup}) and over three independent ID classifiers.}
    \label{Atab:wmse-cifar10-LT}
\vspace{-1em} \end{table*}
\setlength\tabcolsep{6pt}

\setlength\tabcolsep{0.8pt} 
\begin{table*}[ht]
    \centering
    \scriptsize
    \renewcommand{\arraystretch}{1.1}
    \begin{tabular}{c | c | c||c c c | c c c | c c c } \hline\hline
    \multicolumn{3}{c||}{Dataset} & \multicolumn{9}{c}{CIFAR10} \\ \hline
       \multicolumn{3}{c||}{ID label Shift param}  & \multicolumn{3}{c|}{LT10 Backward} & \multicolumn{3}{c|}{LT50 Backward} & \multicolumn{3}{c}{LT100 Backward} \\ \hline
      \multicolumn{3}{c||}{OOD label shift param $r$} & $1.0$ & $0.1$ & $0.01$  & $1.0$ & $0.1$ & $0.01$  & $1.0$ & $0.1$ & $0.01$ \\ \hline \hline

\multicolumn{12}{c}{Closed Set Label Shift estimation models} \\ \hline\hline 
\multicolumn{2}{c|}{\multirow{2}{*}{BBSE}}  & Near & $0.232_{\scaleto{\pm 0.023}{3pt}}$ & $0.008_{\scaleto{\pm 0.003}{3pt}}$ & $0.001_{\scaleto{\pm 0.000}{3pt}}$ & $0.433_{\scaleto{\pm 0.034}{3pt}}$ & $0.017_{\scaleto{\pm 0.002}{3pt}}$ & $0.002_{\scaleto{\pm 0.001}{3pt}}$ & $0.508_{\scaleto{\pm 0.029}{3pt}}$ & $0.021_{\scaleto{\pm 0.002}{3pt}}$ & $0.002_{\scaleto{\pm 0.001}{3pt}}$\\ 
\multicolumn{2}{c|}{} & Far &  $0.568_{\scaleto{\pm 0.032}{3pt}}$ & $0.020_{\scaleto{\pm 0.003}{3pt}}$ & $0.001_{\scaleto{\pm 0.000}{3pt}}$ & $0.796_{\scaleto{\pm 0.030}{3pt}}$ & $0.029_{\scaleto{\pm 0.003}{3pt}}$ & $0.002_{\scaleto{\pm 0.001}{3pt}}$ & $0.905_{\scaleto{\pm 0.037}{3pt}}$ & $0.034_{\scaleto{\pm 0.002}{3pt}}$ & $0.002_{\scaleto{\pm 0.001}{3pt}}$\\ \hline 
\multicolumn{2}{c|}{\multirow{2}{*}{MLLS}}  & Near & $0.222_{\scaleto{\pm 0.030}{3pt}}$ & $0.009_{\scaleto{\pm 0.004}{3pt}}$ & $0.002_{\scaleto{\pm 0.001}{3pt}}$ & $0.394_{\scaleto{\pm 0.049}{3pt}}$ & $0.016_{\scaleto{\pm 0.005}{3pt}}$ & $0.003_{\scaleto{\pm 0.001}{3pt}}$ & $0.455_{\scaleto{\pm 0.045}{3pt}}$ & $0.019_{\scaleto{\pm 0.006}{3pt}}$ & $0.003_{\scaleto{\pm 0.002}{3pt}}$\\
\hline

         \hline
    \end{tabular} \vspace{-1em}
    \caption{\textbf{Estimation Error $(w-\hat{w})^2/K(\downarrow)$ of our OSLS estimation model on the CIFAR10 dataset with Near OOD datasets and Far OOD datasets comparison under Ordered-LT (Backward) ID and OOD label shift. Settings in which our model outperforms baselines are colored in gray.} Our model outperforms baselines under most label shift settings. Each metric is averaged among corresponding OOD test set (Tab.~\ref{tab:dataset-setup}) and over three independent ID classifiers.}
    \label{Atab:wmse-cifar10-LTr}
\vspace{-1em} \end{table*}
\setlength\tabcolsep{6pt}

\setlength\tabcolsep{0.8pt} 
\begin{table*}[ht]
    \centering
    \scriptsize
    \renewcommand{\arraystretch}{1.1}
    \begin{tabular}{c | c | c||c c c | c c c } \hline\hline 
    \multicolumn{3}{c||}{Dataset} & \multicolumn{6}{c}{CIFAR10} \\ \hline
       \multicolumn{3}{c||}{ID label Shift param} & \multicolumn{3}{c|}{Dir $\alpha=1.0$} & \multicolumn{3}{c}{Dir $\alpha=10.0$} \\ \hline
      \multicolumn{3}{c||}{OOD label shift param $r$} & $1.0$ & $0.1$ & $0.01$  & $1.0$ & $0.1$ & $0.01$  \\ \hline

\multicolumn{9}{c}{Closed Set Label Shift estimation models} \\ \hline\hline 
\multicolumn{2}{c|}{\multirow{2}{*}{BBSE}}  & Near & $0.309_{\scaleto{\pm 0.103}{3pt}}$ & $0.008_{\scaleto{\pm 0.002}{3pt}}$ & $0.001_{\scaleto{\pm 0.000}{3pt}}$ & $0.079_{\scaleto{\pm 0.039}{3pt}}$ & $0.003_{\scaleto{\pm 0.001}{3pt}}$ & $0.002_{\scaleto{\pm 0.001}{3pt}}$\\ 
\multicolumn{2}{c|}{} & Far &  $0.570_{\scaleto{\pm 0.114}{3pt}}$ & $0.016_{\scaleto{\pm 0.005}{3pt}}$ & $0.001_{\scaleto{\pm 0.000}{3pt}}$ & $0.366_{\scaleto{\pm 0.087}{3pt}}$ & $0.012_{\scaleto{\pm 0.001}{3pt}}$ & $0.002_{\scaleto{\pm 0.001}{3pt}}$\\ \hline 
\multicolumn{2}{c|}{\multirow{2}{*}{MLLS}}  & Near & $0.280_{\scaleto{\pm 0.082}{3pt}}$ & $0.010_{\scaleto{\pm 0.004}{3pt}}$ & $0.002_{\scaleto{\pm 0.001}{3pt}}$ & $0.088_{\scaleto{\pm 0.040}{3pt}}$ & $0.003_{\scaleto{\pm 0.001}{3pt}}$ & $0.001_{\scaleto{\pm 0.000}{3pt}}$\\ 
\multicolumn{2}{c|}{} & Far &  $0.616_{\scaleto{\pm 0.073}{3pt}}$ & $0.016_{\scaleto{\pm 0.005}{3pt}}$ & $0.002_{\scaleto{\pm 0.001}{3pt}}$ & $0.439_{\scaleto{\pm 0.074}{3pt}}$ & $0.012_{\scaleto{\pm 0.000}{3pt}}$ & $0.001_{\scaleto{\pm 0.000}{3pt}}$\\
\hline\hline
    \end{tabular} \vspace{-1em}
    \caption{\textbf{Estimation Error $(w-\hat{w})^2/K(\downarrow)$ of our OSLS estimation model on the CIFAR10 dataset with Near OOD datasets and Far OOD datasets comparison under Dirichlet ID and OOD label shift. Settings in which our model outperforms baselines are colored in gray.} Our model outperforms baselines under most label shift settings. Each metric is averaged among the corresponding OOD test set (Tab.~\ref{tab:dataset-setup}) and over three independent ID classifiers.}
    \label{Atab:acc-cifar10-dir}
\vspace{-1em} \end{table*}
\setlength\tabcolsep{6pt}

\clearpage
\subsection{CIFAR100}

\setlength\tabcolsep{0.8pt} 
\begin{table*}[ht]
    \centering
    \scriptsize
    \renewcommand{\arraystretch}{1.1}
    \begin{tabular}{c | c | c||c c c | c c c | c c c } \hline\hline
    \multicolumn{3}{c||}{Dataset} & \multicolumn{9}{c}{CIFAR100} \\ \hline
       \multicolumn{3}{c||}{ID label Shift param}  & \multicolumn{3}{c|}{LT10 Forward} & \multicolumn{3}{c|}{LT50 Forward} & \multicolumn{3}{c}{LT100 Forward} \\ \hline
      \multicolumn{3}{c||}{OOD label shift param $r$} & $1.0$ & $0.1$ & $0.01$  & $1.0$ & $0.1$ & $0.01$  & $1.0$ & $0.1$ & $0.01$ \\ \hline \hline
\multicolumn{12}{c}{Closed Set Label Shift estimation models} \\ \hline\hline 
\multicolumn{2}{c|}{\multirow{2}{*}{BBSE}}  & Near & $0.529_{\scaleto{\pm 0.017}{3pt}}$ & $0.131_{\scaleto{\pm 0.030}{3pt}}$ & $0.097_{\scaleto{\pm 0.021}{3pt}}$ & $0.750_{\scaleto{\pm 0.033}{3pt}}$ & $0.173_{\scaleto{\pm 0.010}{3pt}}$ & $0.176_{\scaleto{\pm 0.058}{3pt}}$ & $0.850_{\scaleto{\pm 0.052}{3pt}}$ & $0.173_{\scaleto{\pm 0.041}{3pt}}$ & $0.167_{\scaleto{\pm 0.054}{3pt}}$\\ 
\multicolumn{2}{c|}{} & Far &  $4.118_{\scaleto{\pm 0.263}{3pt}}$ & $0.250_{\scaleto{\pm 0.037}{3pt}}$ & $0.099_{\scaleto{\pm 0.021}{3pt}}$ & $4.362_{\scaleto{\pm 0.245}{3pt}}$ & $0.290_{\scaleto{\pm 0.017}{3pt}}$ & $0.176_{\scaleto{\pm 0.057}{3pt}}$ & $4.489_{\scaleto{\pm 0.238}{3pt}}$ & $0.294_{\scaleto{\pm 0.039}{3pt}}$ & $0.168_{\scaleto{\pm 0.053}{3pt}}$\\ \hline 
\multicolumn{2}{c|}{\multirow{2}{*}{MLLS}}  & Near & $0.870_{\scaleto{\pm 0.069}{3pt}}$ & $0.116_{\scaleto{\pm 0.019}{3pt}}$ & $0.080_{\scaleto{\pm 0.022}{3pt}}$ & $1.005_{\scaleto{\pm 0.085}{3pt}}$ & $0.142_{\scaleto{\pm 0.017}{3pt}}$ & $0.116_{\scaleto{\pm 0.030}{3pt}}$ & $1.100_{\scaleto{\pm 0.098}{3pt}}$ & $0.132_{\scaleto{\pm 0.034}{3pt}}$ & $0.113_{\scaleto{\pm 0.039}{3pt}}$\\
\hline 

         \hline
    \end{tabular} \vspace{-1em}
    \caption{\textbf{Estimation Error $(w-\hat{w})^2/K(\downarrow)$ of our OSLS estimation model on the CIFAR100 dataset with Near OOD datasets and Far OOD datasets comparison under Ordered-LT (Forward) ID and OOD label shift. Settings in which our model outperforms baselines are colored in gray.} Our model outperforms baselines under most label shift settings. Each metric is averaged among corresponding OOD test set (Tab.~\ref{tab:dataset-setup}) and over three independent ID classifiers.}
    \label{Atab:wmse-cifar100-LT}
\vspace{-1em} \end{table*}
\setlength\tabcolsep{6pt}

\setlength\tabcolsep{0.8pt} 
\begin{table*}[ht]
    \centering
    \scriptsize
    \renewcommand{\arraystretch}{1.1}
    \begin{tabular}{c | c | c||c c c | c c c | c c c } \hline\hline
    \multicolumn{3}{c||}{Dataset} & \multicolumn{9}{c}{CIFAR100} \\ \hline
       \multicolumn{3}{c||}{ID label Shift param}  & \multicolumn{3}{c|}{LT10 Backward} & \multicolumn{3}{c|}{LT50 Backward} & \multicolumn{3}{c}{LT100 Backward} \\ \hline
      \multicolumn{3}{c||}{OOD label shift param $r$} & $1.0$ & $0.1$ & $0.01$  & $1.0$ & $0.1$ & $0.01$  & $1.0$ & $0.1$ & $0.01$ \\ \hline \hline

\multicolumn{12}{c}{Closed Set Label Shift estimation models} \\ \hline\hline 
\multicolumn{2}{c|}{\multirow{2}{*}{BBSE}}  & Near & $0.553_{\scaleto{\pm 0.020}{3pt}}$ & $0.153_{\scaleto{\pm 0.003}{3pt}}$ & $0.162_{\scaleto{\pm 0.016}{3pt}}$ & $0.730_{\scaleto{\pm 0.049}{3pt}}$ & $0.243_{\scaleto{\pm 0.031}{3pt}}$ & $0.244_{\scaleto{\pm 0.022}{3pt}}$ & $0.778_{\scaleto{\pm 0.043}{3pt}}$ & $0.265_{\scaleto{\pm 0.017}{3pt}}$ & $0.295_{\scaleto{\pm 0.039}{3pt}}$\\ 
\multicolumn{2}{c|}{} & Far &  $3.968_{\scaleto{\pm 0.256}{3pt}}$ & $0.276_{\scaleto{\pm 0.014}{3pt}}$ & $0.164_{\scaleto{\pm 0.016}{3pt}}$ & $4.099_{\scaleto{\pm 0.290}{3pt}}$ & $0.351_{\scaleto{\pm 0.030}{3pt}}$ & $0.245_{\scaleto{\pm 0.024}{3pt}}$ & $4.104_{\scaleto{\pm 0.206}{3pt}}$ & $0.373_{\scaleto{\pm 0.028}{3pt}}$ & $0.296_{\scaleto{\pm 0.042}{3pt}}$\\ \hline 
\multicolumn{2}{c|}{\multirow{2}{*}{MLLS}}  & Near & $0.937_{\scaleto{\pm 0.034}{3pt}}$ & $0.140_{\scaleto{\pm 0.016}{3pt}}$ & $0.116_{\scaleto{\pm 0.009}{3pt}}$ & $1.133_{\scaleto{\pm 0.107}{3pt}}$ & $0.208_{\scaleto{\pm 0.021}{3pt}}$ & $0.171_{\scaleto{\pm 0.015}{3pt}}$ & $1.146_{\scaleto{\pm 0.083}{3pt}}$ & $0.212_{\scaleto{\pm 0.038}{3pt}}$ & $0.193_{\scaleto{\pm 0.011}{3pt}}$\\
\hline

         \hline
    \end{tabular} \vspace{-1em}
    \caption{\textbf{Estimation Error $(w-\hat{w})^2/K(\downarrow)$ of our OSLS estimation model on the CIFAR100 dataset with Near OOD datasets and Far OOD datasets comparison under Ordered-LT (Backward) ID and OOD label shift. Settings in which our model outperforms baselines are colored in gray.} Our model outperforms baselines under most label shift settings. Each metric is averaged among the corresponding OOD test set (Tab.~\ref{tab:dataset-setup}) and over three independent ID classifiers.}
    \label{Atab:wmse-cifar100-LTr}
\vspace{-1em} \end{table*}
\setlength\tabcolsep{6pt}

\setlength\tabcolsep{0.8pt} 
\begin{table*}[ht]
    \centering
    \scriptsize
    \renewcommand{\arraystretch}{1.1}
    \begin{tabular}{c | c | c||c c c | c c c } \hline\hline 
    \multicolumn{3}{c||}{Dataset} & \multicolumn{6}{c}{CIFAR100} \\ \hline
       \multicolumn{3}{c||}{ID label Shift param} & \multicolumn{3}{c|}{Dir $\alpha=1.0$} & \multicolumn{3}{c}{Dir $\alpha=10.0$} \\ \hline
      \multicolumn{3}{c||}{OOD label shift param $r$} & $1.0$ & $0.1$ & $0.01$  & $1.0$ & $0.1$ & $0.01$  \\ \hline
\multicolumn{9}{c}{Closed Set Label Shift estimation models} \\ \hline\hline 
\multicolumn{2}{c|}{\multirow{2}{*}{BBSE}}  & Near & $0.659_{\scaleto{\pm 0.011}{3pt}}$ & $0.138_{\scaleto{\pm 0.018}{3pt}}$ & $0.207_{\scaleto{\pm 0.017}{3pt}}$ & $0.466_{\scaleto{\pm 0.051}{3pt}}$ & $0.108_{\scaleto{\pm 0.009}{3pt}}$ & $0.106_{\scaleto{\pm 0.013}{3pt}}$\\ 
\multicolumn{2}{c|}{} & Far &  $4.243_{\scaleto{\pm 0.232}{3pt}}$ & $0.263_{\scaleto{\pm 0.011}{3pt}}$ & $0.210_{\scaleto{\pm 0.019}{3pt}}$ & $3.874_{\scaleto{\pm 0.157}{3pt}}$ & $0.240_{\scaleto{\pm 0.015}{3pt}}$ & $0.110_{\scaleto{\pm 0.013}{3pt}}$\\ \hline 
\multicolumn{2}{c|}{\multirow{2}{*}{MLLS}}  & Near & $1.006_{\scaleto{\pm 0.072}{3pt}}$ & $0.139_{\scaleto{\pm 0.002}{3pt}}$ & $0.162_{\scaleto{\pm 0.047}{3pt}}$ & $0.845_{\scaleto{\pm 0.108}{3pt}}$ & $0.112_{\scaleto{\pm 0.012}{3pt}}$ & $0.095_{\scaleto{\pm 0.022}{3pt}}$\\ 
\multicolumn{2}{c|}{} & Far &  $9.871_{\scaleto{\pm 1.504}{3pt}}$ & $0.369_{\scaleto{\pm 0.030}{3pt}}$ & $0.162_{\scaleto{\pm 0.046}{3pt}}$ & $9.281_{\scaleto{\pm 1.478}{3pt}}$ & $0.337_{\scaleto{\pm 0.047}{3pt}}$ & $0.097_{\scaleto{\pm 0.021}{3pt}}$\\
\hline\hline
    \end{tabular} \vspace{-1em}
    \caption{\textbf{Estimation Error $(w-\hat{w})^2/K(\downarrow)$ of our OSLS estimation model on the CIFAR100 dataset with Near OOD datasets and Far OOD datasets comparison under Dirichlet ID and OOD label shift. Settings in which our model outperforms baselines are colored in gray.} Our model outperforms baselines under most label shift settings. Each metric is averaged among the corresponding OOD test set (Tab.~\ref{tab:dataset-setup}) and over three independent ID classifiers.}
    \label{Atab:acc-cifar100-dir}
\vspace{-1em} \end{table*}
\setlength\tabcolsep{6pt}

\clearpage
\subsection{ImageNet-200}

\setlength\tabcolsep{0.8pt} 
\begin{table*}[ht]
    \centering
    \scriptsize
    \renewcommand{\arraystretch}{1.1}
    \begin{tabular}{c | c | c||c c c | c c c | c c c } \hline\hline
    \multicolumn{3}{c||}{Dataset} & \multicolumn{9}{c}{ImageNet-200} \\ \hline
       \multicolumn{3}{c||}{ID label Shift param}  & \multicolumn{3}{c|}{LT10 Forward} & \multicolumn{3}{c|}{LT50 Forward} & \multicolumn{3}{c}{LT100 Forward} \\ \hline
      \multicolumn{3}{c||}{OOD label shift param $r$} & $1.0$ & $0.1$ & $0.01$  & $1.0$ & $0.1$ & $0.01$  & $1.0$ & $0.1$ & $0.01$ \\ \hline \hline
\multicolumn{12}{c}{Closed Set Label Shift estimation models} \\ \hline\hline 
\multicolumn{2}{c|}{\multirow{2}{*}{BBSE}}  & Near & $0.564_{\scaleto{\pm 0.014}{3pt}}$ & $0.119_{\scaleto{\pm 0.015}{3pt}}$ & $0.107_{\scaleto{\pm 0.012}{3pt}}$ & $0.664_{\scaleto{\pm 0.030}{3pt}}$ & $0.128_{\scaleto{\pm 0.016}{3pt}}$ & $0.115_{\scaleto{\pm 0.012}{3pt}}$ & $0.735_{\scaleto{\pm 0.040}{3pt}}$ & $0.132_{\scaleto{\pm 0.009}{3pt}}$ & $0.112_{\scaleto{\pm 0.017}{3pt}}$\\ 
\multicolumn{2}{c|}{} & Far &  $1.148_{\scaleto{\pm 0.042}{3pt}}$ & $0.134_{\scaleto{\pm 0.018}{3pt}}$ & $0.108_{\scaleto{\pm 0.012}{3pt}}$ & $1.301_{\scaleto{\pm 0.031}{3pt}}$ & $0.141_{\scaleto{\pm 0.016}{3pt}}$ & $0.115_{\scaleto{\pm 0.013}{3pt}}$ & $1.389_{\scaleto{\pm 0.039}{3pt}}$ & $0.146_{\scaleto{\pm 0.012}{3pt}}$ & $0.112_{\scaleto{\pm 0.016}{3pt}}$\\ \hline 
\multicolumn{2}{c|}{\multirow{2}{*}{MLLS}}  & Near & $1.152_{\scaleto{\pm 0.101}{3pt}}$ & $0.131_{\scaleto{\pm 0.019}{3pt}}$ & $0.099_{\scaleto{\pm 0.017}{3pt}}$ & $1.233_{\scaleto{\pm 0.128}{3pt}}$ & $0.143_{\scaleto{\pm 0.029}{3pt}}$ & $0.111_{\scaleto{\pm 0.023}{3pt}}$ & $1.272_{\scaleto{\pm 0.128}{3pt}}$ & $0.146_{\scaleto{\pm 0.025}{3pt}}$ & $0.116_{\scaleto{\pm 0.027}{3pt}}$\\
\hline

         \hline
    \end{tabular} \vspace{-1em}
    \caption{\textbf{Estimation Error $(w-\hat{w})^2/K(\downarrow)$ of our OSLS estimation model on the ImageNet-200 dataset with Near OOD datasets and Far OOD datasets comparison under Ordered-LT (Forward) ID and OOD label shift. Settings in which our model outperforms baselines are colored in gray.} Our model outperforms baselines under most label shift settings. Each metric is averaged among the corresponding OOD test set (Tab.~\ref{tab:dataset-setup}) and over three independent ID classifiers.}
    \label{Atab:wmse-imagenet200-LT}
\vspace{-1em} \end{table*}
\setlength\tabcolsep{6pt}

\setlength\tabcolsep{0.8pt} 
\begin{table*}[ht]
    \centering
    \scriptsize
    \renewcommand{\arraystretch}{1.1}
    \begin{tabular}{c | c | c||c c c | c c c | c c c } \hline\hline
    \multicolumn{3}{c||}{Dataset} & \multicolumn{9}{c}{ImageNet-200} \\ \hline
       \multicolumn{3}{c||}{ID label Shift param}  & \multicolumn{3}{c|}{LT10 Backward} & \multicolumn{3}{c|}{LT50 Backward} & \multicolumn{3}{c}{LT100 Backward} \\ \hline
      \multicolumn{3}{c||}{OOD label shift param $r$} & $1.0$ & $0.1$ & $0.01$  & $1.0$ & $0.1$ & $0.01$  & $1.0$ & $0.1$ & $0.01$ \\ \hline \hline
      
\multicolumn{12}{c}{Closed Set Label Shift estimation models} \\ \hline\hline 
\multicolumn{2}{c|}{\multirow{2}{*}{BBSE}}  & Near & $0.651_{\scaleto{\pm 0.079}{3pt}}$ & $0.240_{\scaleto{\pm 0.077}{3pt}}$ & $0.215_{\scaleto{\pm 0.049}{3pt}}$ & $0.848_{\scaleto{\pm 0.074}{3pt}}$ & $0.332_{\scaleto{\pm 0.086}{3pt}}$ & $0.323_{\scaleto{\pm 0.085}{3pt}}$ & $0.930_{\scaleto{\pm 0.063}{3pt}}$ & $0.364_{\scaleto{\pm 0.112}{3pt}}$ & $0.336_{\scaleto{\pm 0.090}{3pt}}$\\ 
\multicolumn{2}{c|}{} & Far &  $1.088_{\scaleto{\pm 0.049}{3pt}}$ & $0.253_{\scaleto{\pm 0.055}{3pt}}$ & $0.214_{\scaleto{\pm 0.046}{3pt}}$ & $1.237_{\scaleto{\pm 0.054}{3pt}}$ & $0.342_{\scaleto{\pm 0.075}{3pt}}$ & $0.322_{\scaleto{\pm 0.085}{3pt}}$ & $1.329_{\scaleto{\pm 0.045}{3pt}}$ & $0.374_{\scaleto{\pm 0.096}{3pt}}$ & $0.338_{\scaleto{\pm 0.090}{3pt}}$\\ \hline 
\multicolumn{2}{c|}{\multirow{2}{*}{MLLS}}  & Near & $1.297_{\scaleto{\pm 0.163}{3pt}}$ & $0.272_{\scaleto{\pm 0.086}{3pt}}$ & $0.220_{\scaleto{\pm 0.043}{3pt}}$ & $1.448_{\scaleto{\pm 0.143}{3pt}}$ & $0.381_{\scaleto{\pm 0.099}{3pt}}$ & $0.346_{\scaleto{\pm 0.092}{3pt}}$ & $1.520_{\scaleto{\pm 0.135}{3pt}}$ & $0.406_{\scaleto{\pm 0.111}{3pt}}$ & $0.346_{\scaleto{\pm 0.081}{3pt}}$\\
\hline

         \hline
    \end{tabular} \vspace{-1em}
    \caption{\textbf{Estimation Error $(w-\hat{w})^2/K(\downarrow)$ of our OSLS estimation model on the ImageNet-200 dataset with Near OOD datasets and Far OOD datasets comparison under Ordered-LT (Backward) ID and OOD label shift. Settings in which our model outperforms baselines are colored in gray.} Our model outperforms baselines under most label shift settings. Each metric is averaged among the corresponding OOD test set (Tab.~\ref{tab:dataset-setup}) and over three independent ID classifiers.}
    \label{Atab:wmse-imagenet200-LTr}
\vspace{-1em} \end{table*}
\setlength\tabcolsep{6pt}

\setlength\tabcolsep{0.8pt} 
\begin{table*}[ht]
    \centering
    \scriptsize
    \renewcommand{\arraystretch}{1.1}
    \begin{tabular}{c | c | c||c c c | c c c } \hline\hline 
    \multicolumn{3}{c||}{Dataset} & \multicolumn{6}{c}{ImageNet-200} \\ \hline
       \multicolumn{3}{c||}{ID label Shift param} & \multicolumn{3}{c|}{Dir $\alpha=1.0$} & \multicolumn{3}{c}{Dir $\alpha=10.0$} \\ \hline
      \multicolumn{3}{c||}{OOD label shift param $r$} & $1.0$ & $0.1$ & $0.01$  & $1.0$ & $0.1$ & $0.01$  \\ \hline
      
\multicolumn{9}{c}{Closed Set Label Shift estimation models} \\ \hline\hline 
\multicolumn{2}{c|}{\multirow{2}{*}{BBSE}}  & Near & $0.808_{\scaleto{\pm 0.125}{3pt}}$ & $0.211_{\scaleto{\pm 0.044}{3pt}}$ & $0.190_{\scaleto{\pm 0.045}{3pt}}$ & $0.556_{\scaleto{\pm 0.048}{3pt}}$ & $0.152_{\scaleto{\pm 0.038}{3pt}}$ & $0.123_{\scaleto{\pm 0.015}{3pt}}$\\ 
\multicolumn{2}{c|}{} & Far &  $1.330_{\scaleto{\pm 0.085}{3pt}}$ & $0.230_{\scaleto{\pm 0.042}{3pt}}$ & $0.192_{\scaleto{\pm 0.045}{3pt}}$ & $1.051_{\scaleto{\pm 0.065}{3pt}}$ & $0.180_{\scaleto{\pm 0.039}{3pt}}$ & $0.122_{\scaleto{\pm 0.015}{3pt}}$\\ \hline 
\multicolumn{2}{c|}{\multirow{2}{*}{MLLS}}  & Near & $1.372_{\scaleto{\pm 0.224}{3pt}}$ & $0.140_{\scaleto{\pm 0.026}{3pt}}$ & $0.152_{\scaleto{\pm 0.061}{3pt}}$ & $1.213_{\scaleto{\pm 0.168}{3pt}}$ & $0.161_{\scaleto{\pm 0.047}{3pt}}$ & $0.122_{\scaleto{\pm 0.028}{3pt}}$\\ 
\multicolumn{2}{c|}{} & Far &  $4.343_{\scaleto{\pm 0.354}{3pt}}$ & $0.180_{\scaleto{\pm 0.041}{3pt}}$ & $0.155_{\scaleto{\pm 0.062}{3pt}}$ & $4.027_{\scaleto{\pm 0.316}{3pt}}$ & $0.205_{\scaleto{\pm 0.049}{3pt}}$ & $0.121_{\scaleto{\pm 0.028}{3pt}}$\\
\hline
    \end{tabular} \vspace{-1em}
    \caption{\textbf{Estimation Error $(w-\hat{w})^2/K(\downarrow)$ of our OSLS estimation model on the ImageNet200 dataset with Near OOD datasets and Far OOD datasets comparison under Dirichlet ID and OOD label shift. Settings in which our model outperforms baselines are colored in gray.} Our model outperforms baselines under most label shift settings. Each metric is averaged among the corresponding OOD test set (Tab.~\ref{tab:dataset-setup}) and over three independent ID classifiers.}
    \label{Atab:wmse-imagenet200-dir}
\vspace{-1em} \end{table*}
\setlength\tabcolsep{6pt}

\clearpage
\section{More Accuracy Results}\label{Asec:detailed-acc}
We provide the full open set label shift Top1 Accuracy on the CIFAR10 and CIFAR100 datasets. Similar to the estimation error metric, our model outperforms baseline in most of the experimental setup.

\subsection{CIFAR10}

\setlength\tabcolsep{2pt} 
\begin{table*}[ht]
    \centering
    \scriptsize
    \renewcommand{\arraystretch}{1.2}
    \begin{tabular}{c | c | c||c c c | c c c | c c c } \hline\hline 
       \multicolumn{3}{c||}{ID label Shift param} & \multicolumn{3}{|c}{LT-10} & \multicolumn{3}{|c}{LT-50} & \multicolumn{3}{|c}{LT 100} \\ \hline
      \multicolumn{3}{c||}{OOD label shift param $r$} & $1.0$ & $0.1$ & $0.01$  & $1.0$ & $0.1$ & $0.01$  & $1.0$ & $0.1$ & $0.01$ \\ \hline
\multirow{6}{*}{OpenMax}  & \multirow{3}{*}{Near} &  Original & $74.650_{\scaleto{\pm 1.220}{3pt}}$ & $86.800_{\scaleto{\pm 0.020}{3pt}}$ & $89.690_{\scaleto{\pm 0.340}{3pt}}$ & $74.340_{\scaleto{\pm 0.530}{3pt}}$ & $87.650_{\scaleto{\pm 0.080}{3pt}}$ & $89.940_{\scaleto{\pm 0.410}{3pt}}$ & $74.750_{\scaleto{\pm 0.910}{3pt}}$ & $87.160_{\scaleto{\pm 0.220}{3pt}}$ & $90.190_{\scaleto{\pm 0.440}{3pt}}$\\ 
 & & Baseline & $75.330_{\scaleto{\pm 0.830}{3pt}}$ & $86.580_{\scaleto{\pm 0.140}{3pt}}$ & $89.340_{\scaleto{\pm 0.140}{3pt}}$ & $75.060_{\scaleto{\pm 0.310}{3pt}}$ & $87.340_{\scaleto{\pm 0.070}{3pt}}$ & $89.540_{\scaleto{\pm 0.270}{3pt}}$ & $75.300_{\scaleto{\pm 0.550}{3pt}}$ & $86.930_{\scaleto{\pm 0.360}{3pt}}$ & $89.800_{\scaleto{\pm 0.340}{3pt}}$\\ 
&  &  \textbf{ours} & \cc $75.330_{\scaleto{\pm 0.830}{3pt}}$ & $86.580_{\scaleto{\pm 0.140}{3pt}}$ & $89.340_{\scaleto{\pm 0.140}{3pt}}$ & \cc $75.060_{\scaleto{\pm 0.310}{3pt}}$ & $87.390_{\scaleto{\pm 0.090}{3pt}}$ & $89.600_{\scaleto{\pm 0.310}{3pt}}$ & \cc $75.300_{\scaleto{\pm 0.550}{3pt}}$ & $86.980_{\scaleto{\pm 0.410}{3pt}}$ & $89.930_{\scaleto{\pm 0.400}{3pt}}$\\ \cline{2-12} 
& \multirow{3}{*}{Far} & Original & $76.560_{\scaleto{\pm 0.560}{3pt}}$ & $87.110_{\scaleto{\pm 0.080}{3pt}}$ & $89.720_{\scaleto{\pm 0.390}{3pt}}$ & $76.550_{\scaleto{\pm 0.120}{3pt}}$ & $87.940_{\scaleto{\pm 0.170}{3pt}}$ & $89.910_{\scaleto{\pm 0.460}{3pt}}$ & $76.730_{\scaleto{\pm 0.240}{3pt}}$ & $87.530_{\scaleto{\pm 0.340}{3pt}}$ & $90.240_{\scaleto{\pm 0.460}{3pt}}$\\
\hline

         \hline
    \end{tabular} \vspace{-1em} 
    \label{Atab:acc-cifar10-LT}
    \caption{\textbf{Top1 Accuracy $(\uparrow)$ of our OSLS correction model on the CIFAR10 dataset with Near OOD datasets and Far OOD datasets comparison under under Ordered-LT (Backward) ID and OOD label shift. Settings in which our model outperforms baselines are colored in gray.} Our model outperforms baselines under most label shift settings. Each metric is averaged among the corresponding OOD test set (Tab.~\ref{tab:dataset-setup}) and over three independent ID classifiers.}

    \begin{tabular}{c | c | c||c c c | c c c | c c c } \hline\hline
       \multicolumn{3}{c||}{ID label Shift param} & \multicolumn{3}{|c}{LT-10} & \multicolumn{3}{|c}{LT-50} & \multicolumn{3}{|c}{LT 100} \\ \hline
      \multicolumn{3}{c||}{OOD label shift param $r$} & $1.0$ & $0.1$ & $0.01$  & $1.0$ & $0.1$ & $0.01$  & $1.0$ & $0.1$ & $0.01$ \\ \hline
\multirow{6}{*}{OpenMax}  & \multirow{3}{*}{Near} &  Original & $75.010_{\scaleto{\pm 0.970}{3pt}}$ & $88.340_{\scaleto{\pm 0.440}{3pt}}$ & $90.400_{\scaleto{\pm 0.410}{3pt}}$ & $75.610_{\scaleto{\pm 1.210}{3pt}}$ & $88.670_{\scaleto{\pm 0.320}{3pt}}$ & $91.100_{\scaleto{\pm 0.330}{3pt}}$ & $75.430_{\scaleto{\pm 1.280}{3pt}}$ & $88.860_{\scaleto{\pm 0.160}{3pt}}$ & $91.260_{\scaleto{\pm 0.060}{3pt}}$\\ 
 & & Baseline & $75.720_{\scaleto{\pm 0.660}{3pt}}$ & $88.210_{\scaleto{\pm 0.480}{3pt}}$ & $90.100_{\scaleto{\pm 0.310}{3pt}}$ & $76.300_{\scaleto{\pm 0.870}{3pt}}$ & $88.610_{\scaleto{\pm 0.380}{3pt}}$ & $90.860_{\scaleto{\pm 0.370}{3pt}}$ & $76.170_{\scaleto{\pm 0.940}{3pt}}$ & $88.830_{\scaleto{\pm 0.230}{3pt}}$ & $91.070_{\scaleto{\pm 0.030}{3pt}}$\\ 
&  &  \textbf{ours} & \cc $75.720_{\scaleto{\pm 0.660}{3pt}}$ & $88.200_{\scaleto{\pm 0.480}{3pt}}$ & $90.110_{\scaleto{\pm 0.300}{3pt}}$ & \cc $76.300_{\scaleto{\pm 0.880}{3pt}}$ & $88.650_{\scaleto{\pm 0.370}{3pt}}$ & $90.950_{\scaleto{\pm 0.350}{3pt}}$ & \cc $76.180_{\scaleto{\pm 0.950}{3pt}}$ & \cc $88.950_{\scaleto{\pm 0.230}{3pt}}$ & $91.200_{\scaleto{\pm 0.040}{3pt}}$\\ \cline{2-12} 
& \multirow{3}{*}{Far} & Original & $76.930_{\scaleto{\pm 0.520}{3pt}}$ & $88.770_{\scaleto{\pm 0.350}{3pt}}$ & $90.450_{\scaleto{\pm 0.340}{3pt}}$ & $77.750_{\scaleto{\pm 0.650}{3pt}}$ & $88.870_{\scaleto{\pm 0.370}{3pt}}$ & $91.070_{\scaleto{\pm 0.240}{3pt}}$ & $77.550_{\scaleto{\pm 0.610}{3pt}}$ & $89.260_{\scaleto{\pm 0.160}{3pt}}$ & $91.360_{\scaleto{\pm 0.060}{3pt}}$\\
\hline

         \hline
    \end{tabular} \vspace{-1em}
    \label{Atab:acc-cifar10-LTr}
    \caption{\textbf{Top1 Accuracy $(\uparrow)$ of our OSLS correction model on the CIFAR10 dataset with Near OOD datasets and Far OOD datasets comparison under Ordered-LT (Backward) ID and OOD label shift. Settings in which our model outperforms baselines are colored in gray.} Our model outperforms baselines under most label shift settings. Each metric is averaged among the corresponding OOD test set (Tab.~\ref{tab:dataset-setup}) and over three independent ID classifiers.}

\vspace{-1em} \end{table*}
\setlength\tabcolsep{6pt} 

\clearpage
\setlength\tabcolsep{2pt} 
\begin{table*}[ht]
    \centering
    \scriptsize
    \renewcommand{\arraystretch}{1.2}
    \begin{tabular}{c | c | c||c c c | c c c } \hline\hline 
       \multicolumn{3}{c||}{ID label Shift param} & \multicolumn{3}{c|}{Dir $\alpha=1.0$} & \multicolumn{3}{c}{Dir $\alpha=10.0$} \\ \hline
      \multicolumn{3}{c||}{OOD label shift param $r$} & $1.0$ & $0.1$ & $0.01$  & $1.0$ & $0.1$ & $0.01$  \\ \hline
\multirow{6}{*}{OpenMax}  & \multirow{3}{*}{Near} &  Original & $74.54_{\scaleto{\pm 0.65}{3pt}}$ & $87.11_{\scaleto{\pm 0.57}{3pt}}$ & $89.33_{\scaleto{\pm 1.08}{3pt}}$ & $74.80_{\scaleto{\pm 1.17}{3pt}}$ & $86.84_{\scaleto{\pm 0.30}{3pt}}$ & $89.85_{\scaleto{\pm 0.61}{3pt}}$\\ 
 & & Baseline & $76.73_{\scaleto{\pm 1.53}{3pt}}$ & $86.45_{\scaleto{\pm 0.54}{3pt}}$ & $88.00_{\scaleto{\pm 1.22}{3pt}}$ & $76.37_{\scaleto{\pm 1.00}{3pt}}$ & $86.29_{\scaleto{\pm 0.31}{3pt}}$ & $88.65_{\scaleto{\pm 0.99}{3pt}}$\\ 
&  &  \textbf{ours} & \cc $76.73_{\scaleto{\pm 1.53}{3pt}}$ & $86.46_{\scaleto{\pm 0.55}{3pt}}$ & $88.00_{\scaleto{\pm 1.22}{3pt}}$ & \cc $76.37_{\scaleto{\pm 1.00}{3pt}}$ & $86.29_{\scaleto{\pm 0.31}{3pt}}$ & $88.65_{\scaleto{\pm 0.99}{3pt}}$\\ \cline{2-9} 
& \multirow{3}{*}{Far} & Original & $76.35_{\scaleto{\pm 0.39}{3pt}}$ & $87.63_{\scaleto{\pm 0.58}{3pt}}$ & $89.40_{\scaleto{\pm 1.02}{3pt}}$ & $76.79_{\scaleto{\pm 0.85}{3pt}}$ & $87.11_{\scaleto{\pm 0.20}{3pt}}$ & $89.88_{\scaleto{\pm 0.61}{3pt}}$\\ 
 & & Baseline & $78.91_{\scaleto{\pm 2.30}{3pt}}$ & $86.95_{\scaleto{\pm 0.39}{3pt}}$ & $88.07_{\scaleto{\pm 1.23}{3pt}}$ & $78.86_{\scaleto{\pm 1.75}{3pt}}$ & $86.63_{\scaleto{\pm 0.35}{3pt}}$ & $88.71_{\scaleto{\pm 1.06}{3pt}}$\\ 
&  &  \textbf{ours} & \cc $78.91_{\scaleto{\pm 2.30}{3pt}}$ & $86.96_{\scaleto{\pm 0.41}{3pt}}$ & $88.07_{\scaleto{\pm 1.22}{3pt}}$ & \cc $78.86_{\scaleto{\pm 1.75}{3pt}}$ & $86.63_{\scaleto{\pm 0.35}{3pt}}$ & $88.71_{\scaleto{\pm 1.06}{3pt}}$\\
\hline 

         \hline
    \end{tabular} \vspace{-1em}
    \caption{\textbf{Estimation Error $(w-\hat{w})^2/K(\downarrow)$ of our OSLS estimation model on the CIFAR10 dataset with Near OOD datasets and Far OOD datasets comparison under Dirichlet ID and OOD label shift. Settings in which our model outperforms baselines are colored in gray.} Our model outperforms baselines under most label shift settings. Each metric is averaged among the corresponding OOD test set (Tab.~\ref{tab:dataset-setup}) and over three independent ID classifiers.}
    \label{Atab:wmse-cifar10-dir}
\vspace{-1em} \end{table*}
\setlength\tabcolsep{6pt}

\clearpage
\subsection{CIFAR100}

\setlength\tabcolsep{2pt} 
\begin{table*}[ht]
    \centering
    \scriptsize
    \renewcommand{\arraystretch}{1.2}
    \begin{tabular}{c | c | c||c c c | c c c | c c c } \hline\hline
       \multicolumn{3}{c||}{ID label Shift param} & \multicolumn{3}{|c}{LT-10} & \multicolumn{3}{|c}{LT-50} & \multicolumn{3}{|c}{LT 100} \\ \hline
      \multicolumn{3}{c||}{OOD label shift param $r$} & $1.0$ & $0.1$ & $0.01$  & $1.0$ & $0.1$ & $0.01$  & $1.0$ & $0.1$ & $0.01$ \\ \hline
\multirow{6}{*}{OpenMax}  & \multirow{3}{*}{Near} &  Original & $63.540_{\scaleto{\pm 0.260}{3pt}}$ & $65.200_{\scaleto{\pm 0.350}{3pt}}$ & $65.350_{\scaleto{\pm 0.370}{3pt}}$ & $63.700_{\scaleto{\pm 0.360}{3pt}}$ & $65.120_{\scaleto{\pm 0.480}{3pt}}$ & $64.960_{\scaleto{\pm 0.630}{3pt}}$ & $63.420_{\scaleto{\pm 0.330}{3pt}}$ & $64.860_{\scaleto{\pm 0.360}{3pt}}$ & $65.200_{\scaleto{\pm 0.700}{3pt}}$\\ 
 & & Baseline & $62.710_{\scaleto{\pm 0.170}{3pt}}$ & $64.920_{\scaleto{\pm 0.320}{3pt}}$ & $65.130_{\scaleto{\pm 0.300}{3pt}}$ & $62.950_{\scaleto{\pm 0.350}{3pt}}$ & $64.720_{\scaleto{\pm 0.550}{3pt}}$ & $64.610_{\scaleto{\pm 0.640}{3pt}}$ & $62.670_{\scaleto{\pm 0.270}{3pt}}$ & $64.610_{\scaleto{\pm 0.280}{3pt}}$ & $65.100_{\scaleto{\pm 0.520}{3pt}}$\\ 
&  &  \textbf{ours} & $62.950_{\scaleto{\pm 0.200}{3pt}}$ & \cc $65.250_{\scaleto{\pm 0.340}{3pt}}$ & \cc $65.700_{\scaleto{\pm 0.400}{3pt}}$ & $63.430_{\scaleto{\pm 0.410}{3pt}}$ & \cc $65.780_{\scaleto{\pm 0.510}{3pt}}$ & \cc $65.900_{\scaleto{\pm 0.400}{3pt}}$ & $63.230_{\scaleto{\pm 0.370}{3pt}}$ & \cc $66.030_{\scaleto{\pm 0.680}{3pt}}$ & \cc $66.760_{\scaleto{\pm 0.880}{3pt}}$\\ \cline{2-12} 
& \multirow{3}{*}{Far} & Original & $66.770_{\scaleto{\pm 0.340}{3pt}}$ & $65.930_{\scaleto{\pm 0.360}{3pt}}$ & $65.430_{\scaleto{\pm 0.400}{3pt}}$ & $67.110_{\scaleto{\pm 0.380}{3pt}}$ & $65.950_{\scaleto{\pm 0.530}{3pt}}$ & $65.010_{\scaleto{\pm 0.690}{3pt}}$ & $66.760_{\scaleto{\pm 0.060}{3pt}}$ & $65.320_{\scaleto{\pm 0.380}{3pt}}$ & $65.300_{\scaleto{\pm 0.720}{3pt}}$\\
\hline 

\hline
    \end{tabular} \vspace{-1em}
    \label{Atab:acc-cifar100-LT}
    \caption{\textbf{Top1 Accuracy $(\uparrow)$  of our OSLS estimation and correction model on the CIFAR100 dataset with Near OOD datasets and Far OOD datasets comparison under Ordered-LT (Forward) ID and OOD label shift. Settings in which our model outperforms baselines are colored in gray.} Our model outperforms baselines under most label shift settings. Each metric is averaged among the corresponding OOD test set (Tab.~\ref{tab:dataset-setup}) and over three independent ID classifiers.}


    \begin{tabular}{c | c | c||c c c | c c c | c c c } \hline\hline
       \multicolumn{3}{c||}{ID label Shift param} & \multicolumn{3}{|c}{LT-10} & \multicolumn{3}{|c}{LT-50} & \multicolumn{3}{|c}{LT 100} \\ \hline
      \multicolumn{3}{c||}{OOD label shift param $r$} & $1.0$ & $0.1$ & $0.01$  & $1.0$ & $0.1$ & $0.01$  & $1.0$ & $0.1$ & $0.01$ \\ \hline
\multirow{6}{*}{OpenMax}  & \multirow{3}{*}{Near} &  Original & $63.770_{\scaleto{\pm 0.360}{3pt}}$ & $65.680_{\scaleto{\pm 0.340}{3pt}}$ & $65.990_{\scaleto{\pm 0.380}{3pt}}$ & $63.760_{\scaleto{\pm 0.440}{3pt}}$ & $66.280_{\scaleto{\pm 0.990}{3pt}}$ & $66.390_{\scaleto{\pm 0.320}{3pt}}$ & $63.960_{\scaleto{\pm 0.580}{3pt}}$ & $65.920_{\scaleto{\pm 0.590}{3pt}}$ & $66.550_{\scaleto{\pm 1.100}{3pt}}$\\ 
 & & Baseline & $63.000_{\scaleto{\pm 0.300}{3pt}}$ & $65.540_{\scaleto{\pm 0.280}{3pt}}$ & $65.480_{\scaleto{\pm 0.270}{3pt}}$ & $63.030_{\scaleto{\pm 0.190}{3pt}}$ & $65.920_{\scaleto{\pm 0.830}{3pt}}$ & $65.890_{\scaleto{\pm 0.220}{3pt}}$ & $63.290_{\scaleto{\pm 0.550}{3pt}}$ & $65.490_{\scaleto{\pm 0.350}{3pt}}$ & $66.240_{\scaleto{\pm 0.740}{3pt}}$\\ 
&  &  \textbf{ours} & $63.210_{\scaleto{\pm 0.280}{3pt}}$ & \cc $65.960_{\scaleto{\pm 0.390}{3pt}}$ & \cc $65.990_{\scaleto{\pm 0.220}{3pt}}$ & $63.680_{\scaleto{\pm 0.130}{3pt}}$ & \cc $67.050_{\scaleto{\pm 0.930}{3pt}}$ & \cc $67.200_{\scaleto{\pm 0.110}{3pt}}$ & $63.930_{\scaleto{\pm 0.460}{3pt}}$ & \cc $67.080_{\scaleto{\pm 0.420}{3pt}}$ & \cc $67.910_{\scaleto{\pm 0.990}{3pt}}$\\ \cline{2-12} 
& \multirow{3}{*}{Far} & Original & $67.160_{\scaleto{\pm 0.970}{3pt}}$ & $66.180_{\scaleto{\pm 0.350}{3pt}}$ & $66.020_{\scaleto{\pm 0.330}{3pt}}$ & $66.800_{\scaleto{\pm 0.780}{3pt}}$ & $66.960_{\scaleto{\pm 0.700}{3pt}}$ & $66.510_{\scaleto{\pm 0.340}{3pt}}$ & $67.390_{\scaleto{\pm 0.950}{3pt}}$ & $66.490_{\scaleto{\pm 0.720}{3pt}}$ & $66.630_{\scaleto{\pm 1.070}{3pt}}$\\
\hline 

\hline
    \end{tabular} \vspace{-1em}
    \label{Atab:acc-cifar100-LTr}
    \caption{\textbf{Top1 Accuracy $(\uparrow)$ of our OSLS estimation and correction model on the CIFAR100 dataset with Near OOD datasets and Far OOD datasets comparison under Ordered-LT (Backward) ID and OOD label shift. Settings in which our model outperforms baselines are colored in gray.} Our model outperforms baselines under most label shift settings. Each metric is averaged among the corresponding OOD test set (Tab.~\ref{tab:dataset-setup}) and over three independent ID classifiers.}

\vspace{-1em} 
\end{table*}
\setlength\tabcolsep{6pt} 

\clearpage
\setlength\tabcolsep{2pt} 
\begin{table*}[ht]
    \centering
    \scriptsize
    \renewcommand{\arraystretch}{1.2}
    \begin{tabular}{c | c | c||c c c | c c c } \hline\hline 
       \multicolumn{3}{c||}{ID label Shift param} & \multicolumn{3}{c|}{Dir $\alpha=1.0$} & \multicolumn{3}{c}{Dir $\alpha=10.0$} \\ \hline
      \multicolumn{3}{c||}{OOD label shift param $r$} & $1.0$ & $0.1$ & $0.01$  & $1.0$ & $0.1$ & $0.01$  \\ \hline
\multirow{6}{*}{OpenMax}  & \multirow{3}{*}{Near} &  Original & $63.77_{\scaleto{\pm 0.34}{3pt}}$ & $66.25_{\scaleto{\pm 0.75}{3pt}}$ & $65.58_{\scaleto{\pm 0.70}{3pt}}$ & $64.18_{\scaleto{\pm 0.14}{3pt}}$ & $65.16_{\scaleto{\pm 0.32}{3pt}}$ & $66.96_{\scaleto{\pm 0.22}{3pt}}$\\ 
 & & Baseline & $66.59_{\scaleto{\pm 0.46}{3pt}}$ & $64.31_{\scaleto{\pm 0.19}{3pt}}$ & $62.61_{\scaleto{\pm 1.53}{3pt}}$ & $66.69_{\scaleto{\pm 0.95}{3pt}}$ & $63.13_{\scaleto{\pm 0.88}{3pt}}$ & $64.22_{\scaleto{\pm 0.73}{3pt}}$\\ 
&  &  \textbf{ours} & \cc $66.96_{\scaleto{\pm 0.36}{3pt}}$ & $65.19_{\scaleto{\pm 0.13}{3pt}}$ & $63.30_{\scaleto{\pm 1.54}{3pt}}$ & \cc $66.72_{\scaleto{\pm 0.85}{3pt}}$ & $63.22_{\scaleto{\pm 0.97}{3pt}}$ & $64.19_{\scaleto{\pm 0.82}{3pt}}$\\ \cline{2-9} 
& \multirow{3}{*}{Far} & Original & $67.16_{\scaleto{\pm 0.69}{3pt}}$ & $66.89_{\scaleto{\pm 0.72}{3pt}}$ & $65.68_{\scaleto{\pm 0.66}{3pt}}$ & $67.29_{\scaleto{\pm 0.44}{3pt}}$ & $65.73_{\scaleto{\pm 0.71}{3pt}}$ & $67.06_{\scaleto{\pm 0.22}{3pt}}$\\ 
 & & Baseline & $69.64_{\scaleto{\pm 0.94}{3pt}}$ & $64.90_{\scaleto{\pm 0.16}{3pt}}$ & $62.70_{\scaleto{\pm 1.51}{3pt}}$ & $69.71_{\scaleto{\pm 1.06}{3pt}}$ & $63.72_{\scaleto{\pm 1.17}{3pt}}$ & $64.27_{\scaleto{\pm 0.72}{3pt}}$\\ 
&  &  \textbf{ours} & \cc $70.02_{\scaleto{\pm 0.77}{3pt}}$ & $65.77_{\scaleto{\pm 0.16}{3pt}}$ & $63.37_{\scaleto{\pm 1.54}{3pt}}$ & $69.70_{\scaleto{\pm 0.96}{3pt}}$ & $63.74_{\scaleto{\pm 1.22}{3pt}}$ & $64.23_{\scaleto{\pm 0.80}{3pt}}$\\
\hline

         \hline
    \end{tabular} \vspace{-1em}
    \caption{\textbf{Top1 Accuracy $(\uparrow)$ of our OSLS correction model on the CIFAR100 dataset with Near OOD datasets and Far OOD datasets comparison under Dirichlet ID and OOD label shift. Settings in which our model outperforms baselines are colored in gray.} Our model outperforms baselines under most label shift settings. Each metric is averaged among the corresponding OOD test set (Tab.~\ref{tab:dataset-setup}) and over three independent ID classifiers.}
    \label{Atab:wmse-cifar100-dir}
\vspace{-1em} \end{table*}
\setlength\tabcolsep{6pt}

\end{document}